\documentclass{article}
\usepackage[utf8]{inputenc}
\usepackage{amsmath}
\usepackage{amsfonts}
\usepackage{amsthm}
\usepackage{graphicx}
\usepackage[colorinlistoftodos]{todonotes}
\usepackage[most]{tcolorbox}
\usepackage[colorlinks=true, allcolors=blue]{hyperref}
\usepackage{algorithm}

\usepackage{amssymb}
\usepackage[noend]{algpseudocode}
\usepackage[normalem]{ulem}
\usepackage{caption}
\usepackage{subcaption}
\usepackage{multirow}
\newcommand{\Delete}[1]{}
\graphicspath{ {./images/} }

\newlength{\rightside}
\usepackage[header,title,page,toc]{appendix}
\usepackage[toc]{appendix}
\usepackage[T1]{fontenc}
\usepackage{listings}
\title{ORPilot: A Production-Oriented Agentic LLM-for-OR Tool for Optimization Modeling}

\author{Guangrui Xie \thanks{Corresponding author: guanx92@vt.edu}}
\date{
    \today
    }

\begin{document}
\maketitle 
\abstract{This paper presents ORPilot, an open-source agentic AI system that translates
real-world natural-language business problems into working and solver-ready
optimization models. Unlike prior LLM-for-OR tools created in a pure academic setting, which assume a clean and complete problem specification with pre-formatted data embedded in problem text, ORPilot is designed
from the ground up for production conditions: ambiguous problem descriptions,
large-scale raw operational data, and the need for reproducibility and portability across different solver backends.
The system achieves this through four novel components rarely present in existing
work: (1) a conversational interview agent that elicits a complete problem
specification before any modeling efforts begin, (2) a data collection agent that collects data independently instead of embedding data into prompts, (3) a parameter computation agent
that bridges the gap between raw tabular data and model-ready parameters, and
(4) a solver-agnostic Intermediate Representation (IR) that enables
deterministic, zero-LLM-call recompilation to different solver backends (Gurobi, CPLEX, PuLP, Pyomo, or OR-Tools). Moreover, self-correcting retry loops are introduced to feed solver errors and tracebacks back to the LLMs
for targeted repairs. To my knowledge, ORPilot is the first-ever attempt for creating a LLM-for-OR tool that is targeted against production-level business problems, rather than textbook operations research (OR) problems. I evaluated ORPilot on real-world sized problems and obtained promising results. Although textbook OR problems are not in the comfort zone of ORPilot, I also tested ORPilot on three traditional academic benchmarks -- IndustryOR,
NL4OPT, and NLP4LP. ORPilot delivered performance superior to other state-of-the-art LLM-for-OR tools in terms of the
overall accuracy on the industrial IndustryOR benchmark, and comparable performance on NL4OPT and NLP4LP.} \\

\vspace{0.5em}

\noindent \textbf{Code:} \url{https://github.com/GuangruiXieVT/ORPilot}

\section{Introduction}\label{sec:introduction}
Operations Research (OR) has long been one of the most practically valuable
disciplines in applied mathematics, underpinning decisions in logistics, supply
chain management, scheduling, finance, and beyond. Despite decades of
mature solver technologies (e.g., Gurobi, CPLEX, etc.), the barrier to
building and deploying optimization models remains high. Industry practitioners who want to solve a business problem through optimization often lack expertise in mathematical programming, while a well-trained OR consultant may lack domain knowledge. Translating a real-world
business problem into a mathematical optimization model often requires iterative dialogs between a domain expert and an OR consultant.

Large Language Models (LLMs) have opened a plausible path to democratizing
optimization modeling. A growing body of work, e.g., OptiMUS \cite{AhmadiTeshnizi2023,AhmadiTeshnizi2024,AhmadiTeshnizi2025}, OR-LLM \cite{Huang2025},
and others, have demonstrated that state-of-the-art LLMs can generate correct solver code for well-specified linear programming (LP) and mixed integer programming (MIP) problems. These results, however, have been obtained almost exclusively in a controlled academic setting: the problem description is complete and unambiguous, all numerical data appear inline in the prompt, and the task reduces to code synthesis from a mathematics textbook problem.

However, real-world OR deployments are categorically different. Industry practitioners are barely able to describe all aspects of the business problem they are facing in a one-go prompt. A business analyst describing a production-planning problem will likely omit assumptions they consider obvious, conflate decision variables with parameters, and provide data in operational spreadsheets rather than as a neatly formatted numerical table in the prompt. Existing LLM-for-OR tools fall short on addressing these issues.

ORPilot is aimed at addressing this gap. It is, to my knowledge, the first LLM-based OR tool designed explicitly for production conditions, featuring the following architectural novelties:

1. An interview agent that conducts a structured dialog with the user, asking targeted clarifying questions until the problem specification is complete and unambiguous before any modeling efforts begin.

2. A data collection agent that instructs the user on what data to prepare with clear schema requirements for the data tables.

3. A parameter computation agent that inspects raw data tables (e.g., CSV files) the user provides, identifies what derived quantities (e.g., distance matrices, aggregated parameters, BigM values) the model requires, generates and executes a Python script to compute them, and exposes the results as additional computed data tables.

4. A solver-agnostic Intermediate Representation (IR) -- a compact, typed JSON schema capturing sets, parameters, variables, objectives, and constraints with references to the CSV files generated in the previous steps as data sources. The IR decouples model formulation from solver-specific code and enables deterministic recompilation, reproducibility, and backend portability without further LLM calls.

These components are embedded in a self-correcting pipeline: LLM-generated code that fails at the solver level is automatically repaired by feeding error messages, infeasibility signals, and full tracebacks back to the LLM for a targeted revision, up to a configurable number of retries.

I evaluated the performance of ORPilot using three optimization problems in the supply chain industry with real-life sized data, mirroring how an OR consultant would perform OR modeling for business users in practice. Although ORPilot is not designed for academic benchmarks, I also evaluated ORPilot on three public academic benchmarks (IndustryOR \cite{Huang2025}, NL4OPT \cite{Ramamonjison2023}, NLP4LP \cite{AhmadiTeshnizi2023,AhmadiTeshnizi2024,AhmadiTeshnizi2025}) and reported accuracy figures to provide a comparison with other existing LLM-for-OR tools. On IndustryOR (the most industrially representative benchmark), ORPilot with GPT-4o outperforms OptiMUS-0.3 (37\% accuracy \cite{AhmadiTeshnizi2025}) with 46\% accuracy, while ORPilot with Claude-Sonnet-4.6 achieves 75\% overall accuracy on the same benchmark.

The remainder of this paper is organized as follows. Section 2 surveys related work. Section 3 describes the system architecture in detail. Section 4
presents the experimental results. Finally, section 5 concludes the paper with some discussions on the experimental results and future improvements.

\section{Relevant Work}\label{sec:literature}
The application of LLMs to OR has grown rapidly into a broad field. A recent survey paper \cite{Wang2025} provided a systematic review covering three research pathways: automatic
modeling (translating natural language into solver code), LLM-assisted
optimization (e.g., heuristic generation, evolutionary algorithms), and direct solving (LLM as black-box optimizer). ORPilot belongs to the automatic modeling pathway and is most directly related to prompt-driven code generation systems. Hence, the discussion in this section is organized accordingly based on this survey paper.

\subsection{Prompt-Driven Automatic Modeling}
The dominant approach to LLM-based OR modeling is to construct a prompt that
contains the problem description and ask the LLM to generate solver code or a
mathematical formulation in a single pass or a fixed multi-step pipeline.

The OptiMUS series \cite{AhmadiTeshnizi2023,AhmadiTeshnizi2024,AhmadiTeshnizi2025} is the most developed work in this direction. The initial version OptiMUS progressively converts natural language into a solvable optimization model through structure generation, code generation, execution validation and feedback correction, and introduced the NLP4LP dataset. OptiMUS-0.2 added a connection-graph mechanism to track dependencies between variables, constraints and parameters. The latest version, OptiMUS-0.3, further integrated a structure detection agent and structure pool, combined with prompt-retry and self-correction loops, and reported 37\% accuracy on the IndustryOR benchmark using GPT-4o. Chain-of-Experts \cite{Xiao2024} proposed a multi-agent chain in which role-specific agents handle terminology parsing, model construction, code generation, and verification, with a two-stage reasoning mechanism of forward construction and backward reflection. OR-LLM-Agent \cite{Zhang2025} introduced a reasoning-driven closed-loop framework that uses structured prompts to transform natural language into LP models, automatically generates executable code, and performs repairs and validations in a sandbox -- the closest in spirit to ORPilot's retry loop. Autoformulator \cite{Astorga2025} combined hierarchical decomposition and Monte Carlo Tree Search (MCTS) with structured templates to systematically explore alternative formulations. NL4OPT \cite{Ramamonjison2023} first introduced the shared task and dataset for formulating LP problems from natural language while establishing a standard evaluation protocol.

Beyond prompting, a parallel line of work combines structured prompts with
model fine-tuning to improve accuracy and robustness on complex problems.
ORLM \cite{Huang2025} constructed the IndustryOR benchmark -- the most industrially representative dataset in this area, and trained the ORLM series of fine-tuned models that outperformed GPT-4 on NL4OPT, MAMO, and
IndustryOR. LLMOPT \cite{Jiang2025} defined a unified five-element modeling structure: sets, parameters, variables, objectives, and constraints, and used supervised fine-tuning, model alignment, and self-correction to significantly improve modeling reliability. This five-element schema is structurally similar to ORPilot's intermediate representation (IR); however, LLMOPT uses it as a training target for a fine-tuned LLM, whereas ORPilot uses it as an executable IR that is compiled deterministically to solver code without further LLM calls, enabling
reproducibility and backend portability. LLaMoCo \cite{Ma2024} introduced code-to-code instruction tuning with a large-scale instruction set, achieving stronger semantic understanding and generalization in code generation for optimization tasks. OptMATH \cite{Lu2025}
constructed high-quality triplet datasets (natural language, mathematical expression, solver code) with feedback-based difficulty adjustment.

All of these systems discussed above share a common assumption: the problem description is
complete and unambiguous, and all numerical data appear inline with clean formats. ORPilot relaxes both assumptions.

\subsection{Intermediate Representation in OR Modeling}
The use of structured intermediate representations (IR) between natural language and
solver code is an emerging theme. NL2OR \cite{Li2024} generated a
domain-specific language (DSL) as an intermediate step with grammar correction and JSON Schema validation before producing solver code. EquivaMap \cite{Zhai2025} used LLM-generated variable mappings to assess the semantic equivalence of different formulations of the same problem on the NLP4LP dataset.

ORPilot's IR goes further: it is a fully typed, Pydantic-validated JSON schema that captures the complete model structure; it is compiled deterministically to any of the five solver backends (Gurobi, CPLEX, PuLP, Pyomo, OR-Tools) with zero LLM calls, and
persists as a portable artifact that can be archived, edited, and re-solved independently of any LLM connection.

\subsection{Interactive and Conversational OR Systems}
OptiGuide \cite{Li2023} integrated GPT-4 with a supply chain
optimizer, translating user queries into optimization code modifications and forming a closed feedback loop achieving over 90\% accuracy in Microsoft Azure deployment. However, OptiGuide assumes an existing fully specified model. It handles what-if queries on top of a fixed formulation rather than eliciting the formulation from scratch. OptLLM \cite{Zhang2024} enabled multi-round conversation between the user and the system to elicit problem specifications and handles external data outside prompts. However, the tool still assumes model-ready data is available and lacks an IR for reproducibility and portability. Decision Optimization CoPilot (DOCP) \cite{Wasserkrug2024} articulated the research vision for conversational decision-optimization tools, identifying problem elicitation, data handling, and reproducibility as open challenges. ORPilot is, to my knowledge, the first concrete system to address all these challenges.

\subsection{Parameter Derivation from Raw Data}
The gap between raw operational data and model-ready parameters has received almost no attention in the LLM-for-OR literature. \cite{Peng2025} proposed an automatic MILP construction framework for multi-robot task allocation that
uses a domain knowledge base to guide variable and constraint generation, but assumed the relevant numerical quantities are already available. OR-LLM-Agent \cite{Zhang2025} incorporated a validation chain but did not address parameter
transformation from raw tables.

ORPilot's parameter computation agent is, to my knowledge, the first component specifically designed to inspect raw tabular data, identify what derived quantities (e.g., pairwise distance matrices, aggregated parameters, BigM values) the model requires, generate and execute a Python script to compute them, and pass the results as additional structured inputs to the code generation step.

\section{System Architecture}\label{sec:framework}
\begin{figure}[h]
	\centering%
	{\includegraphics[width=15cm]{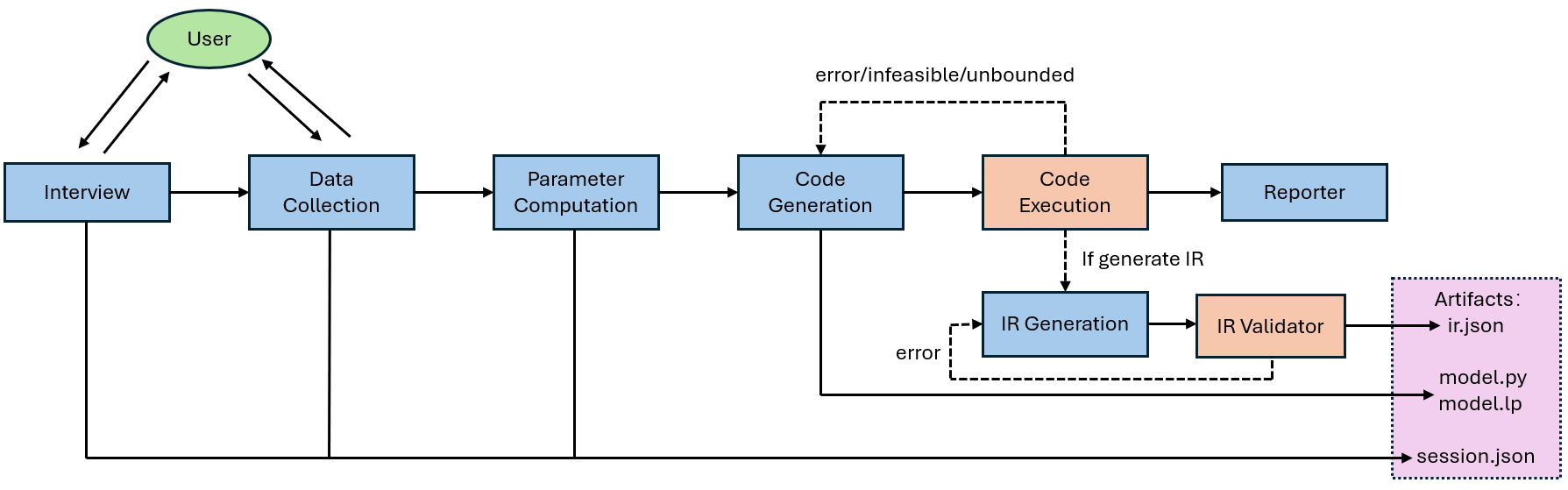}}
	\caption{ORPilot standard pipeline. Blue indicates an LLM-involved step, while orange indicates a deterministic step. Double arrows in opposite directions indicate the interactive nature of this step. Solid arrows represent unconditional transitions between steps, while dashed arrows represent conditional transitions between steps.  }
	\label{fig:pipeline}
\end{figure}

ORPilot is implemented as a LangGraph state machine \cite{langgraph2024}. Each pipeline step
is an independently testable agent node. The standard pipeline of ORPilot is illustrated in Figure
\ref{fig:pipeline}. 

The pipeline starts by the interview agent asking the user questions about the business problem to be solved through an interactive conversation. Once the interview agent completes the conversation with the user, the data collection agent thinks about what data would be needed to model this problem and gives instructions to the user on what data to prepare and how to prepare them through an interactive conversation. Once the data collection agent decides that all needed data are provided in the correct format, the problem specification and the collected data schema will be passed to the parameter computation agent to derive any secondary parameters that are required by the optimization model, but not readily available in the raw data the user provided. Then the code generation agent receives the information of problem specification, raw data tables, and derived parameter tables to generate the code for building and solving the optimization model. The solver code is then executed in a secured sandbox to try to find the optimal solution. A self-correcting loop is embedded here to fix any mistake that the code generation agent could have made when the solver execution step returns an error or infeasible or unbounded solution. The maximum number of retries is limited by a user-defined parameter to prevent infinite loops. Upon successful code executions returning an optimal solution, the solution is passed to the reporter agent to summarize the optimization results. The reporter agent will communicate the findings to the user in plain English. Useful artifacts can be generated and saved along with the pipeline for later analysis the user might need to do. These artifacts include (1) an intermediate representation (IR) which is a structured json blob that extracts the skeleton of this optimization problem, (2) a python file that stores the code for generating the optimization model, (3) the lp file of this particular optimization model and (4) a session file that stores necessary information along with the pipeline which allows the user to resume from the current session later if this session is interrupted. The following subsections discuss the necessity and contribution of each novel component of ORPilot standard pipeline independently.  

\subsection{Interview Agent}

The interview agent is the entry point for end users. It receives a free-form
natural-language description of a business problem which may be vague,
incomplete, or self-contradictory and engages the user in a structured
dialog to elicit a complete problem specification.

The key design principle is that no modeling efforts should begin until the interview
is complete. This mirrors the practice of a human OR consultant. Business users are barely able to describe the business problems they want to model accurately enough in a single paragraph. A good
consultant will spend substantial time with business users to understand the problem before writing a single constraint. Premature modeling based on an underspecified problem produces models that are technically correct but practically wrong. This design of ORPilot is fundamentally different than previous work on LLM-for-OR tools, which are aimed at solving textbook OR problems with clean problem descriptions and model-ready data.

The interview agent is prompted to (a) identify information gaps in the current
description, (b) ask at most one targeted clarifying question per turn to
avoid overwhelming the user, and (c) terminate the interview and advance the
pipeline once the objective function, decision variables, constraints, and data
requirements are unambiguously specified. The agent maintains a running
structured summary of the problem that is updated after each exchange and
passed to downstream steps. Before ending the interview, the agent would provide the user with a summary clearly stating the objective, decision variables, constraints, parameters and indices of each parameter. The user has the chance to make any final corrections by communicating with the agent before this structure summary is passed to downstream processes. Refer to Section \ref{sec:conv_supply_chain_interview}, \ref{sec:conv_mip_interview} and \ref{sec:conv_freight_assignment_interview} for real examples of interview agent -- user conversations. 

\subsection{Data Collection Agent}

This step has no counterpart in any existing LLM-for-OR tool. Existing tools are limited to accepting a single, pre-written problem text with clean and model-ready data embedded in it. However, in practice, solving a real-life optimization problem often requires a huge amount of data that can hardly be directly embedded in the prompt fed to LLMs, due to the maximum token limits. Even if the maximum token limits are not a hurdle, combining data into the prompt will likely increase the chance of hallucinations of LLMs. ORPilot is designed with the notion that data should live separately from the prompt and can only be accessed by AI agents through writing code. Hence, ORPilot adopts a data collection agent dedicated to understand what data the user has available and instruct the user on data preparation.

Based on the problem specification the interview agent obtained from the user, the data collection agent determines exactly
what tabular data is required: which entities (sets) exist, what attributes
(parameters) each entity has, and what data types and semantics apply. The agent
presents this schema to the user and waits for them to supply data as CSV files to the file path specified by the agent. It then validates the completeness and consistency of the uploaded data before
proceeding. If the user does not have the model-ready data to satisfy the schema specified by the agent (e.g., the agent asks for the distance between two locations, but the user only has coordinates of the two locations), the user can indicate this to the data collection agent, and the agent will update the schema requirement based on what the user actually has at hand and let the parameter computation agent compute model-ready parameters by passing a message to it. Refer to Section \ref{sec:conv_supply_chain_data_collection}, \ref{sec:conv_mip_data_collection} and \ref{sec:conv_freight_assignment_data_collection} for real examples of data collection agent -- user conversations. 

% Crucially, the agent distinguishes between the data the user has (e.g., raw
% transaction logs, city coordinates, historical demand) and the data the model
% needs (e.g., distance matrices, per-period aggregate costs). This distinction
% motivates the next step: parameter computation.

\subsection{Parameter Computation Agent}
This step is perhaps the most significant departure from prior work. Most existing LLM-for-OR tools assume that all numerical quantities needed by the model appear directly in the problem statement or attached data. This is essentially never true in practice. Consider two representative examples:
\begin {enumerate}

\item A vehicle routing model needs a distance matrix between locations. The user only has city coordinates. Computing pairwise Euclidean or geographic distances requires a transformation step that is entirely outside the scope of LP/MIP formulation.

\item A multi-period production scheduling model needs aggregate demand per time period. The user only has a transaction ledger with one row per order. The model parameter is a sum-aggregation over the raw data.
\end {enumerate}

The parameter computation agent bridges this gap. It receives the structured problem specification from the interview agent and the raw CSV files from the data collection agent. It then (1) identifies which model parameters cannot be read directly from the raw tables (derived parameters), (2) generates a Python script to compute these derived parameters from the raw data, (3) executes the script in a sandboxed environment, and (4) writes the results as additional CSV files, which are passed to the code generation step.

This ensures that the code generation step always receives model-ready parameters (clean, correctly typed, and correctly indexed, rather than raw operational
data), which results in substantially more reliable code generation and fewer
retries.

\subsection{Code Generation and Self-Correcting Retry Loop}
The code generation step receives the full problem specification and all
parameter CSV files and generates a complete Python solver script for the chosen
backend. ORPilot currently supports five backends (Gurobi, CPLEX, PuLP, Pyomo and OR-Tools), and the
generated code is backend-specific.

The generated code is immediately executed. If the solver returns an error
(syntax error or runtime exception) or infeasibility signal or unbounded signal, the full error message and traceback are fed back to the LLM along with the previously generated code, and code generation is retried. This loop runs up to $N$ times
(configurable). In practice, the majority of failures are resolved within one or two retries. Refer to Section \ref{sec:code_for_supply_chain}, \ref{sec:code_for_mip} and \ref{sec:code_for_freight_assignment} for real examples of the code generated by the code generation agent. 

\subsection{Intermediate Representation (IR)}
A central design goal of ORPilot is reproducibility and portability: once a
model has been solved successfully, it should be possible to re-run it with a
different solver backend, to perform what-if analysis by editing the data, or to archive the model in a form that does not depend on a live LLM connection.

To achieve this, ORPilot defines a solver-agnostic Intermediate Representation
(IR) -- a compact, typed JSON schema that captures all information needed to
reconstruct any optimization model:

  1. Sets -- named collections of entities (e.g., Plants, Products, Periods)
  
  2. Parameters -- indexed numerical quantities, with references to CSV files (indicating where to read the parameter data)
  
  3. Variables -- decision variable types, domains, bounds, and index structure
  
  4. Objective -- a symbolic expression tree over variables and parameters
  
  5. Constraints -- named symbolic expressions with domains and senses

IR generation is optional and governed by a user-specified configuration parameter. When enabled, a LLM is invoked to produce the IR. Due to the inherently non-deterministic nature of LLMs, the generation process is constrained by a set of predefined rules embedded within the prompt to prevent hallucinations. After generation, the IR is passed to a validator to ensure correctness. The validator is designed to detect common failure modes associated with LLM-generated IR. If validation fails, the identified errors are fed back to the agent, which then regenerates the IR.

Compilation from IR to solver code is a fully deterministic process enabled by an IR compiler, with zero LLM calls. This means that once an IR has been validated, it can be recompiled to any supported backend at any time without
incurring API costs or non-determinism. 

% - Aliased sets: the same set appearing twice in a summation domain (e.g., distance[i,j] over all (i,j) location pairs) is handled via alias notation in the IR (e.g., "Locations:l1", "Locations:l2"), producing
% correctly distinct loop variables.

% - Temporal lag: inventory balance constraints of the form inventory[t] = inventory[t-1] + purchase[t] - demand[t]
% are expressed with a "lag": -1 annotation on variable references, which the compiler translates into an enumerate()-based loop with a boundary guard for the first period.

% - Diagonal exclusion: arc variables (e.g., flow[i,j] where i != j) carry an "exclude_diagonal": true flag; the compiler emits the appropriate filtering guards for all three backends.

% - MTZ subtour elimination: the IR supports set_size expression nodes that compile to len(SetName), enabling the standard MTZ big-M formulation for vehicle routing without hardcoded constants.

\begin{figure}[h]
	\centering%
	{\includegraphics[width=15cm]{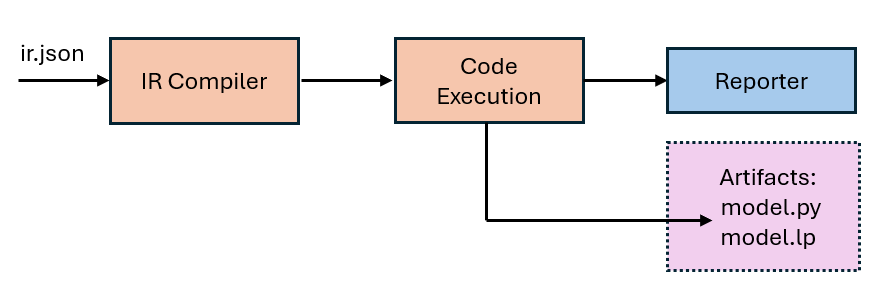}}
	\caption{IR compilation pipeline. Blue indicates an LLM-involved step, while orange indicates a deterministic step.}
	\label{fig:ir_pipeline}
\end{figure}

The existence of the IR enables a lightweight IR compilation pipeline in which a pre-built ir.json file together with CSV files can be compiled and solved with zero LLM calls, see Figure \ref{fig:ir_pipeline}. This pipeline is useful for CI/CD validation and solver regression testing and also crucial to what-if analyses that business users usually perform to support decision making in practice. A more detailed discussion about the role of IR in what-if analysis can be found in Section \ref{sec:what-if}.

\subsection{Reporter Agent}
After a successful solve, a reporter agent generates a plain-English summary of the solution: the optimal objective value, the values of key decision variables, and a brief interpretation of key findings and operational suggestions in the domain language of the original problem. This output is intended for consumption by the business user who initiated the problem description, not by an OR expert. Refer to Section \ref{sec:report_for_supply_chain}, \ref{sec:report_for_MIP} and \ref{sec:report_for_freight_assignment} for real examples of optimization reports produced by the reporter agent.

\section{Experiments}\label{sec:experiments}
This section presents the experimental results obtained when testing ORPilot against various OR problems. Since ORPilot is targeted at real-world optimization problems, Section \ref{sec:supply_chain}, \ref{sec:MIP} and \ref{sec:freight_assignment} evaluate ORPilot on three classic OR problem instances that business users would commonly encounter in practice with real-world sized data. Finally, Section \ref{sec:academic_benchmark} evaluates ORPilot on three academic benchmark datasets in order to compare the performance of ORPilot with other existing LLM-for-OR tools. Refer to \url{https://github.com/GuangruiXieVT/ORPilot} for access to the data used for testing and the testing outputs in Section \ref{sec:supply_chain}, \ref{sec:MIP} and \ref{sec:freight_assignment}.

\subsection{Large-Scale Supply Chain Network Optimization (Pure LP)} \label{sec:supply_chain}
Consider a multi-period and multi-product supply chain network optimization problem defined over the following sets: $I$ -- set of production sites (PS), $J$ -- set of distribution centers (DC), $K$ -- set of customers, $P$ -- set of products, and $T$ -- set of periods. Not all $(i,j)$ and $(j,k)$ pairs are connected. The network topology is defined by a sparse set of links $E_{IJ} \in I \times J$ and $E_{JK} \in J \times K$. The decision variables are: $x_{ipt}$ -- units of product $p$ produced at PS $i$ in period $t$, $f_{ijpt}$ -- units of product $p$ shipped from PS $i$ to DC $j$ in period $t$ defined only for $(i,j) \in E_{IJ}$, $f_{jkpt}$ -- units of product $p$ shipped from DC $j$ to customer $k$ in period $t$ defined only for $(j,k) \in E_{JK}$,
$u_{ipt}$ -- end-of-period inventory of product $p$ at PS $i$ in period $t$, 
$u_{jpt}$ -- end-of-period inventory of product $p$ at DC $j$ in period $t$. All variables are continuous and non-negative and initial inventory is zero at all facilities (PSs and DCs). The objective is to maximize annual profit:
\begin{equation*}
{maximize} \sum_{j,k,p,t} r_p * f_{jkpt} - \sum_{i,p,t} c^{p}_{ip}*x_{ipt} - \sum_{i,j,p,t} c^{t}_{ij}*f_{ijpt} - \sum_{j,k,p,t} c^{t}_{jk}*f_{jkpt} - \sum_{i,p,t} c^h_{ip}*u_{ipt} - \sum_{j,p,t}   c^h_{jp}* u_{jpt},
\end{equation*}
where $r_p$ is the revenue of product $p$, $c^p_{ip}$ is the cost for producing one unit of product $p$ at PS $i$, $c^{t}_{ij}$ is the transportation cost for shipping one unit of any product from PS $i$ to DC $j$, $c^{t}_{jk}$ is the transportation cost for shipping one unit of any product from DC $j$ to customer $k$, $c^h_{ip}$ is the inventory holding cost for one unit of product $p$ at PS $i$ and $c^h_{jp}$ is the inventory holding cost for one unit of product $p$ at DC $j$. The problem is subject to the following constraints:
\begin{eqnarray*}
production \ \ capacity: \ \ \ \sum_p x_{ipt} &\le& cap_{it} \ \ \ \  \forall i \in I, t \in T  \\
flow \ \ balance \ \ at \ \ production \ \ sites: \ \ \ u_{ip,t-1} + x_{ipt} &=& \sum_j f_{ijpt} + u_{ipt} \ \ \ \  \forall i \in I, p \in P, t \in T  \\
inventory \ \ limit \ \ at \ \ production \ \ sites: \ \ \ \sum_p u_{ipt} &\le& lim_i \ \ \ \ \forall i \in I, t \in T \\
flow \ \ balance \ \ at \ \ distribution \ \ centers: \ \ \ u_{jp,t-1} + \sum_i f_{ijpt} &=& \sum_k f_{jkpt} + u_{jpt} \ \ \ \ \forall j \in J, p \in P, t \in T \\
throughput \ \ capacity \ \ at \ \ distribution \ \ centers: \ \ \ \sum_{p,k} f_{jkpt} &\le& cap_{jt} \ \ \ \ \forall j \in J, t \in T \\
inventory \ \ limit \ \ at \ \ distribution \ \ centers: \ \ \ \sum_p u_{jpt} &\le& lim_j  \ \ \ \ \forall j \in J, t \in T \\
demand \ \ requirements: \ \ \ \sum_j f_{jkpt} &=& dem_{kpt}  \ \ \ \ \forall k \in K, p \in P, t \in T,
\end{eqnarray*}
where $cap_{it}$ is the production capacity at PS $i$ in period $t$, $cap_{jt}$ is the throughput capacity at DC $j$ in period $t$, $limit_{i}$ is the inventory limit at production site $i$, $limit_{j}$ is the inventory limit at distribution center $j$, and $dem_{kpt}$ is the demand of customer $k$ for product $p$ in period $t$.

To prepare the data for this problem instance, all numerical data were generated synthetically with a fixed random seed 42 to ensure
reproducibility. The problem instance comprises 50 PSs, 50 DCs,
500 customers, 500 products, and 12 monthly planning periods. All facilities and customers are placed on a $1000 \times 1000$ unit grid.
PSs are drawn uniformly from the interior band $[200, 800]^2$, DCs from the near-full grid $[50, 950]^2$, and customers are sampled from 20 Gaussian clusters whose centers are drawn uniformly from $[100, 900]^2$, with per-cluster standard deviation $\sigma = 30$ units. Each PS is connected to its 8 nearest DCs by a directed link,
yielding a sparse PS$\to$DC subnetwork; isolated DCs (those with no incoming link) are connected to their single nearest PS as a fallback, resulting in 404 PS$\to$DC links in total. Similarly, each DC is connected to its 20 nearest customers for the DC$\to$customer links; any customer left uncovered is connected to its nearest DC,
giving 1{,}065 DC$\to$customer links in total. Transportation costs are proportional to Euclidean distance (\$0.05 per product unit per distance unit on the PS$\to$DC leg and \$0.08 per product unit per distance unit on the DC$\to$customer leg) with independent multiplicative noise that follows a distribution
$U(0.9, 1.1)$. Unit revenues are drawn from a log-normal distribution with parameters $\mu = 3.9$ and $\sigma = 0.6$, giving a median price of approximately \$49, a mean of approximately \$59,
and a coefficient of variation of 0.66. Production costs for each (PS, product) pair are set to a site-level efficiency factor
drawn from distribution $U(0.30, 0.55)$ multiplied by the product revenue, multiplied by independent noise that follows distribution
$U(0.85, 1.15)$.
Inventory holding costs are set to $U(1.0\%, 2.0\%)$ of product revenue per period at
PSs and $U(0.8\%, 1.8\%)$ at DCs. Capacities are set after demand data to make sure the problem instance is feasible.
For each DC $j$, the throughput capacity is set to
$1.5 D_{jt}$ (with $\pm 5\%$ multiplicative noise), where $D_{jt}$ is the
total demand of all customers that are reachable by DC $j$ in period $t$.
For each production site $i$, the production capacity is set to
$1.5 D_{it}$ (with $\pm 5\%$ multiplicative noise), where
$D_{it}$ is the total demand of all customers that are reachable by PS $i$ in period $t$.
This allocation ensures that the aggregate production capacity across all sites is at least
$1.5\times$ total system demand in every period, while each site individually covers its
mandatory supply obligations.
A floor of 10{,}000 units is applied to all capacity parameters.
Storage capacities at PSs are drawn from distribution $U(5{,}000, 12{,}000)$ and
at DCs from distribution $U(8{,}000, 15{,}000)$. The resulting system-wide production-to-demand ratio ranges from 1.52 in December (the peak demand month) to 2.49, with a period-average of approximately 1.98. Each customer is assigned a random subset of 40--80 products (drawn uniformly without replacement from the full product set).
Per-product base demand for each customer is drawn from a log-normal distribution
with $\mu = 2.5$ and $\sigma = 0.6$, and scaled by a monthly seasonal multiplier that
follows the pattern
$s_t \in \{0.85, 0.80, 0.90, 1.00, 1.05, 1.10, 1.15, 1.10, 1.05, 1.00, 0.95, 1.30\}$
for months $t = 1, \ldots, 12$, peaking in December ($+30\%$).
An independent uniform noise factor that follows distribution $U(0.8, 1.2)$ is applied to each
(customer, product, period) triple, and entries that round to zero are dropped.
The resulting demand table contains 361{,}092 non-zero entries across
500 customers, 500 products, and 12 periods. The resulting number of variables and number of constraints are summarized in Table \ref{table:number}. This is a large-scale pure LP problem with 9,714,000 variables and 963,492 constraints. 

\begin{table}[h]
	\centering
	{\small
	\begin{tabular}{ l l l l c l l l }
		\hline
		& \multicolumn{3}{c}{Number of Variables}  & & \multicolumn{3}{c}{Number of Constraints} \\
 		\cline{2-4}  \cline{6-8}
 		& production & shipment & inventory & & inventory balance & demand & capacity \\ 
		\hline
		Individual & 300,000 & 8,814,000 &  600,000  && \ \ \ \ \ 600,000 & 361,092  & 2,400 \\
		Total & & 9,714,000  & &   & & 963,492    \\
		\hline
	\end{tabular}
	}
    \caption{Number of variables and number of constraints}
    \label{table:number}
\end{table}

ORPilot was then tested on this problem instance. In consideration of both cost and speed, I selected Deepseek-Chat model as the LLM for this experiment. Considering the scale of this problem, Gurobi was selected as the optimization solver. The purpose of this experiment is to test the performance of each component in the ORPilot standard pipeline in Figure \ref{fig:pipeline}. The test results for each component are presented in the following paragraphs in sequence. 
\begin{enumerate}
\item Interview Agent. A standard ORPilot pipeline starts with a conversation with the interview agent. The whole conversation with the interview agent obtained from this experiment is presented in Section \ref{sec:conv_supply_chain_interview}. We see that the interview agent was able to guide the user to describe the problem thoroughly and ask important clarifying questions like ``does this model handle a single product or multiple products" to fully grasp every detail of the model. When the agent thought it had all the necessary information to formulate the model, it provided a summary of the problem and asked the user to confirm whether any important details was not captured. This offered the user an opportunity to correct problem specifications if needed. For example, the interview agent automatically assumed that the inventory storage capacity, production cost, inventory holding cost and revenue were time-dependent parameters and there were initial inventories at production sites and distribution centers. However, later the user added a clarification that the inventory storage capacity, production cost, inventory holding cost and revenue did not vary across time, and that there were no initial inventories. Then the agent was able to correct the problem specifications accordingly by removing the time index from these parameters and making initial inventory parameter optional. 

\item Data Collection Agent and Parameter Computation Agent. After the interview is finished, the standard pipeline calls the data collection agent and parameter computation agent to prepare the data needed for modeling this problem. The data collection agent instructs the user on what CSV files are needed and what are the corresponding schema through an interactive conversation, and then the data computation agent decides if any parameters need to be computed based on the raw data the user provided. The conversation between the data collection agent and the user is presented in Section \ref{sec:conv_supply_chain_data_collection}. We see that the agent clearly listed out all the 12 CSV files required for modeling this problem along with the associated schema. Once the user provided the required CSV files, the parameter computation agent then worked on the files and decided no additional CSV file was needed for modeling the problem.

\item Code Generation Agent. Once parameter computation is finished, ORPilot reads the generated CSV files and serializes them into a JSON object named ``data". Then the code generation agent generates the solver code based on the problem specifications and the JSON object. The generated solver code is presented in Section \ref{sec:code_for_supply_chain}. 

\item Code Execution. ORPilot executes the generated solver code in a sandbox, parses the optimization results and produces the output CSV files. Specifically, in this problem, the code was executed successfully without any error, and 5 output CSV files containing the values of the decision variables were produced. The details of these 5 output CSV files can be found in Section \ref{sec:report_for_supply_chain}.

\item Reporter Agent. Once the optimization model is solved successfully, the reporter agent looks at the output CSV files and produces a report that explains what each output CSV file contains and main findings in the solution. The report generated for this problem is presented in Section \ref{sec:report_for_supply_chain}.

\item IR generation. The generated IR for this problem is presented in Section \ref{sec:ir_supply_chain}.
The correctness of the IR was tested by recompiling it to solver code and executed with the same data in a secured sandbox environment. The process returned the same solution as that from the code execution step of the original run. 

\end{enumerate}

\subsection{Large-Scale Supply Chain Network Optimization with Binary Decisions (MIP)} \label{sec:MIP}

Section \ref{sec:supply_chain} shows that ORPilot is able to tackle a pure LP problem with real-life sized data. In this experiment, I further escalated this problem by introducing binary decision variables to make it a MIP problem and evaluated the performance of ORPilot. 

The binary decisions added to this problem are the opening decisions of PSs and DCs. In any period, the user is allowed to open a PS or a DC with a fixed opening cost incurred. Once a PS or a DC is opened, it remains open for the rest of the periods. For any period in which a PS or DC is open, a fixed operating cost will be incurred. The fixed opening cost and operating cost numbers are generated as random draws calibrated to be economically meaningful relative to
the revenue and flow volumes in the problem instance. To make the problem solvable in a reasonable amount of time, 25 PSs (out of 50), 25 DCs (out of 50) and 250 customers (out of 500) were retained from the original LP problem, while all 500 products and 12 monthly periods were kept unchanged. The transportation links were subsetted accordingly,
reducing PS$\to$DC links from 404 to 109 and DC$\to$customer links from 1,065 to 299.

The objective of this MIP problem can be given as:
\begin{eqnarray*}
{maximize} && \sum_{j,k,p,t} r_p * f_{jkpt} - \sum_{i,p,t} c^{p}_{ip}*x_{ipt} - \sum_{i,j,p,t} c^{t}_{ij}*f_{ijpt} - \sum_{j,k,p,t} c^{t}_{jk}*f_{jkpt} - \sum_{i,p,t} c^h_{ip}*u_{ipt} - \sum_{j,p,t}   c^h_{jp}* u_{jpt}\\
&& - \sum_{i} c^o_{i}*o_{i,t=12} -\sum_{j} c^o_{j}*o_{j,t=12} - \sum_{i,t} c^{op}_i*o_{it} - \sum_{j,t} c^{op}_j*o_{jt},
\end{eqnarray*}
where $c^o_{i}$ is the fixed opening cost of PS $i$, $c^o_{j}$ is the fixed opening cost of DC $j$, $c^{op}_i$ is the fixed operating cost of PS $i$, and $c^{op}_j$ is the fixed operating cost of DC $j$, $o_{it}$ and $o_{jt}$ are binary decision variables denoting whether PS $i$ and DC $j$ are open in period $t$. One trick to calculate the fixed opening cost for a PS or DC is to multiply the opening cost with $o_{it}$ or $o_{jt}$ in the last period, due to the assumption that once a facility is open it remains open for the rest of the periods. This problem is subject to the following constraints:

\begin{eqnarray*}
production \ \ capacity: \ \ \ \sum_p x_{ipt} &\le& cap_{it} * o_{it} \ \ \ \  \forall i \in I, t \in T  \\
flow \ \ balance \ \ at \ \ production \ \ sites: \ \ \ u_{ip,t-1} + x_{ipt} &=& \sum_j f_{ijpt} + u_{ipt} \ \ \ \  \forall i \in I, p \in P, t \in T  \\
inventory \ \ limit \ \ at \ \ production \ \ sites: \ \ \ \sum_p u_{ipt} &\le& lim_i \ \ \ \ \forall i \in I, t \in T \\
flow \ \ balance \ \ at \ \ distribution \ \ centers: \ \ \ u_{jp,t-1} + \sum_i f_{ijpt} &=& \sum_k f_{jkpt} + u_{jpt} \ \ \ \ \forall j \in J, p \in P, t \in T \\
throughput \ \ capacity \ \ at \ \ distribution \ \ centers: \ \ \ \sum_{p,k} f_{jkpt} &\le& cap_{jt} * o_{jt} \ \ \ \ \forall j \in J, t \in T \\
inventory \ \ limit \ \ at \ \ distribution \ \ centers: \ \ \ \sum_p u_{jpt} &\le& lim_j  \ \ \ \ \forall j \in J, t \in T \\
demand \ \ requirements: \ \ \ \sum_j f_{jkpt} &=& dem_{kpt}  \ \ \ \ \forall k \in K, p \in P, t \in T\\
production \ \ site \ \ opening \ \ logic: \ \ \ o_{it} &\ge& o_{i,t-1}  \ \ \ \ \forall i \in I, t = 2, 3, ..., 12 \\
distribution \ \ center \ \ opening \ \ logic: \ \ \ o_{jt} &\ge& o_{j,t-1}  \ \ \ \ \forall j \in J, t = 2, 3, ..., 12 \\
linking \ \ inflows \ \ to \ \ open \ \ status \ \ at \ \ distribution \ \ centers: \sum_{p,i} f_{ijpt} &\le& (cap_{jt}+lim_j) * o_{jt} \ \ \ \ \forall j \in J, t \in T. \\
\end{eqnarray*}
This formulation results in 2,898,000 continuous variables, 600 binary variables, and 2,102,050 constraints. 

ORPilot was then tested on this problem instance. As the problem was escalated to a MIP model, I selected a more capable LLM, Deepseek-Reasoner model for this experiment. Gurobi was selected as the optimization solver. The test results for each component are presented in the following paragraphs in sequence.
\begin{enumerate}
\item Interview Agent. The whole conversation with the interview agent obtained from this experiment is presented in Section \ref{sec:conv_mip_interview}. The conversation is similar to that in Section \ref{sec:conv_supply_chain_interview}. Regarding the newly added facility opening decisions, the agent was able to ask questions to clarify the facility opening rules.

\item Data Collection Agent and Parameter Computation Agent. The conversation between the data collection agent and the user is shown in Section \ref{sec:conv_mip_data_collection}. The data collection agent specified 18 CSV files needed for this problem with their corresponding schema and the parameter computation agent calculated a BigM value to be used in constraints linking continuous variables to binary variables. 

\item Code Generation Agent. The generated solver code is given in Appendix \ref{sec:code_for_mip}. An interesting finding is that from the comments the LLM wrote, it realized the trick of calculating the total fixed opening cost by multiplying the opening cost with $o_{it}$ or $o_{jt}$ in the last period, however, it still decided to adopt a more explicit approach by adding up $c^o_{i}*(o_{i,t}-o_{i,t-1})$ and $c^o_{j}*(o_{j,t}-o_{j,t-1})$ across all periods except for the first period, which is also correct.

\item Code Execution. In this problem, the code was executed successfully without any error, and 7 output CSV files containing the values of the decision variables were produced. The details of these 7 output CSV files can be found in Section \ref{sec:report_for_MIP}.

\item Reporter Agent. The report generated for this problem is presented in Section \ref{sec:report_for_MIP}. 

\item IR generation. The IR generated for this problem is presented in Section \ref{sec:ir_MIP}. The correctness of the IR was tested by recompiling it to solver code and executed with the same data in a secured sandbox environment. The process returned the same solution as that from the code execution step of the original run. 
\end{enumerate}

\subsection{Large-Scale Freight Assignment Problem (Pure IP)} \label{sec:freight_assignment}

Consider a freight assignment problem, where the user has a set of shipments $I$ to assign to a set of carriers $J$. Each shipment has its own revenue $r_i$, and the assignment cost for assigning shipment $i$ to carrier $j$ is $c_{ij}$. The decision variables are $x_{ij}$, which are binary variables representing whether shipment $i$ is assigned to carrier $j$. Any shipment can be left unassigned and there is no penalty for not assigning a shipment. The objective of is to maximize the profit for assigning the shipments:

\begin{equation*}
{maximize} \sum_{ij} (r_i-c_{ij})x_{ij}.
\end{equation*}

This problem is subject to the following constraints:

\begin{eqnarray*}
carrier \ \ capacity: \ \ \ \sum_i o_{ij}x_{ij} &\le& cap_{j} \ \ \ \  \forall j \in J \\
each \ \ shipment \ \ assigned \ \ at \ \ most \ \ once: \ \ \ \sum_j x_{ij} &\le& 1 \ \ \ \  \forall i \in I, \\
\end{eqnarray*}
where $o_{ij}$ is the capacity the shipment $i$ occupies on carrier $j$ if shipment $i$ is assigned to carrier $j$ and $cap_j$ is the capacity of carrier $j$.

All data are generated synthetically with a fixed random seed 42 to ensure reproducibility. The instance comprises 400
carriers and 3,200 shipments, yielding a pure IP problem with 1,280,000 binary decision variables. Each carrier is assigned a fixed depot location drawn uniformly from a 1,000 × 1,000 unit grid. Each shipment has an origin point, also drawn uniformly from the same grid.
Each carrier has a per-unit-distance cost rate drawn from distribution $U(0.04, 0.12)$ and a handling efficiency factor drawn from distribution $U(0.85, 1.15)$. Carrier weight capacities are set after shipment weights are generated: the target per-carrier capacity is $1.3 \times (\text{total demand} / \text{number of carriers})$, and individual capacities are drawn from distribution $U(0.7, 1.3)$ times that target, giving a fleet-wide capacity-to-demand ratio of 1.31 and a per-carrier range of 670 --1,240 kg. Shipment weights are drawn from a log-normal distribution with parameters $\mu = 4.4$ and $\sigma = 0.5$ (median $\approx$ 81 kg), truncated at a minimum of 5 kg. Unit revenue for each shipment is its weight multiplied by a price factor drawn from distribution $U(2.5, 6.0)$, yielding revenues between \$57 and \$2,181. For each shipment-carrier pair $(i, j)$, the assignment cost is computed as cost $\text{rate}_j \times w_i \times d_{ij} / 100 + \epsilon_{ij}$, where $w_i$ is the weight of shipment $i$, $d_{ij}$ is the Euclidean distance from carrier $j$'s depot to shipment $i$'s origin and $\epsilon_{ij} \sim U(20, 80)$ is a handling fee. The capacity consumed when carrier $j$ takes shipment $i$ is $r_{ij} = w_i \times h_j \times \eta_{ij}$, where $h_j \sim U(0.85, 1.15)$ is the carrier's loading efficiency and $\eta_{ij} \sim U(0.92, 1.08)$ is a per-pair noise term. This carrier-specific resource consumption is what distinguishes the instance as a generalized assignment problem rather than a standard bin-packing formulation.

ORPilot was then tested on this problem instance with Deepseek-Chat as the LLM and Gurobi as the optimization solver. The test results for each component are presented in the following paragraphs in sequence.
\begin{enumerate}
\item Interview Agent. The whole conversation with the interview agent obtained from this experiment is presented in Section \ref{sec:conv_freight_assignment_interview}. This is a conceptually easier model to formulate than the problems in Section \ref{sec:supply_chain} and \ref{sec:MIP}, hence the agent needed fewer rounds of conversations to confirm the problem specifications.

\item Data Collection Agent and Parameter Computation Agent. The conversation between the data collection agent and the user is shown in Section \ref{sec:conv_freight_assignment_data_collection}. The data collection agent specified 5 CSV files needed for this problem
with their corresponding schema. The parameter computation agent didn't compute any additional parameters for this problem.

\item Code Generation Agent. The generated solver code is given in Section \ref{sec:code_for_freight_assignment}.

\item Code Execution. In this problem, the code was executed successfully without any error, and only one output CSV file containing the values of the decision variables was produced. The details of the produced output CSV file can be found in Section \ref{sec:report_for_freight_assignment}.

\item Reporter Agent. The report generated for this problem is presented in Section \ref{sec:report_for_freight_assignment}.

\item The generated IR for this problem is presented in Section \ref{sec:ir_assignment}. A recompilation of the IR with the same data showed exactly the same solution. 

\end{enumerate}

\subsection{Academic Benchmarks}\label{sec:academic_benchmark}
Existing LLM-for-OR tools and ORPilot are targeted at different types of problems. While most existing tools strive to improve the accuracy on solving textbook style OR problems (unambiguous problem descriptions in one go and clean model-ready data), ORPilot is aimed at more industrial style real-life OR problems with large scale data. However, in order to compare ORPilot with other tools, I evaluated the accuracy performance of ORPilot on three academic benchmarks: IndustryOR, NL4OPT and NLP4LP, which existing tools are usually evaluated on, although these three benchmarks are not quite in the comfort zone of ORPilot.  

Since ORPilot is originally designed for dialog-based problem description intake, not for prepared problem text, the standard pipeline of ORPilot cannot be directly applied to the academic benchmarks. Hence, I designed a benchmark mode for the ORPilot pipeline specially for running benchmark tests.

IndustryOR \cite{Huang2025} is a benchmark of 100 real-world industrial OR problems drawn from supply chain, scheduling, routing, and resource allocation. Problems are labeled as easy (39 instances), medium (41 instances), or hard (20 instances). It is the most industrially representative of the three benchmarks evaluated here. NL4OPT \cite{Ramamonjison2023} is an academic benchmark of natural-language linear programming
problems, originally introduced as a shared task at NeurIPS 2022. Each problem is a self-contained textbook-style LP with numerical data embedded in the problem text. I evaluated on the test split and restricted to problems that have a finite optimal solution (skipping infeasible or unbounded instances). NLP4LP \cite{AhmadiTeshnizi2023,AhmadiTeshnizi2024,AhmadiTeshnizi2025} is a gated benchmark (udell-lab/NLP4LP) containing a diverse set
of LP problems with natural-language descriptions and associated parameter
tables. It is more challenging than NL4OPT in that the problems are longer, more ambiguously stated, and include parameter data as structured JSON rather than inline text.

For the existing LLM-for-OR tools to compare, I selected three representative works to compare: LLMOPT \cite{Jiang2025}, ORLM \cite{Huang2025} and OptiMUS \cite{AhmadiTeshnizi2023,AhmadiTeshnizi2024,AhmadiTeshnizi2025} due to their similar pipeline to ORPilot. The test results are summarized in Table \ref{table:comparison}. All test results of ORPilot were obtained using the pulp solver backend.

\begin{table}[h]
	
	\centering
{	\small
	\begin{tabular}{ l l l  c l l c l l }
		\hline
	\multirow{2}{*}{} &
	 \multicolumn{2}{c}{IndustryOR}  &	 & \multicolumn{2}{c}{NL4OPT} && \multicolumn{2}{c}{NLP4LP}\\
 	\cline{2-3}  \cline{5-6} \cline{8-9}
 	&  \multicolumn{1}{c}{SA} & \multicolumn{1}{c}{AST} & &  \multicolumn{1}{c}{SA} & \multicolumn{1}{c}{AST} &&  \multicolumn{1}{c}{SA} & \multicolumn{1}{c}{AST} \\ 
	\hline
	 LLMOPT with Qwen1.5-14B & 46.0\%  & 8.35s  && 93.0\% & 3.07s && 83.8\% & 7.00s\\
%	\hline 
	ORLM with LLaMa3-8B & 38.0\% & \ \ \ - & & 85.7\% & \ \ \ - &&  \ \ \ - &  \ \ \ -\\
%	\hline
    ORLM with Deepseek-Math & 33.0\% & \ \ \ - & & 86.5\% & \ \ \ - && 72.9\% &  \ \ \ -\\
%	\hline
	OptiMUS0.3 with GPT-4o & 37.0\% & \ \ \ - & & 86.6\% & \ \ \ - && 73.7\% &  \ \ \ -\\
%	\hline 
    OptiMUS0.3 with o1 & 46.0\% & \ \ \ - & & \ \ \ - & \ \ \ - && 80.6\% &  \ \ \ -\\
%	\hline 
	ORPilot with Claude Snonnet 4.6 &  75.0\% & 31.0s & & 73.2\% & 16.3s  && 79.8\% & 14.8s\\
    ORPilot with Deepseek-Reasoner &  74.0\% & 421.8s & & 76.6\% & 235.1s  && 76.7\% & 216.3s \\
    ORPilot with GPT-4o &  46.0\% & 16.2s & & 71.0\% & 15.3s  && 69.8\% & 16.1s\\
	\hline
	\end{tabular}
	}
    \caption{Comparison between ORPilot and existing tools on solving accuracy (SA) and average solving time (AST)}
    \label{table:comparison}
\end{table}

On IndustryOR benchmark, a notable observation is that ORPilot with GPT-4o (46\% overall) exceeds the 37\% accuracy reported for OptiMUS-0.3 with GPT-4o \cite{AhmadiTeshnizi2025} on this benchmark, when using the same underlying LLM. Note that GPT-4o is no longer among the top models by 2026. The reason why GPT-4o was selected to be tested with ORPilot is to ensure ORPilot was using the same LLM as OptMUS0.3 for a fair comparison. Another observation is that ORPilot with the most advanced commercial LLMs as of 2026 significantly outperforms other tools with models fine-tuned on similar tasks in terms of accuracy, with DeepSeek-Reasoner achieving an accuracy (74\%) comparable to Claude-Sonnet-4.6 (75\%) but at substantially higher latency. For latency-sensitive applications, Claude-Sonnet-4.6 is the preferred LLM. However, for cost sensitive applications, Deepseek-Reasoner is undoubtedly the top choice. Overall, it can be concluded that ORPilot achieves a better accuracy performance as compared to other tools, due to the more industrial nature of IndustryOR dataset.

On NL4OPT and NLP4LP datasets, however, ORPilot is showing an inferior accuracy performance, as these problems are less industrial in nature. 

% \subsection{Ablation Study}
% In Section \ref{sec:academic_benchmark}, we have seen that ORPilot underperforms other tools on NL4OPT and NLP4LP. I attribute this to the fact that ORPilot has an explicit component dedicated to compute derived parameters from raw data, while other tools don't. In this section, an ablation study is conducted to specifically evaluate the value of the parameter computation component of ORPilot, by evaluating the performance of ORPilot again on the NL4OPT and NLP4LP without the parameter computation agent.

\section{Conclusions and Discussions}
This paper presents ORPilot, a production-oriented LLM-powered AI agent for OR modeling with a design principle that mirrors how OR consultants would work with business users for modeling their problems. By introducing four novel pipeline components -- a conversational interview agent, a data collection agent, a parameter computation agent, and a solver-agnostic
Intermediate Representation (IR), ORPilot addresses the full gap between an
unconstrained business dialog and a deployable and reproducible optimization
model. These components have no counterpart in most existing LLM-for-OR tools, which operate exclusively in the idealized setting of clean, complete and inline problem descriptions.

Experimental results in Section \ref{sec:experiments} show that ORPilot works well with OR problems with large scale data in supply chain industry and delivers a superior performance than existing LLM-for-OR tools when evaluated on industry-style academic benchmark problems. I believe that the pipeline introduced here -- iterative problem elicitation, data collection, automated parameter derivation, and IR-based reproducibility, represents a necessary and general framework for any serious LLM-for-OR production-level deployment, and I hope it will serve as a foundation for future work in this area. Although the experimental results establish ORPilot as a competitive production-grade pipeline prototype, but they also expose a few structural limitations that define the most productive directions for future work.

\subsection{Optimal Solution Validation}
The current pipeline of ORPilot does not have any mechanism to examine the solution when an optimal solution is produced. The retry loop is only triggered when the model generated runs into an error, infeasibility or unboundedness. However, the generated model can be semantically wrong (misidentifying a decision variable,
omitting a constraint, or misinterpreting a bound) yet still produces an optimal solution. In fact, on the three academic benchmarks (IndustryOR, NL4OPT and NLP4LP), most of the failed cases of ORPilot are due to a wrong optimal objective value without entering any retry loop. 

\begin{figure}[h]
	\centering%
	{\includegraphics[width=15cm]{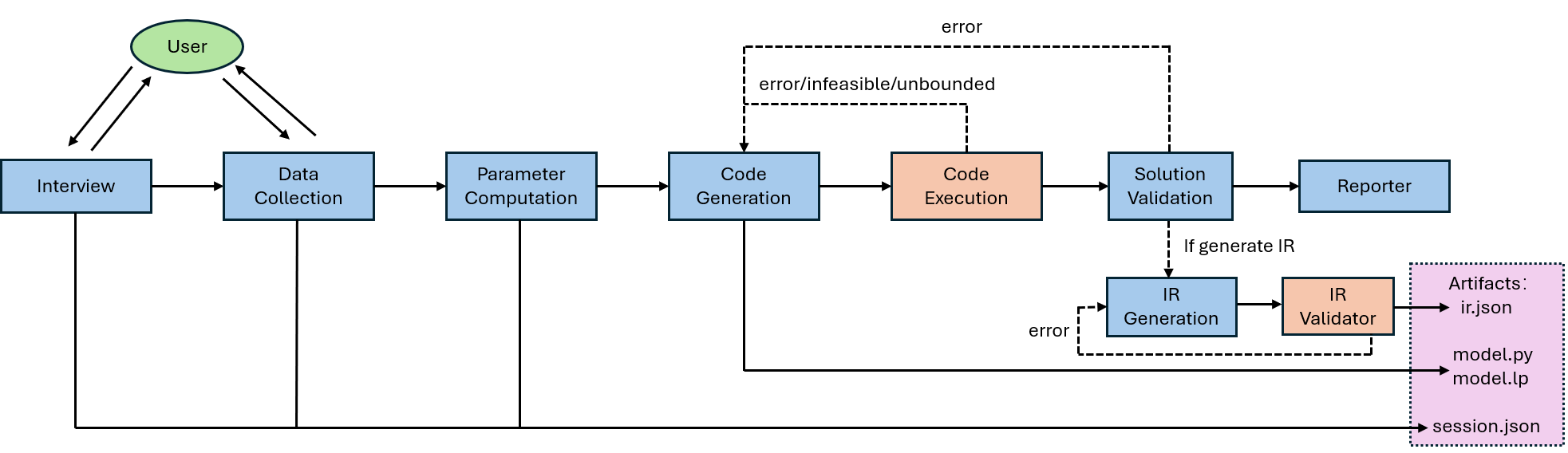}}
	\caption{ORPilot pipeline with a solution validation agent. Blue indicates an LLM-involved step, while orange indicates a deterministic step. Double arrows in opposite directions indicate the interactive nature of this step. Solid arrows represent unconditional transitions between steps, while dashed arrows represent conditional transitions between steps.}
	\label{fig:solution_validation_pipeline}
\end{figure}

Addressing this limitation requires a solution validation layer between the
solver and the reporter. A solution validation agent prompted with the original
problem description and the full solution vector could verify that the
reported values satisfy the business constraints as stated in natural
language, that variable assignments are semantically meaningful, and that
the solution is not merely optimal within a misspecified model. When this agent decides the optimal solution is semantically wrong, it triggers the retry loop to regenerate the model with the new message the solution validation agent produces. This is analogous to the backward-reflection step proposed in Chain-of-Experts \cite{Xiao2024}, but applied to the solution rather than to the formulation. The pipeline with a solution validation agent can be illustrated by the diagram in Figure \ref{fig:solution_validation_pipeline}.

However, adding another LLM-powered agent to examine the semantical correctness of the optimal solution of the generated model is a double-edged sword. While it could identify possible optimal solutions that are semantically wrong, it could also produce false positives (flagging solutions that are semantically correct as wrong and triggering the retry loop), which could turn a correct solution into a wrong solution in the end. Some preliminary experiments I conducted have shown that such an agent would bring more issues than benefits without a fine-tuned prompt or heuristic. Developing robust prompts or heuristics for validating optimal solutions generated by LLM-for-OR tools is an important open problem for the field, not only for ORPilot.

\subsection{Infeasibility and Unboundedness Analysis}
A semantically wrong model generated by LLM-for-OR tools could produce an optimal solution. Conversely, a semantically correct model could run into infeasibility/unboundedness due to the nature of the data provided or the constraints specified (e.g., the total capacity of all facilities is not large enough to satisfy the demands of all customers in a supply chain problem). ORPilot's retry loop treats infeasibility/unboundedness as a recoverable failure: when the solver returns an infeasible signal, the error is fed back to the LLM with a prompt to revise the model, and a new attempt is made. This heuristic is effective when infeasibility/unboundedness is caused by a modeling error (e.g., a misspecified constraint, an incorrect bound, or a missing relaxation). However, it is counterproductive when infeasibility is inherent to the problem as specified. No reformulation of the model will yield a feasible solution because the user's constraints are mutually contradictory or the problem data make the feasible region empty. In this case, the retry loop wastes API calls and latency and ultimately fails after exhausting its budget, reporting a failure that could have been diagnosed immediately.

Two complementary capabilities are needed. First, an infeasibility classifier should distinguish between (1) formulation infeasibility -- the model as written is infeasible due to a modeling error, for which a retry is appropriate  and (2) problem infeasibility -- the constraints implied by the problem description are inherently contradictory, for which the correct response is to inform the user and terminate. A lightweight classifier trained on solver status codes, constraint structure, and LLM-extracted semantic features could make this distinction before invoking the retry loop. Second, for confirmed problem infeasibility, an infeasibility diagnosis agent should apply techniques from the OR literature, e.g., irreducible infeasible subsystems (IIS) extraction or constraint relaxation enumeration, to identify the minimal conflicting subset of constraints and translate it back into natural-language terms the user can understand. This closes the loop for business users: rather than a generic failure message, they receive an explanation such as ``the weekly production capacity constraint conflicts with the minimum demand requirement in week 3 and one of these must be relaxed." Such explanations are standard output from commercial solver tools (e.g., Gurobi, CPLEX) but are not yet integrated into any LLM-for-OR tools.

\subsection{Fine-Tuned Language Models for OR Modeling}\label{sec:fine-tune}

All ORPilot results reported in this paper are based on general-purpose frontier LLMs (e.g., Claude-Sonnet-4.6, DeepSeek-Reasoner) via API. This approach has significant advantages -- no training infrastructure, immediate access to the latest models, but also significant drawbacks -- API cost, latency, data privacy constraints, and a hard dependency on external services for a production system.

The IR structure in ORPilot provides a natural training signal for fine-tuning. Each successful ORPilot run produces a (problem specification, ir.json) pair that is a high-quality supervised training example. The input is an unambiguous natural-language problem description with the schema information of associated data tables needed for building the OR model and the output is a structurally valid, solver-verified IR that compiles correctly to executable code. Unlike raw solver code, which is backend-specific, verbose, and sensitive to API idiom, the IR is compact, backend-agnostic, and schema-validated, making
it a substantially cleaner training target. In practice, a smaller open-weights LLM (e.g., a 7B or 13B parameter model) can be trained to generate IR JSON from problem specifications, using successful and verified ORPilot runs as supervised examples.

The current pipeline of ORPilot uses IR generation as a post-processing step after an optimal solution is generated, see Figure \ref{fig:pipeline}. IR generation was not designed as a step prior to solver code generation in ORPilot in the first place. This is because the experiments I conducted showed LLM-powered IR generation was not very robust and stable using only problem specifications and data table schemas. Feeding artifacts like solver code (model.py) together into LLMs can enhance the quality of the generated IR. 

\begin{figure}[h]
	\centering%
	{\includegraphics[width=15cm]{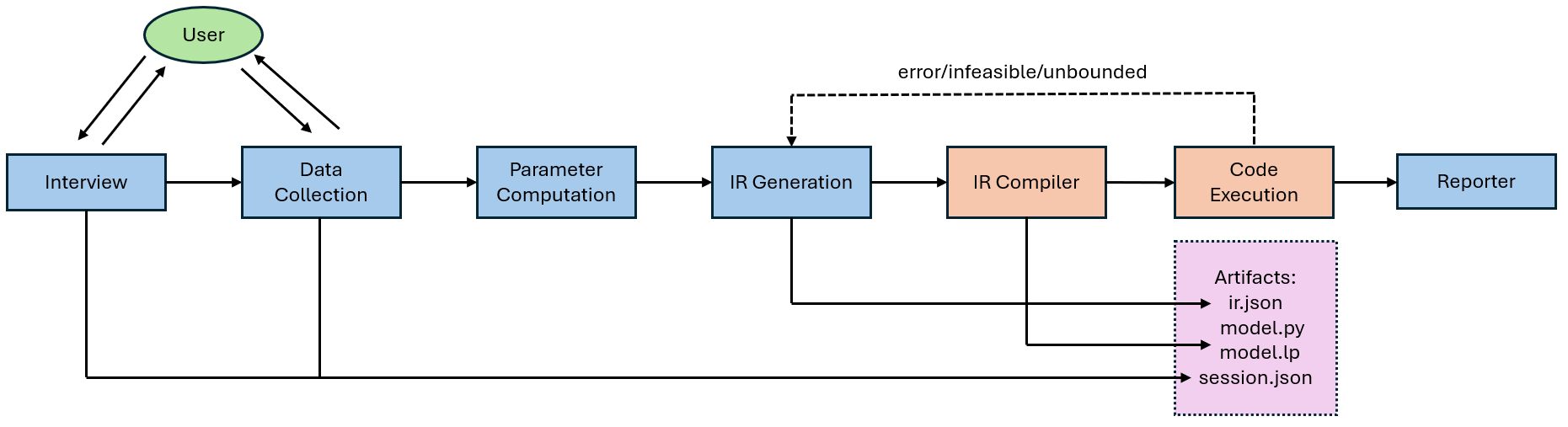}}
	\caption{ORPilot pipeline with IR generation prior to generating solver code. Blue indicates an LLM-involved step, while orange indicates a deterministic step. Double arrows in opposite directions indicate the interactive nature of this step. Solid arrows represent unconditional transitions between steps, while dashed arrows represent conditional transitions between steps.}
	\label{fig:fine_tune_pipeline}
\end{figure}

However, with fine-tuned LLMs that use verified IRs as the learning target, a new pipeline can be established which employs IR generation prior to solver code generation, see Figure \ref{fig:fine_tune_pipeline}. As the fine-tuning process is expected to significantly improve the quality of the generated IR, IR generation can serve as a stable component that would not impact the accuracy of the downstream steps. Since ORPilot has a deterministic IR compiler, solver code generation in this new pipeline will be a deterministic step. This decouples the modeling intelligence (LLM to IR) from the compilation step (IR to solver code), concentrating the LLM's task on structured JSON generation rather than solver API syntax.

This direction is consistent with work on LLM fine-tuning for OR, including
ORLM \cite{Huang2025} and LLaMoCo \cite{Ma2024}, but benefits from the IR's cleaner target structure and the availability of verified training data generated by the ORPilot pipeline itself. 

\subsection{What-if Analysis via IR Manipulation}\label{sec:what-if}
A natural extension of the IR's reproducibility and portability properties is systematic what-if analysis. Once a model has been solved and its IR persists as a validated JSON artifact, a business user may wish to explore how the optimal solution changes under modified assumptions without re-engaging the LLM or re-specifying the problem from scratch.

What-if queries fall into two structurally distinct categories, and the IR structure is able to support both. The first category is data changes -- modifications to parameter values that
leave the model structure intact. Examples include asking ``what if demand
increases by 20\% in Q3?" or ``what if the cost of raw material B rises to \$15 per unit?" In these cases, only the CSV files referenced by the IR need to be updated and the IR JSON itself is unchanged. The deterministic compiler then recompiles the same IR against the new data and a re-solve is triggered, which is a zero-LLM-call operation. This form of sensitivity analysis is already latent in the current architecture and requires only a thin user-facing interface to expose it.

The second category is structural changes: modifications to constraints, objectives or variable domains that alter the model's mathematical form. Examples include ``what if we add a constraint that no single supplier accounts for more than 40\% of total procurement?" or ``what if we remove the overtime limit and instead penalize it in the objective?" These changes require editing the IR -- adding, removing, or modifying constraint or objective terms. Because the IR is a typed, schema-validated document with a well-defined expression tree, such edits are localized and predictable:
adding a constraint is a matter of appending a new constraint object with
the appropriate domain and expression, rather than modifying dozens of lines
of solver-specific code.

\begin{figure}[h]
	\centering%
	{\includegraphics[width=15cm]{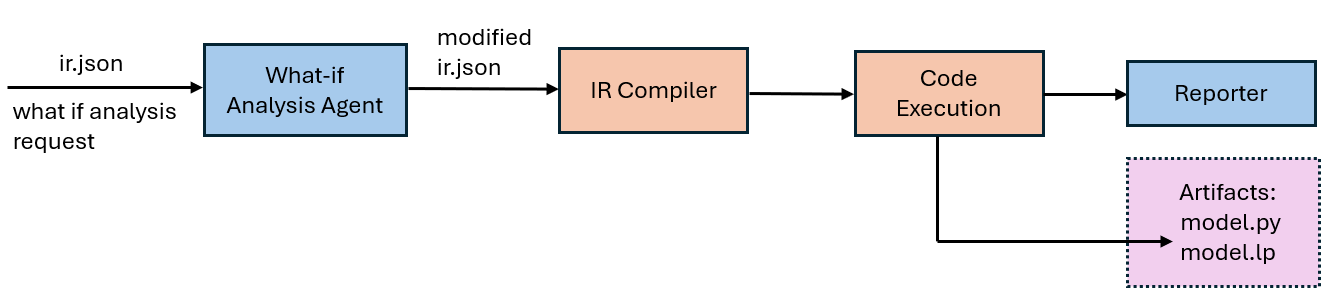}}
	\caption{ORPilot pipeline with a what-if analysis agent. Blue indicates an LLM-involved step, while orange indicates a deterministic step.}
	\label{fig:what_if_pipeline}
\end{figure}

A what-if analysis agent can be added into the current IR compilation pipeline (see Figure \ref{fig:what_if_pipeline}) which accepts natural-language change requests, classifies them as data changes or structural changes, and either updates the relevant values in CSV files or generate a targeted IR modification. The LLM's task in this setting is substantially simpler than full model generation: it operates on a verified IR and only needs to express an incremental change in the same structured schema. This narrower task is well-suited to a fine-tuned or smaller LLM as discussed in Section \ref{sec:fine-tune}.

This capability would bring ORPilot significantly closer to the vision
articulated by \cite{Wasserkrug2024} of a conversational decision-optimization tool that supports iterative exploration of the solution space, not being just a one-shot model builder, which is a significant enabler for enterprise adoption.

\subsection{Database Connection}
A key limitation of the current ORPilot pipeline is its reliance on user-prepared CSV files as the data interface. While this design keeps the system lightweight and portable, it places a significant burden on practitioners, who must extract, transform, and format data from their operational systems before optimization can begin. A natural and impactful extension of ORPilot is to eliminate this preprocessing step through native database connectivity.

In future versions, ORPilot could accept a database connection string -- pointing to a relational database, data warehouse, or enterprise resource planning (ERP) system  and autonomously perform the data engineering tasks currently delegated to the user. This capability would unfold in three stages. First, a schema discovery phase would catalog the available tables, columns, data types, foreign key relationships, and cardinalities, producing a structured representation of what data the system has access to. Second, a data sufficiency assessment phase would evaluate whether the discovered schema contains the entities and relationships required to instantiate the optimization model identified during the interview phase. Drawing on the IR specification, ORPilot would map each required set, parameter, and variable domain to candidate tables in the database, flagging gaps where data is absent or ambiguous and querying the user only for genuinely missing information rather than demanding a complete data hand-off upfront. Third, a model-ready data preparation phase would issue the necessary SQL queries, perform unit validation, resolve naming inconsistencies, and materialize the result as the structured inputs consumed by the IR compiler all transparently, without requiring the user to write a single query.

This architecture would bring ORPilot substantially closer to the vision of end-to-end decision support: a
practitioner could describe their OR problem in natural language, grant the system read access to their existing database, and receive a solved, deployable optimization model with minimal manual intervention. It would also make re-optimization operationally practical -- as the underlying database is updated with new demand signals, cost figures, or capacity changes, the model could be rebuilt and re-solved automatically, enabling
near-real-time decision support rather than one-off studies.

Several technical challenges merit attention in pursuing this direction. Semantic alignment between database schemas and optimization concepts is non-trivial; a table named orders in one organization may correspond to demand in another, and the LLM agent responsible for schema mapping must resolve such ambiguities reliably. Handling sparse or missing data requires robust imputation and flagging strategies so that the resulting model is both tractable and trustworthy. Privacy and access control are also important considerations, as practitioners may only be able to grant query-level access to subsets of their data; the system must respect these boundaries while still assembling complete model inputs. Finally, for very large databases, scalable query planning and data sampling strategies will be necessary to keep the data preparation phase practical within the time constraints of an interactive optimization workflow.

\subsection{Retrieval Augmented Generation (RAG) System}
The current approach to improving ORPilot's reliability follows an empirical, rule-accumulation paradigm: when a failure mode is identified through testing, such as a hallucinated column name, an incorrect sparse network encoding, or a malformed IR expression, a corrective rule is appended to the relevant system prompt. This strategy is effective in the short term, as each new rule directly addresses an observed failure and can be validated immediately
through re-testing. However, it is fundamentally unsustainable at scale. As the library of rules grows, prompts become increasingly long, placing pressure on the context windows of the underlying language models, increasing inference latency and cost, and diluting the signal-to-noise ratio of the prompt itself. A model attending to a prompt containing hundreds of rules may fail to surface the specific guidance most relevant to the problem at hand.

A more principled and scalable alternative is to replace static and monolithic prompts with a retrieval-augmented generation (RAG) architecture. Under this design, ORPilot would maintain a curated knowledge base of high-quality examples, including reference IR specifications for canonical problem classes, correct solver code for representative model structures, and documented patterns for common modeling challenges such as inventory balance constraints, sparse network flows, and multi-period temporal dependencies. Rather than embedding all of this knowledge statically in every prompt, agents would issue a semantic query against the knowledge base before each generation step,
retrieving only the most relevant examples for the specific problem structure currently being modeled. The retrieved examples would be injected dynamically into the prompt context, providing targeted, high-precision guidance without the overhead of irrelevant rules.

This architecture would yield several concrete benefits beyond prompt length control. First, it decouples knowledge maintenance from prompt engineering: adding a new problem type or correcting a modeling pattern becomes an operation on the knowledge base rather than a prompt rewrite, making the system easier to maintain and extend. Second, retrieval-based selection is inherently adaptive, a routing problem would surface arc-flow examples and subtour elimination patterns, while an inventory management problem would retrieve temporal balance constraints and holding cost formulations, without either example polluting the other's prompt context. Third, the knowledge base itself becomes an auditable artifact: practitioners and researchers can inspect which examples were retrieved for a given generation, providing a degree of transparency and debuggability that is difficult to achieve with opaque monolithic
prompts.

However, building an effective RAG system for ORPilot also raises several research questions worth investigating. The design of the embedding space is particularly important: standard semantic similarity over natural language descriptions may be insufficient to distinguish structurally similar but semantically distinct optimization patterns; embeddings that
encode mathematical structure, e.g., constraint types, variable domains, sparsity patterns, may be necessary for high-precision retrieval. The construction and curation of the knowledge base also requires care: examples must be verified correct, sufficiently diverse to cover the problem space, and annotated with enough metadata to support accurate retrieval. Finally, the interaction between retrieval and in-context learning deserves study. The ordering, formatting, and number of retrieved examples all influence generation quality, and optimal retrieval strategies may differ between the IR generation and code generation stages of the pipeline.
\bibliographystyle{IEEEtran}
\bibliography{reference}

\clearpage
\begin{appendices}
\section{Conversation History between Interview Agent and User for Problem in Section \ref{sec:supply_chain}}\label{sec:conv_supply_chain_interview}

% \begin{chatbox}{Interview Agent - User Conversation}{conv1}

\textbf{Agent}: What is the main goal you want to achieve with this optimization model? For example, are you trying to minimize costs, maximize profit, or optimize something else?\\
 
\textbf{User}: I'm looking to maximize the profit of a supply chain network.\\

\textbf{Agent}: What are the key decisions you need to make in this supply chain network? For example, are you deciding which facilities to open, how much to produce, or how much to ship between locations?\\

\textbf{User}: I need to make production decisions for production sites, transportation decisions between production sites and distribution centers, transportation decisions between distribution centers and customers, and inventory decisions for production sites and distribution centers.\\

\textbf{Agent}: Thank you. To clarify, are you making these decisions over a single time period (static model) or over multiple time periods (dynamic model with inventory carried over time)?\\

\textbf{User}: Over multiple time periods. I have 12 periods and each period is one month.\\

\textbf{Agent}: Are there any capacity constraints at the production sites, such as limits on how much can be produced in a given month?\\

\textbf{User}: Yes, there is a production capacity for each production site in each period.\\

\textbf{Agent}: Are there any capacity constraints for transportation between locations, or for inventory storage at production sites and distribution centers?\\

\textbf{User}: There are inventory storage capacities for both production sites and distribution centers. There are throughput capacities for distribution centers that limit the total quantity of products that can be shipped through each distribution center.\\

\textbf{Agent}: Are there any constraints on how much can be shipped along a specific route (e.g., from a given production site to a given distribution center) in a given period?\\

\textbf{User}: No, there are no such constraints.\\

\textbf{Agent}: Are there any other constraints or requirements, such as needing to meet all customer demand each period, or having minimum production levels?\\

\textbf{User}: All customers' demands must be met in each period.\\

\textbf{Agent}: Thank you. Now, to help me structure the parameters, can you please list all the different types of locations involved? For example, production sites, distribution centers, and customers.\\

\textbf{User}: Production sites, distribution centers and customers.\\

\textbf{Agent}: Perfect. Now, to define the indices for the parameters, I need to understand the product scope. Does this model handle a single product, or multiple different products?\\

\textbf{User}: Multiple different products.\\
    
% \end{tcolorbox}
% \end{chatbox}

% \begin{chatbox}{Interview Agent - User Conversation Continued (1)}{conv2}

\textbf{Agent}: Thank you. I now have enough information to summarize my understanding of your problem. Please review the following summary carefully. 

\begin{enumerate}
\item \textbf{Objective:}

Maximize total profit over a 12-month planning horizon.

\item \textbf{Decision Variables:}
\begin{itemize}
\item Production Quantity: How much of each product to produce at each production site in each time period. 

\item Shipment Quantity (First Leg): How much of each product to ship from each production site to each distribution center in each time period. 

\item Shipment Quantity (Second Leg): How much of each product to ship from each distribution center to each customer in each time period. 

\item Inventory Level: The amount of each product held in inventory at each production site and at each distribution center at the end of each time period.
\end{itemize}

\item \textbf{Constraints:}
\begin{itemize}
\item Production Capacity: For each production site and each time period, the total production of all products cannot exceed a given capacity. 

\item Inventory Storage Capacity: For each production site and each distribution center, the total inventory of all products held at the end of a period cannot exceed a given storage capacity for that location. 

\item Distribution Center Throughput Capacity: For each distribution center and each time period, the total quantity of all products shipped out from it (to customers) cannot exceed a given throughput capacity. 

\item Demand Satisfaction: For each customer, product, and time period, the total quantity shipped from all distribution centers must exactly meet the customer's demand. 

\item Inventory Balance: For each product at each production site and distribution center, the inventory level changes according to a standard balance equation (e.g., ending inventory = starting inventory + incoming shipments - outgoing shipments). This links production, shipments, and inventory across time periods.

\item Non-negativity: All decision variables (production, shipments, inventory) must be zero or positive. 
\end{itemize}

\item \textbf{Parameters Needed (with Indices):}
\begin{itemize}
\item \texttt{demand[c, p, t]}: Demand for product $p$ from customer $c$ in period $t$.

\item \texttt{production\_capacity[i, t]}: Maximum total production at production site $i$ in period $t$.

\item \texttt{storage\_capacity[l, t]}: Maximum total inventory storage at location $l$ (which can be a production site $i$ or distribution center $j$) in period $t$.

\item \texttt{throughput\_capacity[j, t]}: Maximum total quantity that can be shipped out from distribution center $j$ in period $t$.

\item \texttt{production\_cost[i, p, t]}: Cost to produce one unit of product $p$ at production site $i$ in period $t$.

\item \texttt{transport\_cost\_1[i, j, p, t]}: Cost to ship one unit of product $p$ from production site $i$ to distribution center $j$ in period $t$.

\item \texttt{transport\_cost\_2[j, c, p, t]}: Cost to ship one unit of product $p$ from distribution center $j$ to customer $c$ in period $t$.

\item \texttt{holding\_cost[l, p, t]}: Cost to hold one unit of inventory of product $p$ at location $l$ at the end of period $t$.

\item \texttt{revenue[p, t]}: Revenue earned per unit of product $p$ sold in period $t$.

\item \texttt{initial\_inventory[l, p]}: Starting inventory level of product $p$ at location $l$ at the beginning of the first period.
\end{itemize}

\item \textbf{Indices}: 

$i\in I$: Set of production sites.

$j \in J$: Set of distribution centers.

$c\in C$: Set of customers.

$p\in P$: Set of products.

$t \in T$: Set of time periods (1 to 12).
\end{enumerate}

The objective function would sum revenue from met demand and subtract total costs (production, transportation, and inventory holding) across all periods.\\

Is there anything else you'd like to add, or anything I may have missed?\\
% \end{chatbox}

% \begin{chatbox}{Interview Agent - User Conversation Continued (2)}{conv3} 
\textbf{User}: The inventory storage capacity, production cost, inventory holding cost and revenue do not change over time, so they don't have to be indexed over periods. The transportation cost does not change over time or depend on products, so doesn't need to be indexed over periods and products. Also the initial inventory for all locations and products are zero.\\

\textbf{Agent}: Thank you for those important clarifications. I will update the parameter list accordingly.

\begin{enumerate}
\item 
\textbf{Objective:}

Maximize total profit over a 12-month planning horizon.

\item
\textbf{Decision Variables:}
\begin{itemize}
\item Production Quantity: How much of each product to produce at each production site in each time period. 

\item Shipment Quantity (First Leg): How much of each product to ship from each production site to each distribution center in each time period.

\item Shipment Quantity (Second Leg):How much of each product to ship from each distribution center to each customer in each time period.

\item Inventory Level: The amount of each product held in inventory at each production site and at each distribution center at the end of each time period.\\
\end{itemize}

\item 
\textbf{Constraints:}
\begin{itemize}
\item Production Capacity: For each production site and each time period, the total production of all products cannot exceed a given capacity.

\item Inventory Storage Capacity: For each production site and each distribution center, the total inventory of all products held at the end of a period cannot exceed a given storage capacity for that location.

\item Distribution Center Throughput Capacity: For each distribution center and each time period, the total quantity of all products shipped out from it (to customers) cannot exceed a given throughput capacity.

\item Demand Satisfaction: For each customer, product, and time period, the total quantity shipped from all distribution centers must exactly meet the customer's demand.

\item Inventory Balance: For each product at each production site and distribution center, the inventory level changes according to a standard balance equation (e.g., ending inventory = starting inventory + incoming shipments - outgoing shipments). This links production, shipments, and inventory across time periods. Initial inventory is zero for all locations and products.

\item Non-negativity: All decision variables (production, shipments, inventory) must be zero or positive.
\end{itemize}

\item 
\textbf{Parameters Needed (with Indices):}
\begin{itemize}
\item \texttt{demand[c, p, t]}: Demand for product $p$ from customer $c$ in period $t$. 

\item \texttt{production\_capacity[i, t]}: Maximum total production at production site $i$ in period $t$.

\item \texttt{storage\_capacity[l]}: Maximum total inventory storage at location $l$ (which can be a production site $i$ or distribution center $j$). Constant over time.

\item \texttt{throughput\_capacity[j, t]}: Maximum total quantity that can be shipped out from distribution center $j$ in period $t$.

\item \texttt{production\_cost[i, p]}: Cost to produce one unit of product $p$ at production site $i$. Constant over time.

\item \texttt{transport\_cost\_1[i, j]}: Cost to ship one unit of any product from production site $i$ to distribution center $j$. Constant over time and product-independent.

\item \texttt{transport\_cost\_2[j, c]}: Cost to ship one unit of any product from distribution center $j$ to customer $c$. Constant over time and product-independent.

\item \texttt{holding\_cost[l, p]}: Cost to hold one unit of inventory of product $p$ at location $l$. Constant over time.

\item \texttt{revenue[p]}: Revenue earned per unit of product $p$ sold. Constant over time.

\item \texttt{initial\_inventory[l, p]}: This parameter is zero for all $l$ and $p$, so it can be omitted from the model data.
\end{itemize}

\item \textbf{Indices:}

$i\in I$: Set of production sites.

$j \in J$: Set of distribution centers.

$c\in C$: Set of customers.

$p\in P$: Set of products.

$t \in T$: Set of time periods (1 to 12).

\end{enumerate}

Is there anything else you'd like to add, or anything I may have missed?\\

\textbf{User}: No.

% \end{chatbox}

\section{Conversation History between Data Collection Agent and User for Problem in Section \ref{sec:supply_chain}}\label{sec:conv_supply_chain_data_collection}
\textbf{Agent}: Based on the problem definition and our discussion, you need to provide the following CSV files. Please ensure the column names and data types match exactly.\\

\begin{enumerate}

\item \textbf{\texttt{sets.csv}}
\begin{itemize}
\item \textit{Description:} Defines all the members of the sets used in the model.

\item \textit{Schema:}
\begin{itemize}
    \item \texttt{set\_name} (str): The name of the set this row belongs to.
    \item \texttt{element} (str): The unique identifier for a member of that set.
\end{itemize}

\item \textit{Required Sets:}
\begin{itemize}
    \item production\_sites (e.g., PS\_001, PS\_002)
    \item distribution\_centers (e.g., DC\_001, DC\_002)
    \item customers (e.g., C\_0001, C\_0002)
    \item products (e.g., P\_001, P\_002)
    \item periods (e.g., 1, 2, ..., 12)
\end{itemize}

\item \textit{Example:}
\begin{verbatim}
set_name,element
production_sites,PS_001
production_sites,PS_002
distribution_centers,DC_001
customers,C_0001
customers,C_0002
products,P_001
products,P_002
periods,1
periods,2
...
periods,12
\end{verbatim}
\end{itemize}

\item \textbf{\texttt{demand.csv}}
\begin{itemize}
\item \textit{Description:} Demand for each product from each customer in each time period.

\item \textit{Schema:}
\begin{itemize}
    \item \texttt{customer\_id} (str): Must match an element from \texttt{customers} in \texttt{sets.csv}.
    \item \texttt{product\_id} (str): Must match an element from \texttt{products} in \texttt{sets.csv}.
    \item \texttt{period} (int): Time period (1--12).
    \item \texttt{demand\_quantity} (float): Quantity demanded.
\end{itemize}

\item \textit{Note:} Missing combinations are treated as zero demand.
\end{itemize}

\item \textbf{\texttt{production\_capacity.csv}}
\begin{itemize}
\item \textit{Description:} Maximum total production at each site per period.

\item \textit{Schema:}
\begin{itemize}
    \item \texttt{site\_id} (str): Production site ID.
    \item \texttt{period} (int): Time period (1--12).
    \item \texttt{capacity} (float): Production capacity.
\end{itemize}

\item \textit{Note:} Missing rows imply unlimited capacity.
\end{itemize}

\item \textbf{\texttt{storage\_capacity\_sites.csv}}
\begin{itemize}
\item \textit{Description:} Inventory storage capacity at production sites (constant over time).

\item \textit{Schema:}
\begin{itemize}
    \item \texttt{site\_id} (str): Production site ID.
    \item \texttt{capacity} (float): Storage capacity.
\end{itemize}

\item \textit{Note:} Missing sites imply unlimited capacity.
\end{itemize}

\item \textbf{\texttt{storage\_capacity\_dcs.csv}}
\begin{itemize}
\item \textit{Description:} Inventory storage capacity at distribution centers (constant over time).

\item \textit{Schema:}
\begin{itemize}
    \item \texttt{dc\_id} (str): Distribution center ID.
    \item \texttt{capacity} (float): Storage capacity.
\end{itemize}
\end{itemize}

\item \textbf{\texttt{throughput\_capacity.csv}}
\begin{itemize}
\item \textit{Description:} Maximum shipment capacity from each DC per period.

\item \textit{Schema:}
\begin{itemize}
    \item \texttt{dc\_id} (str): Distribution center ID.
    \item \texttt{period} (int): Time period (1--12).
    \item \texttt{capacity} (float): Throughput capacity.
\end{itemize}

\item \textit{Note:} Missing rows imply unlimited throughput.
\end{itemize}

\item \textbf{\texttt{production\_cost.csv}}
\begin{itemize}
\item \textit{Description:} Unit production cost at each site for each product.

\item \textit{Schema:}
\begin{itemize}
    \item \texttt{site\_id} (str): Production site ID.
    \item \texttt{product\_id} (str): Product ID.
    \item \texttt{unit\_cost} (float): Production cost per unit.
\end{itemize}

\item \textit{Note:} Missing combinations are infeasible (infinite cost).
\end{itemize}

\item \textbf{\texttt{transport\_cost\_site\_to\_dc.csv}}
\begin{itemize}
\item \textit{Description:} Transportation cost from production sites to DCs.

\item \textit{Schema:}
\begin{itemize}
    \item \texttt{from\_site\_id} (str): Production site ID.
    \item \texttt{to\_dc\_id} (str): Distribution center ID.
    \item \texttt{unit\_cost} (float): Shipping cost per unit.
\end{itemize}

\item \textit{Note:} Missing routes are infeasible.
\end{itemize}

\item \textbf{\texttt{transport\_cost\_dc\_to\_customer.csv}}
\begin{itemize}
\item \textit{Description:} Transportation cost from DCs to customers.

\item \textit{Schema:}
\begin{itemize}
    \item \texttt{from\_dc\_id} (str): Distribution center ID.
    \item \texttt{to\_customer\_id} (str): Customer ID.
    \item \texttt{unit\_cost} (float): Shipping cost per unit.
\end{itemize}

\item \textit{Note:} Missing routes are infeasible.
\end{itemize}

\item \textbf{\texttt{holding\_cost\_sites.csv}}
\begin{itemize}
\item \textit{Description:} Inventory holding cost at production sites (constant over time).

\item \textit{Schema:}
\begin{itemize}
    \item \texttt{site\_id} (str): Production site ID.
    \item \texttt{product\_id} (str): Product ID.
    \item \texttt{unit\_cost} (float): Holding cost per unit per period.
\end{itemize}

\item \textit{Note:} Missing combinations imply zero cost.
\end{itemize}

\item \textbf{\texttt{holding\_cost\_dcs.csv}}
\begin{itemize}
\item \textit{Description:} Inventory holding cost at distribution centers (constant over time).

\item \textit{Schema:}
\begin{itemize}
    \item \texttt{dc\_id} (str): Distribution center ID.
    \item \texttt{product\_id} (str): Product ID.
    \item \texttt{unit\_cost} (float): Holding cost per unit per period.
\end{itemize}

\item \textit{Note:} Missing combinations imply zero cost.
\end{itemize}

\item \textbf{\texttt{revenue.csv}}
\begin{itemize}
\item \textit{Description:} Revenue per unit sold for each product (constant over time).

\item \textit{Schema:}
\begin{itemize}
    \item \texttt{product\_id} (str): Product ID.
    \item \texttt{unit\_revenue} (float): Revenue per unit sold.
\end{itemize}

\item \textit{Note:} Missing products imply zero revenue.
\end{itemize}

\end{enumerate}

\textbf{Important Rules for All CSV Files:}
\begin{itemize}
\item Use exact column names as specified.
\item All IDs must match those defined in \texttt{sets.csv}.
\item Period values must be integers from 1 to 12.
\item All costs and revenues must be non-negative.
\item Initial inventory is zero for all locations and products (no file required).
\end{itemize}

Please prepare these 12 CSV files and place them in:
\texttt{ORPilot/data}\\

Type \textbf{ready} when the files are in place.\\

\textbf{User}: Ready.

\section{Generated Solver Code for Problem in Section \ref{sec:supply_chain}}
\label{sec:code_for_supply_chain}

\begin{lstlisting}[language=Python]
import gurobipy as gp
from gurobipy import GRB

def solve(data, time_limit=None, show_solver_log=False):
    SEP = "\x1f"
    
    # Load all sets from sets.csv
    sets_data = data["sets"]
    production_sites = [r["element"] for r in sets_data if r["set_name"] == "production_sites"]
    distribution_centers = [r["element"] for r in sets_data if r["set_name"] == "distribution_centers"]
    customers = [r["element"] for r in sets_data if r["set_name"] == "customers"]
    products = [r["element"] for r in sets_data if r["set_name"] == "products"]
    periods = [r["element"] for r in sets_data if r["set_name"] == "periods"]
    
    # Load parameters with appropriate defaults
    # Demand: missing = 0
    demand = {}
    for r in data["demand"]:
        key = (r["customer_id"], r["product_id"], r["period"])
        demand[key] = float(r["demand_quantity"])

    # Production capacity: missing = unlimited
    prod_cap = {}
    for r in data["production_capacity"]:
        key = (r["site_id"], r["period"])
        prod_cap[key] = float(r["capacity"])

    # Storage capacities: missing = unlimited
    storage_site = {}
    for r in data["storage_capacity_sites"]:
        storage_site[r["site_id"]] = float(r["capacity"])

    storage_dc = {}
    for r in data["storage_capacity_dcs"]:
        storage_dc[r["dc_id"]] = float(r["capacity"])

    # Throughput capacity: missing = unlimited
    throughput_cap = {}
    for r in data["throughput_capacity"]:
        key = (r["dc_id"], r["period"])
        throughput_cap[key] = float(r["capacity"])
    
    # Production cost: missing = unavailable (inf)
    prod_cost = {}
    for r in data["production_cost"]:
        key = (r["site_id"], r["product_id"])
        prod_cost[key] = float(r["unit_cost"])
    
    # Transportation costs: missing = unavailable route
    trans_cost_site_dc = {}
    for r in data["transport_cost_site_to_dc"]:
        key = (r["from_site_id"], r["to_dc_id"])
        trans_cost_site_dc[key] = float(r["unit_cost"])
    
    trans_cost_dc_cust = {}
    for r in data["transport_cost_dc_to_customer"]:
        key = (r["from_dc_id"], r["to_customer_id"])
        trans_cost_dc_cust[key] = float(r["unit_cost"])
    
    # Holding costs: missing = 0
    hold_cost_site = {}
    for r in data["holding_cost_sites"]:
        key = (r["site_id"], r["product_id"])
        hold_cost_site[key] = float(r["unit_cost"])
    
    hold_cost_dc = {}
    for r in data["holding_cost_dcs"]:
        key = (r["dc_id"], r["product_id"])
        hold_cost_dc[key] = float(r["unit_cost"])
    
    # Revenue: missing = 0
    revenue = {}
    for r in data["revenue"]:
        revenue[r["product_id"]] = float(r["unit_revenue"])
    
    # Create sets of valid connections (sparse network)
    valid_site_dc = set(trans_cost_site_dc.keys())
    valid_dc_cust = set(trans_cost_dc_cust.keys())
    
    # Build neighbor mappings for efficient constraint building
    site_to_dc = {}
    dc_from_site = {}
    for i, j in valid_site_dc:
        site_to_dc.setdefault(i, []).append(j)
        dc_from_site.setdefault(j, []).append(i)
    
    dc_to_cust = {}
    cust_from_dc = {}
    for j, c in valid_dc_cust:
        dc_to_cust.setdefault(j, []).append(c)
        cust_from_dc.setdefault(c, []).append(j)
    
    # Initialize model
    env = gp.Env(empty=True)
    env.setParam("LogToConsole", 1 if show_solver_log else 0)
    env.start()
    m = gp.Model(env=env)
    if time_limit is not None:
        m.Params.TimeLimit = time_limit
    
    # Decision variables
    production = {}
    for i in production_sites:
        for p in products:
            for t in periods:
                # Only create variable if production cost is finite
                if prod_cost.get((i, p), float('inf')) < float('inf'):
                    production[i, p, t] = m.addVar(
                        name=f"production_quantity{SEP}{i}{SEP}{p}{SEP}{t}",
                        lb=0.0
                    )
    
    ship_site_dc = {}
    for i, j in valid_site_dc:
        for p in products:
            for t in periods:
                ship_site_dc[i, j, p, t] = m.addVar(
                    name=f"shipment_site_to_dc{SEP}{i}{SEP}{j}{SEP}{p}{SEP}{t}",
                    lb=0.0
                )
    
    ship_dc_cust = {}
    for j, c in valid_dc_cust:
        for p in products:
            for t in periods:
                ship_dc_cust[j, c, p, t] = m.addVar(
                    name=f"shipment_dc_to_customer{SEP}{j}{SEP}{c}{SEP}{p}{SEP}{t}",
                    lb=0.0
                )
    
    inv_site = {}
    for i in production_sites:
        for p in products:
            for t in periods:
                inv_site[i, p, t] = m.addVar(
                    name=f"inventory_site{SEP}{i}{SEP}{p}{SEP}{t}",
                    lb=0.0
                )
    
    inv_dc = {}
    for j in distribution_centers:
        for p in products:
            for t in periods:
                inv_dc[j, p, t] = m.addVar(
                    name=f"inventory_dc{SEP}{j}{SEP}{p}{SEP}{t}",
                    lb=0.0
                )
    
    m.update()
    
    # Objective: maximize total profit = revenue - costs
    # Revenue from met demand
    revenue_terms = []
    for (c, p, t), d in demand.items():
        # Sum of shipments to customer c for product p in period t
        # Only include if there are valid DCs serving this customer
        if c in cust_from_dc:
            for j in cust_from_dc[c]:
                if (j, c, p, t) in ship_dc_cust:
                    revenue_terms.append(revenue.get(p, 0.0) * ship_dc_cust[j, c, p, t])
    
    # Production costs
    prod_cost_terms = []
    for (i, p, t), var in production.items():
        prod_cost_terms.append(prod_cost.get((i, p), float('inf')) * var)
    
    # Transportation costs
    trans_cost_terms = []
    for (i, j, p, t), var in ship_site_dc.items():
        trans_cost_terms.append(trans_cost_site_dc.get((i, j), float('inf')) * var)
    
    for (j, c, p, t), var in ship_dc_cust.items():
        trans_cost_terms.append(trans_cost_dc_cust.get((j, c), float('inf')) * var)
    
    # Holding costs
    hold_cost_terms = []
    for (i, p, t), var in inv_site.items():
        hold_cost_terms.append(hold_cost_site.get((i, p), 0.0) * var)
    
    for (j, p, t), var in inv_dc.items():
        hold_cost_terms.append(hold_cost_dc.get((j, p), 0.0) * var)
    
    m.setObjective(
        gp.quicksum(revenue_terms) - 
        gp.quicksum(prod_cost_terms) - 
        gp.quicksum(trans_cost_terms) - 
        gp.quicksum(hold_cost_terms),
        GRB.MAXIMIZE
    )
    
    # Constraints
    # 1. Production capacity
    for i in production_sites:
        for t in periods:
            cap = prod_cap.get((i, t), float('inf'))
            if cap < float('inf'):
                prod_vars = []
                for p in products:
                    if (i, p, t) in production:
                        prod_vars.append(production[i, p, t])
                if prod_vars:
                    m.addConstr(gp.quicksum(prod_vars) <= cap, name=f"prod_cap_{i}_{t}")
    
    # 2. Site storage capacity
    for i in production_sites:
        cap = storage_site.get(i, float('inf'))
        if cap < float('inf'):
            for t in periods:
                inv_vars = []
                for p in products:
                    inv_vars.append(inv_site[i, p, t])
                if inv_vars:
                    m.addConstr(gp.quicksum(inv_vars) <= cap, name=f"store_site_{i}_{t}")
    
    # 3. DC storage capacity
    for j in distribution_centers:
        cap = storage_dc.get(j, float('inf'))
        if cap < float('inf'):
            for t in periods:
                inv_vars = []
                for p in products:
                    inv_vars.append(inv_dc[j, p, t])
                if inv_vars:
                    m.addConstr(gp.quicksum(inv_vars) <= cap, name=f"store_dc_{j}_{t}")
    
    # 4. DC throughput capacity
    for j in distribution_centers:
        for t in periods:
            cap = throughput_cap.get((j, t), float('inf'))
            if cap < float('inf'):
                ship_vars = []
                if j in dc_to_cust:
                    for c in dc_to_cust[j]:
                        for p in products:
                            ship_vars.append(ship_dc_cust[j, c, p, t])
                if ship_vars:
                    m.addConstr(gp.quicksum(ship_vars) <= cap, name=f"throughput_{j}_{t}")
    
    # 5. Demand satisfaction (equality, always create even if demand is 0)
    for c in customers:
        for p in products:
            for t in periods:
                d = demand.get((c, p, t), 0.0)
                ship_vars = []
                if c in cust_from_dc:
                    for j in cust_from_dc[c]:
                        if (j, c, p, t) in ship_dc_cust:
                            ship_vars.append(ship_dc_cust[j, c, p, t])
                # Always create constraint, even if d == 0
                m.addConstr(gp.quicksum(ship_vars) == d, name=f"demand_{c}_{p}_{t}")
    
    # 6. Site inventory balance
    for i in production_sites:
        for p in products:
            for idx, t in enumerate(periods):
                # Initial inventory is 0
                if idx == 0:
                    prev_inv = 0.0
                else:
                    prev_t = periods[idx - 1]
                    prev_inv = inv_site[i, p, prev_t]
                
                # Production term (if variable exists)
                prod_term = production.get((i, p, t), 0.0)
                
                # Outgoing shipments
                out_ship = []
                if i in site_to_dc:
                    for j in site_to_dc[i]:
                        out_ship.append(ship_site_dc[i, j, p, t])
                
                m.addConstr(
                    inv_site[i, p, t] == prev_inv + prod_term - gp.quicksum(out_ship),
                    name=f"balance_site_{i}_{p}_{t}"
                )
    
    # 7. DC inventory balance
    for j in distribution_centers:
        for p in products:
            for idx, t in enumerate(periods):
                # Initial inventory is 0
                if idx == 0:
                    prev_inv = 0.0
                else:
                    prev_t = periods[idx - 1]
                    prev_inv = inv_dc[j, p, prev_t]
                
                # Incoming shipments
                in_ship = []
                if j in dc_from_site:
                    for i in dc_from_site[j]:
                        in_ship.append(ship_site_dc[i, j, p, t])
                
                # Outgoing shipments
                out_ship = []
                if j in dc_to_cust:
                    for c in dc_to_cust[j]:
                        out_ship.append(ship_dc_cust[j, c, p, t])
                
                m.addConstr(
                    inv_dc[j, p, t] == prev_inv + gp.quicksum(in_ship) - gp.quicksum(out_ship),
                    name=f"balance_dc_{j}_{p}_{t}"
                )
    
    # Write LP file
    try:
        m.write("model.lp")
    except Exception:
        pass
    
    # Solve
    m.optimize()
    
    # Process results
    _status_map = {
        GRB.OPTIMAL: "optimal",
        GRB.SUBOPTIMAL: "feasible",
        GRB.INFEASIBLE: "infeasible",
        GRB.UNBOUNDED: "unbounded",
    }
    status = _status_map.get(m.Status, "error")
    
    obj = None
    variables = {}
    if status in ("optimal", "feasible"):
        obj = m.ObjVal
        for v in m.getVars():
            variables[v.VarName] = v.X
    
    # Group variables
    _groups = {}
    for key, val in variables.items():
        prefix = key.split(SEP, 1)[0] if SEP in key else key
        _groups.setdefault(prefix, {})[key] = val
    
    _dim_labels = {
        "production_quantity": ["site_id", "product_id", "period"],
        "shipment_site_to_dc": ["from_site_id", "to_dc_id", "product_id", "period"],
        "shipment_dc_to_customer": ["from_dc_id", "to_customer_id", "product_id", "period"],
        "inventory_site": ["site_id", "product_id", "period"],
        "inventory_dc": ["dc_id", "product_id", "period"],
    }
    
    variable_groups = [
        {
            "group_name": g,
            "dimension_labels": _dim_labels.get(g, []),
            "variables": gvars
        }
        for g, gvars in _groups.items()
    ]
    
    return {
        "status": status,
        "objective_value": obj,
        "variables": variables,
        "variable_groups": variable_groups
    }
\end{lstlisting}

\section{Solution Report Produced by Reporter Agent for Problem in Section \ref{sec:supply_chain}}
\label{sec:report_for_supply_chain}
\textbf{Multi-period Multi-product Profit Maximization}
\begin{enumerate}
\item \textbf{Executive Summary}

The optimization model successfully determined the maximum-profit operating plan for a supply chain network spanning \textbf{50 production sites}, \textbf{50 distribution centers}, \textbf{500 customers}, and \textbf{500 products} over \textbf{12 monthly periods}. The solver found a provably optimal solution with a \textbf{total network profit of \$155,729,604.44}.

All 500 customers are fully served across every product and period. The plan achieves this by activating 41 of the 50 available production sites and routing goods through 48 of the 50 distribution centers, using 197 distinct site-to-DC shipping lanes.

\item \textbf{Key Decisions}
\begin{itemize}
\item \textbf{Production}

Total production across the planning horizon is approximately \textbf{5.42 million units}. Production ramps up steadily through the year, peaking in Period 12 (520,674 units), which suggests that late-period demand or favorable economics drive maximum output toward year-end.

\begin{center}
\begin{tabular}{|c|c|}
\hline
\textbf{Period} & \textbf{Total Production (units)} \\
\hline
1 & 387,420 \\
2 & 378,164 \\
3 & 411,817 \\
4 & 444,224 \\
5 & 456,138 \\
6 & 471,652 \\
7 & 490,375 \\
8 & 478,568 \\
9 & 463,644 \\
10 & 456,197 \\
11 & 458,481 \\
12 & 520,674 \\
\hline
\end{tabular}
\end{center}

The top five production sites by total output are:

\begin{center}
\begin{tabular}{|c|c|}
\hline
\textbf{Site} & \textbf{Total Output (units)} \\
\hline
PS\_017 & 514,334 \\
PS\_036 & 494,114 \\
PS\_050 & 490,157 \\
PS\_042 & 379,648 \\
PS\_024 & 263,380 \\
\hline
\end{tabular}
\end{center}

\item \textbf{Facility Utilization}

\begin{itemize}
    \item \textbf{9 production sites are idle throughout the entire planning horizon}: PS\_003, PS\_008, PS\_011, PS\_030, PS\_038, PS\_039, PS\_040, PS\_041, and PS\_047. These sites are either more expensive to operate, have lower capacity, or serve market areas that are better covered by nearby active sites.
    \item \textbf{2 distribution centers are unused}: DC\_001 and DC\_042. All other 48 DCs participate in fulfilling customer demand.
\end{itemize}

\item \textbf{Distribution}

The network ships goods through 197 distinct production site-to-DC lane combinations. The top five DCs by throughput volume are:

\begin{center}
\begin{tabular}{|c|c|}
\hline
\textbf{DC} & \textbf{Total Volume (units)} \\
\hline
DC\_024 & 664,979 \\
DC\_025 & 259,758 \\
DC\_028 & 240,684 \\
DC\_030 & 219,501 \\
DC\_038 & 202,703 \\
\hline
\end{tabular}
\end{center}

DC\_024 handles roughly 12\% of total network volume, making it a critical node. Any disruption to this facility would have an outsized impact on service levels.

\item \textbf{Inventory}

The plan uses inventory as a buffer, building stock in mid-year periods and drawing it down toward year-end. Both production sites and DCs follow a similar seasonal pattern:

\begin{center}
\begin{tabular}{|c|c|c|}
\hline
\textbf{Period} & \textbf{Site Inventory (units)} & \textbf{DC Inventory (units)} \\
\hline
1 & 3,606 & 8,474 \\
2 & 8,112 & 28,640 \\
3 & 10,577 & 39,579 \\
4 & 11,763 & 40,142 \\
5 & 9,719 & 34,606 \\
6 & 7,404 & 22,214 \\
7 & 2,267 & 9,283 \\
8 & 358 & 2,807 \\
9 & 286 & 2,011 \\
10 & 2,154 & 13,958 \\
11 & 9,591 & 44,449 \\
12 & 0 & 0 \\
\hline
\end{tabular}
\end{center}

Inventory peaks at \textbf{40,142 units across DCs in Period 4}, then declines sharply. The DC inventory spike in Period 11 (44,449 units) followed by a full drawdown in Period 12 indicates the model is pre-positioning stock to meet high late-year demand. Zero ending inventory in Period 12 is expected behavior for a finite-horizon model: the solver correctly avoids carrying unsold stock past the planning window.
\end{itemize}

\item \textbf{Notable Findings and Concerns}

\begin{itemize}
    \item \textbf{9 idle production sites}: Nearly 18\% of production capacity is left unused. Before removing these sites from the network or reducing their capacity, it is worth investigating whether they become active under different demand scenarios or cost structures. They may serve as surge capacity.
    \item \textbf{DC\_024 concentration risk}: A single DC handles 12\% of all deliveries. If this center faces disruption (weather, labor, capacity ceiling), a large portion of the customer base is affected. A sensitivity run that caps DC\_024 throughput would reveal how much profit is lost and which alternative routes the network would use.
    \item \textbf{Mid-year inventory build-up}: Holding inventory for 3–4 periods adds holding cost. If holding costs are high, a solution with smoother production and lower peak inventory may be preferable even at a small reduction in revenue. A multi-objective run or an added constraint on maximum inventory level would surface this trade-off.
    \item \textbf{End-of-horizon effect}: Zero inventory in Period 12 is mathematically optimal for this model but may leave the business exposed if demand continues into Period 13. Extending the planning horizon or adding a terminal inventory value would give a more realistic picture.
    \item \textbf{Unused DCs (DC\_001, DC\_042)}: These facilities incur no shipping volume. If there are fixed operating costs not captured in this LP model (e.g., lease costs), their ongoing cost should be weighed against the value of having them available as backup capacity.
\end{itemize}

\item \textbf{Suggested Next Steps}

\begin{itemize}
    \item \textbf{Sensitivity analysis on revenue and production cost}: Small changes in unit revenue or production cost can shift which sites and DCs are active. Run a parametric sweep to identify the break-even cost at which the 9 idle sites become economical.
    \item \textbf{Capacity tightening}: Artificially reduce DC\_024's throughput cap to 80\% of its current utilization and re-solve to understand the cost of concentration risk.
    \item \textbf{Extend the horizon}: Add a Period 13 or a terminal inventory value to eliminate the artificial drawdown effect and get a rolling-horizon plan.
    \item \textbf{Add fixed costs}: If opening or operating a site/DC incurs a fixed periodic cost, upgrade to a MIP model (as in the output\_MIP\_new scenario) to jointly optimize facility selection and flow decisions.
    \item \textbf{Stochastic demand}: Replace deterministic demand with scenarios to produce a robust plan that performs well even if demand deviates from forecast.
\end{itemize}

\item \textbf{Output Files}
\begin{itemize}
\item \textbf{\texttt{solution\_production\_quantity.csv}} \\
Records how many units of each product were produced at each production site in each period. Use this to schedule factory workloads — each row tells a site manager exactly what to produce in a given month. \\
Columns: \texttt{site\_id} (which production site), \texttt{product\_id} (which product), \texttt{period} (month number), \texttt{value} (units to produce).

\item \textbf{\texttt{solution\_shipment\_site\_to\_dc.csv}} \\
Records the volume of each product shipped from each production site to each distribution center in each period. Use this to plan outbound logistics from factories — trucks, freight bookings, and warehouse receiving schedules at DCs. \\
Columns: \texttt{from\_site\_id} (origin site), \texttt{to\_dc\_id} (destination DC), \texttt{product\_id}, \texttt{period}, \texttt{value} (units shipped).

\item \textbf{\texttt{solution\_shipment\_dc\_to\_customer.csv}} \\
Records the volume of each product shipped from each DC to each customer in each period. This is the last-mile delivery plan — use it to generate pick-pack-ship orders and communicate delivery quantities to customers. \\
Columns: \texttt{from\_dc\_id} (origin DC), \texttt{to\_customer\_id} (destination customer), \texttt{product\_id}, \texttt{period}, \texttt{value} (units delivered).

\item \textbf{\texttt{solution\_inventory\_site.csv}} \\
Records the end-of-period inventory of each product held at each production site. Use this to plan warehouse space at factories and to verify that storage capacity constraints are not being violated. \\
Columns: \texttt{site\_id}, \texttt{product\_id}, \texttt{period}, \texttt{value} (units in stock at end of period).

\item \textbf{\texttt{solution\_inventory\_dc.csv}} \\
Records the end-of-period inventory of each product held at each distribution center. Use this to manage DC storage space and to anticipate when inventory build-up is largest (Periods 3–4 and Period 11). \\
Columns: \texttt{dc\_id}, \texttt{product\_id}, \texttt{period}, \texttt{value} (units in stock at end of period).
\end{itemize}
\end{enumerate}

\section{Generated IR for Problem in Section \ref{sec:supply_chain}}
\label{sec:ir_supply_chain}
\begin{lstlisting}[language=Python]
{
  "problem_class": "SupplyChainPlanning",
  "model_type": "Linear Program",
  "sense": "maximize",
  "sets": {
    "ProductionSites": {
      "size": null,
      "index_symbol": "i",
      "source": "sets.csv",
      "column": "element",
      "filter_column": "set_name",
      "filter_value": "production_sites",
      "ordered": false
    },
    "DistributionCenters": {
      "size": null,
      "index_symbol": "j",
      "source": "sets.csv",
      "column": "element",
      "filter_column": "set_name",
      "filter_value": "distribution_centers",
      "ordered": false
    },
    "Customers": {
      "size": null,
      "index_symbol": "c",
      "source": "sets.csv",
      "column": "element",
      "filter_column": "set_name",
      "filter_value": "customers",
      "ordered": false
    },
    "Products": {
      "size": null,
      "index_symbol": "p",
      "source": "sets.csv",
      "column": "element",
      "filter_column": "set_name",
      "filter_value": "products",
      "ordered": false
    },
    "Periods": {
      "size": null,
      "index_symbol": "t",
      "source": "sets.csv",
      "column": "element",
      "filter_column": "set_name",
      "filter_value": "periods",
      "ordered": true
    }
  },
  "parameters": {
    "production_capacity": {
      "domain": [
        "ProductionSites",
        "Periods"
      ],
      "type": "float",
      "source": "production_capacity.csv",
      "column": "capacity",
      "index_columns": [
        "site_id",
        "period"
      ],
      "missing_default": "inf"
    },
    "storage_capacity_site": {
      "domain": [
        "ProductionSites"
      ],
      "type": "float",
      "source": "storage_capacity_sites.csv",
      "column": "capacity",
      "index_columns": [
        "site_id"
      ],
      "missing_default": "inf"
    },
    "storage_capacity_dc": {
      "domain": [
        "DistributionCenters"
      ],
      "type": "float",
      "source": "storage_capacity_dcs.csv",
      "column": "capacity",
      "index_columns": [
        "dc_id"
      ],
      "missing_default": "inf"
    },
    "throughput_capacity": {
      "domain": [
        "DistributionCenters",
        "Periods"
      ],
      "type": "float",
      "source": "throughput_capacity.csv",
      "column": "capacity",
      "index_columns": [
        "dc_id",
        "period"
      ],
      "missing_default": "inf"
    },
    "demand": {
      "domain": [
        "Customers",
        "Products",
        "Periods"
      ],
      "type": "float",
      "source": "demand.csv",
      "column": "demand_quantity",
      "index_columns": [
        "customer_id",
        "product_id",
        "period"
      ],
      "missing_default": "zero"
    },
    "production_cost": {
      "domain": [
        "ProductionSites",
        "Products"
      ],
      "type": "float",
      "source": "production_cost.csv",
      "column": "unit_cost",
      "index_columns": [
        "site_id",
        "product_id"
      ],
      "missing_default": "inf"
    },
    "transport_cost_site_to_dc": {
      "domain": [
        "ProductionSites",
        "DistributionCenters"
      ],
      "type": "float",
      "source": "transport_cost_site_to_dc.csv",
      "column": "unit_cost",
      "index_columns": [
        "from_site_id",
        "to_dc_id"
      ],
      "missing_default": "inf"
    },
    "transport_cost_dc_to_customer": {
      "domain": [
        "DistributionCenters",
        "Customers"
      ],
      "type": "float",
      "source": "transport_cost_dc_to_customer.csv",
      "column": "unit_cost",
      "index_columns": [
        "from_dc_id",
        "to_customer_id"
      ],
      "missing_default": "inf"
    },
    "holding_cost_site": {
      "domain": [
        "ProductionSites",
        "Products"
      ],
      "type": "float",
      "source": "holding_cost_sites.csv",
      "column": "unit_cost",
      "index_columns": [
        "site_id",
        "product_id"
      ],
      "missing_default": "inf"
    },
    "holding_cost_dc": {
      "domain": [
        "DistributionCenters",
        "Products"
      ],
      "type": "float",
      "source": "holding_cost_dcs.csv",
      "column": "unit_cost",
      "index_columns": [
        "dc_id",
        "product_id"
      ],
      "missing_default": "inf"
    },
    "revenue": {
      "domain": [
        "Products"
      ],
      "type": "float",
      "source": "revenue.csv",
      "column": "unit_revenue",
      "index_columns": [
        "product_id"
      ],
      "missing_default": "zero"
    }
  },
  "variables": {
    "production_quantity": {
      "description": "Amount of product p produced at production site i in period t",
      "label": "production_quantities",
      "domain": [
        "ProductionSites",
        "Products",
        "Periods"
      ],
      "type": "continuous",
      "lower_bound": 0,
      "upper_bound": null,
      "domain_filter": "production_cost"
    },
    "shipment_site_to_dc": {
      "description": "Amount of product p shipped from production site i to distribution center j in period t",
      "label": "shipments_site_to_dc",
      "domain": [
        "ProductionSites",
        "DistributionCenters",
        "Products",
        "Periods"
      ],
      "type": "continuous",
      "lower_bound": 0,
      "upper_bound": null,
      "domain_filter": "transport_cost_site_to_dc"
    },
    "shipment_dc_to_customer": {
      "description": "Amount of product p shipped from distribution center j to customer c in period t",
      "label": "shipments_dc_to_customer",
      "domain": [
        "DistributionCenters",
        "Customers",
        "Products",
        "Periods"
      ],
      "type": "continuous",
      "lower_bound": 0,
      "upper_bound": null,
      "domain_filter": "transport_cost_dc_to_customer"
    },
    "inventory_site": {
      "description": "Amount of product p held in inventory at production site i at end of period t",
      "label": "inventory_at_site",
      "domain": [
        "ProductionSites",
        "Products",
        "Periods"
      ],
      "type": "continuous",
      "lower_bound": 0,
      "upper_bound": null,
      "domain_filter": null
    },
    "inventory_dc": {
      "description": "Amount of product p held in inventory at distribution center j at end of period t",
      "label": "inventory_at_dc",
      "domain": [
        "DistributionCenters",
        "Products",
        "Periods"
      ],
      "type": "continuous",
      "lower_bound": 0,
      "upper_bound": null,
      "domain_filter": null
    }
  },
  "constraints": {
    "production_capacity": {
      "domain": [
        "ProductionSites",
        "Periods"
      ],
      "expression": {
        "operation": "indexed_sum",
        "over": [
          "Products"
        ],
        "body": {
          "type": "variable",
          "name": "production_quantity",
          "indices": [
            "i",
            "p",
            "t"
          ]
        }
      },
      "sense": "<=",
      "rhs": {
        "type": "parameter",
        "name": "production_capacity",
        "indices": [
          "i",
          "t"
        ]
      },
      "sparse_filter": "production_capacity"
    },
    "site_storage_capacity": {
      "domain": [
        "ProductionSites",
        "Periods"
      ],
      "expression": {
        "operation": "indexed_sum",
        "over": [
          "Products"
        ],
        "body": {
          "type": "variable",
          "name": "inventory_site",
          "indices": [
            "i",
            "p",
            "t"
          ]
        }
      },
      "sense": "<=",
      "rhs": {
        "type": "parameter",
        "name": "storage_capacity_site",
        "indices": [
          "i"
        ]
      },
      "sparse_filter": "storage_capacity_site"
    },
    "dc_storage_capacity": {
      "domain": [
        "DistributionCenters",
        "Periods"
      ],
      "expression": {
        "operation": "indexed_sum",
        "over": [
          "Products"
        ],
        "body": {
          "type": "variable",
          "name": "inventory_dc",
          "indices": [
            "j",
            "p",
            "t"
          ]
        }
      },
      "sense": "<=",
      "rhs": {
        "type": "parameter",
        "name": "storage_capacity_dc",
        "indices": [
          "j"
        ]
      },
      "sparse_filter": "storage_capacity_dc"
    },
    "throughput_capacity": {
      "domain": [
        "DistributionCenters",
        "Periods"
      ],
      "expression": {
        "operation": "indexed_sum",
        "over": [
          "Customers",
          "Products"
        ],
        "body": {
          "type": "variable",
          "name": "shipment_dc_to_customer",
          "indices": [
            "j",
            "c",
            "p",
            "t"
          ]
        }
      },
      "sense": "<=",
      "rhs": {
        "type": "parameter",
        "name": "throughput_capacity",
        "indices": [
          "j",
          "t"
        ]
      },
      "sparse_filter": "throughput_capacity"
    },
    "demand_satisfaction": {
      "domain": [
        "Customers",
        "Products",
        "Periods"
      ],
      "expression": {
        "operation": "indexed_sum",
        "over": [
          "DistributionCenters"
        ],
        "body": {
          "type": "variable",
          "name": "shipment_dc_to_customer",
          "indices": [
            "j",
            "c",
            "p",
            "t"
          ]
        }
      },
      "sense": "=",
      "rhs": {
        "type": "parameter",
        "name": "demand",
        "indices": [
          "c",
          "p",
          "t"
        ]
      }
    },
    "inventory_balance_site_init": {
      "domain": [
        "ProductionSites",
        "Products"
      ],
      "expression": {
        "operation": "subtract",
        "left": {
          "type": "variable",
          "name": "inventory_site",
          "indices": [
            "i",
            "p",
            "Periods[0]"
          ]
        },
        "right": {
          "operation": "subtract",
          "left": {
            "type": "variable",
            "name": "production_quantity",
            "indices": [
              "i",
              "p",
              "Periods[0]"
            ]
          },
          "right": {
            "operation": "indexed_sum",
            "over": [
              "DistributionCenters"
            ],
            "body": {
              "type": "variable",
              "name": "shipment_site_to_dc",
              "indices": [
                "i",
                "j",
                "p",
                "Periods[0]"
              ]
            }
          }
        }
      },
      "sense": "=",
      "rhs": {
        "type": "constant",
        "value": 0
      }
    },
    "inventory_balance_site": {
      "domain": [
        "ProductionSites",
        "Products",
        "Periods"
      ],
      "expression": {
        "operation": "subtract",
        "left": {
          "type": "variable",
          "name": "inventory_site",
          "indices": [
            "i",
            "p",
            "t"
          ]
        },
        "right": {
          "operation": "sum",
          "left": {
            "type": "variable",
            "name": "inventory_site",
            "indices": [
              "i",
              "p",
              "t"
            ],
            "lag": -1
          },
          "right": {
            "operation": "subtract",
            "left": {
              "type": "variable",
              "name": "production_quantity",
              "indices": [
                "i",
                "p",
                "t"
              ]
            },
            "right": {
              "operation": "indexed_sum",
              "over": [
                "DistributionCenters"
              ],
              "body": {
                "type": "variable",
                "name": "shipment_site_to_dc",
                "indices": [
                  "i",
                  "j",
                  "p",
                  "t"
                ]
              }
            }
          }
        }
      },
      "sense": "=",
      "rhs": {
        "type": "constant",
        "value": 0
      }
    },
    "inventory_balance_dc_init": {
      "domain": [
        "DistributionCenters",
        "Products"
      ],
      "expression": {
        "operation": "subtract",
        "left": {
          "type": "variable",
          "name": "inventory_dc",
          "indices": [
            "j",
            "p",
            "Periods[0]"
          ]
        },
        "right": {
          "operation": "subtract",
          "left": {
            "operation": "indexed_sum",
            "over": [
              "ProductionSites"
            ],
            "body": {
              "type": "variable",
              "name": "shipment_site_to_dc",
              "indices": [
                "i",
                "j",
                "p",
                "Periods[0]"
              ]
            }
          },
          "right": {
            "operation": "indexed_sum",
            "over": [
              "Customers"
            ],
            "body": {
              "type": "variable",
              "name": "shipment_dc_to_customer",
              "indices": [
                "j",
                "c",
                "p",
                "Periods[0]"
              ]
            }
          }
        }
      },
      "sense": "=",
      "rhs": {
        "type": "constant",
        "value": 0
      }
    },
    "inventory_balance_dc": {
      "domain": [
        "DistributionCenters",
        "Products",
        "Periods"
      ],
      "expression": {
        "operation": "subtract",
        "left": {
          "type": "variable",
          "name": "inventory_dc",
          "indices": [
            "j",
            "p",
            "t"
          ]
        },
        "right": {
          "operation": "sum",
          "left": {
            "type": "variable",
            "name": "inventory_dc",
            "indices": [
              "j",
              "p",
              "t"
            ],
            "lag": -1
          },
          "right": {
            "operation": "subtract",
            "left": {
              "operation": "indexed_sum",
              "over": [
                "ProductionSites"
              ],
              "body": {
                "type": "variable",
                "name": "shipment_site_to_dc",
                "indices": [
                  "i",
                  "j",
                  "p",
                  "t"
                ]
              }
            },
            "right": {
              "operation": "indexed_sum",
              "over": [
                "Customers"
              ],
              "body": {
                "type": "variable",
                "name": "shipment_dc_to_customer",
                "indices": [
                  "j",
                  "c",
                  "p",
                  "t"
                ]
              }
            }
          }
        }
      },
      "sense": "=",
      "rhs": {
        "type": "constant",
        "value": 0
      }
    }
  },
  "objective": {
    "sense": "maximize",
    "expression": {
      "operation": "subtract",
      "left": {
        "operation": "subtract",
        "left": {
          "operation": "subtract",
          "left": {
            "operation": "indexed_sum",
            "over": [
              "DistributionCenters",
              "Customers",
              "Products",
              "Periods"
            ],
            "body": {
              "operation": "multiply",
              "left": {
                "type": "parameter",
                "name": "revenue",
                "indices": [
                  "p"
                ]
              },
              "right": {
                "type": "variable",
                "name": "shipment_dc_to_customer",
                "indices": [
                  "j",
                  "c",
                  "p",
                  "t"
                ]
              }
            }
          },
          "right": {
            "operation": "indexed_sum",
            "over": [
              "ProductionSites",
              "Products",
              "Periods"
            ],
            "body": {
              "operation": "multiply",
              "left": {
                "type": "parameter",
                "name": "production_cost",
                "indices": [
                  "i",
                  "p"
                ]
              },
              "right": {
                "type": "variable",
                "name": "production_quantity",
                "indices": [
                  "i",
                  "p",
                  "t"
                ]
              }
            }
          }
        },
        "right": {
          "operation": "sum",
          "left": {
            "operation": "indexed_sum",
            "over": [
              "ProductionSites",
              "DistributionCenters",
              "Products",
              "Periods"
            ],
            "body": {
              "operation": "multiply",
              "left": {
                "type": "parameter",
                "name": "transport_cost_site_to_dc",
                "indices": [
                  "i",
                  "j"
                ]
              },
              "right": {
                "type": "variable",
                "name": "shipment_site_to_dc",
                "indices": [
                  "i",
                  "j",
                  "p",
                  "t"
                ]
              }
            }
          },
          "right": {
            "operation": "indexed_sum",
            "over": [
              "DistributionCenters",
              "Customers",
              "Products",
              "Periods"
            ],
            "body": {
              "operation": "multiply",
              "left": {
                "type": "parameter",
                "name": "transport_cost_dc_to_customer",
                "indices": [
                  "j",
                  "c"
                ]
              },
              "right": {
                "type": "variable",
                "name": "shipment_dc_to_customer",
                "indices": [
                  "j",
                  "c",
                  "p",
                  "t"
                ]
              }
            }
          }
        }
      },
      "right": {
        "operation": "sum",
        "left": {
          "operation": "indexed_sum",
          "over": [
            "ProductionSites",
            "Products",
            "Periods"
          ],
          "body": {
            "operation": "multiply",
            "left": {
              "type": "parameter",
              "name": "holding_cost_site",
              "indices": [
                "i",
                "p"
              ]
            },
            "right": {
              "type": "variable",
              "name": "inventory_site",
              "indices": [
                "i",
                "p",
                "t"
              ]
            }
          }
        },
        "right": {
          "operation": "indexed_sum",
          "over": [
            "DistributionCenters",
            "Products",
            "Periods"
          ],
          "body": {
            "operation": "multiply",
            "left": {
              "type": "parameter",
              "name": "holding_cost_dc",
              "indices": [
                "j",
                "p"
              ]
            },
            "right": {
              "type": "variable",
              "name": "inventory_dc",
              "indices": [
                "j",
                "p",
                "t"
              ]
            }
          }
        }
      }
    }
  }
}
\end{lstlisting}

\section{Conversation History between Interview Agent and User for Problem in Section \ref{sec:MIP}}\label{sec:conv_mip_interview}

% \begin{chatbox}{Interview Agent - User Conversation}{conv8}
\textbf{Agent}: What is the primary goal you want to achieve with this optimization model? For example, are you looking to
  minimize costs, maximize profits, improve efficiency, or something else?\\

\textbf{User}: I'm looking to maximize the profit of a supply chain network.\\

  \textbf{Agent}: What are the main decisions you need the model to make to maximize profit? For example, decisions about how
  much to produce, where to ship products, which suppliers to use, or which facilities to open?\\

  \textbf{User}: I need to make production decisions for production sites, transportation decisions between production sites
  and distribution centers, transportation decisions between distribution centers and customers, and inventory decisions for
  production sites and distribution centers. I also need to make decisions about which production sites and distribution
  centers to open and when to open them.\\

  \textbf{Agent}: Is this a multi-period model, meaning decisions are made over a sequence of time periods (like months or
  years)?\\

  \textbf{User}: Yes, it's a multi-period model with 12 periods. Each period is a month.\\

  \textbf{Agent}: To define the structure clearly, what are the sets or categories involved? For example, do you have a
  specific list of potential production sites, distribution centers, and customer locations that will serve as indices in the
  model?\\

  \textbf{User}: I have a specific list of potential production sites, distribution centers. I have a set of customers which
  have demands for different products to satisfy.\\

  \textbf{Agent}: You mentioned customers have demands for different products. To define the parameters correctly, what is the
  complete list of categories or indices that will be needed? For example, would it be: production sites ($i$), distribution
  centers ($j$), customers ($k$), products ($p$), and time periods ($t$)? Are there any other categories, such as
  transportation modes or raw materials?\\

  \textbf{User}: No.\\

  \textbf{Agent}: What are the main constraints or limitations in your supply chain network that will affect these decisions?
  For example, are there production capacities, storage capacities, transportation capacities, demand fulfillment requirements,
   or logical constraints (like a distribution center must be open to send/receive goods)?\\

  \textbf{User}: There are production capacities for production sites, throughput capacities for distribution centers.
  Inventory storage capacities for production sites and distribution centers. Customers' demands must be met. A production site
   must be open to produce products. A distribution center must be open to ship products to customers.\\

  \textbf{Agent}: What are the key components that determine profit in your model? For instance, is it revenue from sales minus
   production costs, transportation costs, inventory holding costs, and fixed costs for opening and operating facilities?\\

  \textbf{User}: The profit is revenue minus cost. The revenue comes from selling products to customers. The cost includes
  production cost, transportation cost, inventory holding cost and facility opening and operating cost.\\

  \textbf{Agent}: Regarding the facility opening decisions: once a production site or distribution center is opened in a
  period, does it remain open for all subsequent periods, or can it be closed in later periods?\\

  \textbf{User}: Once a production site or distribution center is opened in a period, it remains open for all subsequent
  periods.

% \end{chatbox}

% \begin{chatbox}{Interview Agent - User Conversation Continued (1)}{conv9}

 \textbf{Agent}: Thank you. I will now summarize my understanding of the problem structure. \\
 
 \textbf{Structured Summary}

\begin{enumerate}
\item \textbf{Objective:} 

Maximize total profit over a 12-period (month) planning horizon. Profit is defined as: \textit{Total
  Revenue} minus \textit{Total Cost}.
  \begin{itemize}
      \item Revenue comes from selling products to customers.
      \item Cost includes:
      \begin{itemize}
          \item Production cost.
          \item Transportation cost (from production sites to distribution centers, and from distribution centers to
  customers).
          \item Inventory holding cost (at production sites and distribution centers).
          \item Facility opening and operating costs (for production sites and distribution centers).
      \end{itemize}
  \end{itemize}

  \item \textbf{Decision Variables:} The model will determine, for each product, time period, and relevant location:
  \begin{itemize}
      \item \textbf{Production Quantity:} How much of each product to produce at each production site in each period.
      \item \textbf{Transportation Flow:}
      \begin{itemize}
          \item Quantity shipped from each production site to each distribution center.
          \item Quantity shipped from each distribution center to each customer.
      \end{itemize}
      \item \textbf{Inventory Level:} Amount of each product held in inventory at each production site and each distribution
  center at the end of each period.
      \item \textbf{Facility Opening (Binary):} Whether to open each potential production site and each potential distribution
  center in each period. Once opened, a facility remains open for all future periods.
  \end{itemize}

  \item \textbf{Constraints:}
  \begin{itemize}
      \item \textbf{Production Capacity:} Production at each site in each period cannot exceed its capacity.
      \item \textbf{Throughput Capacity:} Total flow through each distribution center in each period cannot exceed its
  capacity.
      \item \textbf{Storage Capacity:} Inventory held at each production site and distribution center in each period cannot
  exceed its storage capacity.
      \item \textbf{Demand Fulfillment:} For each customer, product, and period, the total quantity shipped from all
  distribution centers must equal the demand.
      \item \textbf{Logical / Linking Constraints:}
      \begin{itemize}
          \item A production site can only produce if it is open in that period.
          \item A distribution center can only receive or ship goods if it is open in that period.
      \end{itemize}
      \item \textbf{Inventory Balance:} For each product at each facility (production site and distribution center), the inflow
   plus previous inventory must equal the outflow plus current inventory in each period.
  \end{itemize}
  % \end{chatbox}

 % \begin{chatbox}{Interview Agent - User Conversation Continued (2)}{conv10}

 \item \textbf{Parameters (with Indices):} The following data will be required, indexed by the relevant categories:
  \begin{itemize}
      \item \textbf{Indices:}
      \begin{itemize}
          \item \texttt{i} $\in$ Set of potential production sites.
          \item \texttt{j} $\in$ Set of potential distribution centers.
          \item \texttt{k} $\in$ Set of potential customers.
          \item \texttt{p} $\in$ Set of products.
          \item \texttt{t} $\in$ Set of time periods (1..12).
      \end{itemize}
      \item \textbf{Parameters:}
      \begin{itemize}
          \item \texttt{demand[k, p, t]}: Demand for product \textit{p} from customer \textit{k} in period \textit{t}.
          \item \texttt{production\_capacity[i, p, t]} (or \texttt{[i, t]} if shared): Capacity for producing product
  \textit{p} at site \textit{i} in period \textit{t}.
          \item \texttt{throughput\_capacity[j, t]}: Handling capacity at distribution center \textit{j} in period \textit{t}.
          \item \texttt{storage\_capacity[i, t]}: Inventory storage capacity at production site \textit{i} in period
  \textit{t}.
          \item \texttt{storage\_capacity[j, t]}: Inventory storage capacity at distribution center \textit{j} in period
  \textit{t}.
          \item \texttt{production\_cost[i, p, t]}: Unit cost to produce product \textit{p} at site \textit{i} in period
  \textit{t}.
          \item \texttt{transport\_cost\_prod\_to\_dc[i, j, p, t]}: Unit cost to ship product \textit{p} from site \textit{i}
  to DC \textit{j} in period \textit{t}.
          \item \texttt{transport\_cost\_dc\_to\_cust[j, k, p, t]}: Unit cost to ship product \textit{p} from DC \textit{j} to
  customer \textit{k} in period \textit{t}.
          \item \texttt{holding\_cost\_site[i, p, t]}: Unit cost to hold one unit of product \textit{p} in inventory at site
  \textit{i} at the end of period \textit{t}.
          \item \texttt{holding\_cost\_dc[j, p, t]}: Unit cost to hold one unit of product \textit{p} in inventory at DC
  \textit{j} at the end of period \textit{t}.
          \item \texttt{fixed\_cost\_open\_site[i, t]}: One-time fixed cost to open production site \textit{i} in period
  \textit{t}.
          \item \texttt{fixed\_cost\_open\_dc[j, t]}: One-time fixed cost to open distribution center \textit{j} in period
  \textit{t}.
          \item \texttt{operating\_cost\_site[i, t]}: Fixed operating cost for production site \textit{i} if it is open in
  period \textit{t}.
          \item \texttt{operating\_cost\_dc[j, t]}: Fixed operating cost for distribution center \textit{j} if it is open in
  period \textit{t}.
          \item \texttt{revenue[p, k, t]}: Unit revenue from selling product \textit{p} to customer \textit{k} in period
  \textit{t}.
      \end{itemize}
  \end{itemize}
  \end{enumerate}

  Is there anything else you'd like to add, or anything I may have missed?\\

  \textbf{User}: The production capacity is shared by all products, so it does not depend on products. The inventory storage
  capacity remains the same for all periods, so it does not depend on periods. The production cost does not depend on periods.
  The transportation cost does not depend on products or periods. The holding cost does not depend on periods. The fixed
  opening and operating cost do not depend on periods. The revenue does not depend on customers or periods. The throughput
  capacity is applied to the outbound flows.\\

  \textbf{Agent}: Thank you for the clarifications. I will now provide a revised structured summary incorporating your latest
  details about parameter dependencies.

  \textbf{Structured Summary (Revised)}
 
  % \end{chatbox}

  % \begin{chatbox}{Interview Agent - User Conversation Continued (3)}{conv11}
\begin{enumerate}
  \item \textbf{Objective:} 
  
  Maximize total profit over a 12-period (month) planning horizon. Profit = Total Revenue $-$ Total Cost.
  \begin{itemize}
      \item \textbf{Revenue:} From selling products to customers.
      \item \textbf{Cost Components:}
      \begin{itemize}
          \item Production cost.
          \item Transportation cost (two legs: production sites $\rightarrow$ distribution centers, and distribution centers
  $\rightarrow$ customers).
          \item Inventory holding cost (at production sites and distribution centers).
          \item Fixed costs for opening and operating facilities.
      \end{itemize}
  \end{itemize}

  \item \textbf{Decision Variables:} 
  
  For each product \texttt{p}, time period \texttt{t} (1..12), and relevant locations:
  \begin{itemize}
      \item \textbf{Production Quantity:} Amount of product \texttt{p} produced at production site \texttt{i} in period
  \texttt{t}.
      \item \textbf{Transportation Flow:}
      \begin{itemize}
          \item Quantity of product \texttt{p} shipped from production site \texttt{i} to distribution center \texttt{j} in
  period \texttt{t}.
          \item Quantity of product \texttt{p} shipped from distribution center \texttt{j} to customer \texttt{k} in period
  \texttt{t}.
      \end{itemize}
      \item \textbf{Inventory Level:} Amount of product \texttt{p} held at the end of period \texttt{t} at production site
  \texttt{i} and at distribution center \texttt{j}.
      \item \textbf{Facility Opening (Binary):} Whether to open production site \texttt{i} in period \texttt{t}, and whether to
   open distribution center \texttt{j} in period \texttt{t}. Once opened, a facility remains open for all subsequent periods.
  \end{itemize}

  \item \textbf{Constraints:}
  \begin{itemize}
      \item \textbf{Production Capacity:} For each site \texttt{i} and period \texttt{t}, total production (across all
  products) $\leq$ \texttt{production\_capacity[i, t]}.
      \item \textbf{Throughput Capacity:} For each distribution center \texttt{j} and period \texttt{t}, total
  \textit{outbound} flow (to all customers, for all products) $\leq$ \texttt{throughput\_capacity[j, t]}.
      \item \textbf{Storage Capacity:} For each production site \texttt{i}, inventory (across all products) at the end of any
  period $\leq$ \texttt{storage\_capacity[i]}. Similarly for each distribution center \texttt{j} with its own
  \texttt{storage\_capacity[j]}.
      \item \textbf{Demand Fulfillment:} For each customer \texttt{k}, product \texttt{p}, and period \texttt{t}, the total
  quantity shipped from all distribution centers must equal \texttt{demand[k, p, t]}.
      \item \textbf{Logical / Linking Constraints:}
      \begin{itemize}
          \item Production at a site \texttt{i} in period \texttt{t} can only be positive if the site is open in that period.
          \item A distribution center \texttt{j} can only have positive inbound or outbound flow in period \texttt{t} if it is
  open in that period.
      \end{itemize}
      \item \textbf{Inventory Balance:} For each product \texttt{p}, facility (site \texttt{i} or DC \texttt{j}), and period
  \texttt{t}: \textit{Inventory($t-1$) + Inflow($t$) = Outflow($t$) + Inventory($t$)}.
  \end{itemize}

% \end{chatbox}

 % \begin{chatbox}{Interview Agent - User Conversation Continued (4)}{conv12}
 \item \textbf{Parameters (with Indices):}
  \begin{itemize}
      \item \textbf{Indices:}
      \begin{itemize}
          \item \texttt{i} $\in$ Set of potential production sites.
          \item \texttt{j} $\in$ Set of potential distribution centers.
          \item \texttt{k} $\in$ Set of customers.
          \item \texttt{p} $\in$ Set of products.
          \item \texttt{t} $\in$ Set of time periods (1..12).
      \end{itemize}
      \item \textbf{Parameters:}
      \begin{itemize}
          \item \texttt{demand[k, p, t]}
          \item \texttt{production\_capacity[i, t]} (shared by all products, time-dependent)
          \item \texttt{throughput\_capacity[j, t]} (applied to total outbound flow, time-dependent)
          \item \texttt{storage\_capacity[i]} (for production sites, time-invariant)
          \item \texttt{storage\_capacity[j]} (for distribution centers, time-invariant)
          \item \texttt{production\_cost[i, p]} (time-invariant)
          \item \texttt{transport\_cost\_prod\_to\_dc[i, j]} (product- and time-invariant)
          \item \texttt{transport\_cost\_dc\_to\_cust[j, k]} (product- and time-invariant)
          \item \texttt{holding\_cost\_site[i, p]} (time-invariant)
          \item \texttt{holding\_cost\_dc[j, p]} (time-invariant)
          \item \texttt{fixed\_cost\_open\_site[i]} (one-time cost, incurred when opened, time-invariant)
          \item \texttt{fixed\_cost\_open\_dc[j]} (one-time cost, incurred when opened, time-invariant)
          \item \texttt{operating\_cost\_site[i]} (incurred each period the site is open, time-invariant per period)
          \item \texttt{operating\_cost\_dc[j]} (incurred each period the DC is open, time-invariant per period)
          \item \texttt{revenue[p]} (per unit, customer- and time-invariant)
      \end{itemize}
  \end{itemize}
  \end{enumerate}

  Is there anything else you'd like to add, or anything I may have missed?\\

  \textbf{User}: No.

 \section{Conversation History between Data Collection Agent and User for Problem in Section \ref{sec:MIP}}\label{sec:conv_mip_data_collection}
 
\textbf{Agent}: Here is the complete specification of the required CSV data files. The system is designed to parse these files and automatically build the optimization model.

\textbf{Required CSV Data Files}

Please create and place the following CSV files. If you do not have data for a particular indexed parameter, you may omit the corresponding file — the system will assume default values as specified for each file.
\begin{enumerate}
\item \texttt{sets.csv}
\begin{itemize}
    \item \textit{Description:} Defines every set member for every category in the model.
    \item \textit{Schema:}
    \begin{itemize}
        \item \texttt{set\_name} (str): The category name. Must be one of: \texttt{production\_sites}, \texttt{distribution\_centers}, \texttt{customers}, \texttt{products}, \texttt{periods}.
        \item \texttt{element} (str): The unique ID for each member.
    \end{itemize}
    \item \textit{Example:}

\begin{verbatim}
set_name,element
production_sites,PS_001
production_sites,PS_002
distribution_centers,DC_001
distribution_centers,DC_002
customers,C_0001
customers,C_0002
products,P_100
products,P_101
periods,1
...
periods,12
\end{verbatim}
\end{itemize}

\item \texttt{demand.csv}
\begin{itemize}
    \item \textit{Description:} Demand for each product from each customer in each period.
    \item \textit{Schema:}
    \begin{itemize}
        \item \texttt{customer\_id} (str): Must match an element with \texttt{set\_name = customers} in \texttt{sets.csv}.
        \item \texttt{product\_id} (str): Must match an element with \texttt{set\_name = products} in \texttt{sets.csv}.
        \item \texttt{period\_id} (int): Must match an element with \texttt{set\_name = periods} in \texttt{sets.csv}.
        \item \texttt{demand} (float): Quantity demanded.
    \end{itemize}
    \item \textit{Default:} Omitted pairs are treated as having zero demand.
\end{itemize}

\item  \texttt{production\_capacity.csv}
\begin{itemize}
    \item \textit{Description:} Production capacity at each production site in each period (shared across all products).
    \item \textit{Schema:}
    \begin{itemize}
        \item \texttt{site\_id} (str): Must match an element with \texttt{set\_name = production\_sites} in \texttt{sets.csv}.
        \item \texttt{period\_id} (int): Must match an element with \texttt{set\_name = periods} in \texttt{sets.csv}.
        \item \texttt{capacity} (float): Maximum total production quantity (all products combined) in that period.
    \end{itemize}
    \item \textit{Default:} Omitted \texttt{(site\_id, period\_id)} pairs are treated as having unlimited capacity.
\end{itemize}

\item \texttt{throughput\_capacity.csv}
\begin{itemize}
    \item \textit{Description:} Throughput capacity for outbound flows from each distribution center in each period.
    \item \textit{Schema:}
    \begin{itemize}
        \item \texttt{dc\_id} (str): Must match an element with \texttt{set\_name = distribution\_centers} in \texttt{sets.csv}.
        \item \texttt{period\_id} (int): Must match an element with \texttt{set\_name = periods} in \texttt{sets.csv}.
        \item \texttt{capacity} (float): Maximum total outbound quantity (all products, all customers combined) in that period.
    \end{itemize}
    \item \textit{Default:} Omitted \texttt{(dc\_id, period\_id)} pairs are treated as having unlimited capacity.
\end{itemize}

\item \texttt{storage\_capacity\_sites.csv}
\begin{itemize}
    \item \textit{Description:} Maximum inventory storage capacity at each production site (time-invariant).
    \item \textit{Schema:}
    \begin{itemize}
        \item \texttt{site\_id} (str): Must match an element with \texttt{set\_name = production\_sites} in \texttt{sets.csv}.
        \item \texttt{capacity} (float): Maximum total inventory (all products combined) allowed at any time.
    \end{itemize}
    \item \textit{Default:} Omitted sites are treated as having unlimited storage capacity.
\end{itemize}

\item \texttt{storage\_capacity\_dcs.csv}
\begin{itemize}
    \item \textit{Description:} Maximum inventory storage capacity at each distribution center (time-invariant).
    \item \textit{Schema:}
    \begin{itemize}
        \item \texttt{dc\_id} (str): Must match an element with \texttt{set\_name = distribution\_centers} in \texttt{sets.csv}.
        \item \texttt{capacity} (float): Maximum total inventory (all products combined) allowed at any time.
    \end{itemize}
    \item \textit{Default:} Omitted DCs are treated as having unlimited storage capacity.
\end{itemize}

\item \texttt{production\_cost.csv}
\begin{itemize}
    \item \textit{Description:} Unit production cost per product at each production site (time-invariant).
    \item \textit{Schema:}
    \begin{itemize}
        \item \texttt{site\_id} (str): Must match an element with \texttt{set\_name = production\_sites} in \texttt{sets.csv}.
        \item \texttt{product\_id} (str): Must match an element with \texttt{set\_name = products} in \texttt{sets.csv}.
        \item \texttt{unit\_cost} (float): Cost to produce one unit of the product at that site.
    \end{itemize}
    \item \textit{Default:} Omitted \texttt{(site\_id, product\_id)} pairs are treated as having infinite cost (production of that product at that site is not allowed).
\end{itemize}

\item \texttt{transport\_cost\_prod\_to\_dc.csv}
\begin{itemize}
    \item \textit{Description:} Unit transportation cost from each production site to each distribution center (product and time-invariant).
    \item \textit{Schema:}
    \begin{itemize}
        \item \texttt{site\_id} (str): Must match an element with \texttt{set\_name = production\_sites} in \texttt{sets.csv}.
        \item \texttt{dc\_id} (str): Must match an element with \texttt{set\_name = distribution\_centers} in \texttt{sets.csv}.
        \item \texttt{unit\_cost} (float): Cost to ship one unit from the production site to the distribution center.
    \end{itemize}
    \item \textit{Default:} Omitted \texttt{(site\_id, dc\_id)} pairs are treated as having infinite cost (that lane is not available).
\end{itemize}

\item \texttt{transport\_cost\_dc\_to\_cust.csv}
\begin{itemize}
    \item \textit{Description:} Unit transportation cost from each distribution center to each customer (product and time-invariant).
    \item \textit{Schema:}
    \begin{itemize}
        \item \texttt{dc\_id} (str): Must match an element with \texttt{set\_name = distribution\_centers} in \texttt{sets.csv}.
        \item \texttt{customer\_id} (str): Must match an element with \texttt{set\_name = customers} in \texttt{sets.csv}.
        \item \texttt{unit\_cost} (float): Cost to ship one unit from the distribution center to the customer.
    \end{itemize}
    \item \textit{Default:} Omitted \texttt{(dc\_id, customer\_id)} pairs are treated as having infinite cost (that lane is not available).
\end{itemize}

\item  \texttt{holding\_cost\_sites.csv}
\begin{itemize}
    \item \textit{Description:} Unit inventory holding cost per product at each production site (time-invariant).
    \item \textit{Schema:}
    \begin{itemize}
        \item \texttt{site\_id} (str): Must match an element with \texttt{set\_name = production\_sites} in \texttt{sets.csv}.
        \item \texttt{product\_id} (str): Must match an element with \texttt{set\_name = products} in \texttt{sets.csv}.
        \item \texttt{unit\_cost} (float): Cost to hold one unit of the product in inventory at the site for one period.
    \end{itemize}
    \item \textit{Default:} Omitted \texttt{(site\_id, product\_id)} pairs are treated as having infinite cost (holding that product at that site is not allowed).
\end{itemize}

\item  \texttt{holding\_cost\_dcs.csv}
\begin{itemize}
    \item \textit{Description:} Unit inventory holding cost per product at each distribution center (time-invariant).
    \item \textit{Schema:}
    \begin{itemize}
        \item \texttt{dc\_id} (str): Must match an element with \texttt{set\_name = distribution\_centers} in \texttt{sets.csv}.
        \item \texttt{product\_id} (str): Must match an element with \texttt{set\_name = products} in \texttt{sets.csv}.
        \item \texttt{unit\_cost} (float): Cost to hold one unit of the product in inventory at the DC for one period.
    \end{itemize}
    \item \textit{Default:} Omitted \texttt{(dc\_id, product\_id)} pairs are treated as having infinite cost (holding that product at that DC is not allowed).
\end{itemize}

\item \texttt{fixed\_cost\_open\_sites.csv}
\begin{itemize}
    \item \textit{Description:} One-time fixed cost incurred when opening each production site (time-invariant).
    \item \textit{Schema:}
    \begin{itemize}
        \item \texttt{site\_id} (str): Must match an element with \texttt{set\_name = production\_sites} in \texttt{sets.csv}.
        \item \texttt{cost} (float): The one-time cost if the site is opened in any period.
    \end{itemize}
    \item \textit{Default:} Omitted sites are treated as having zero opening cost.
\end{itemize}

\item \texttt{fixed\_cost\_open\_dcs.csv}
\begin{itemize}
    \item \textit{Description:} One-time fixed cost incurred when opening each distribution center (time-invariant).
    \item \textit{Schema:}
    \begin{itemize}
        \item \texttt{dc\_id} (str): Must match an element with \texttt{set\_name = distribution\_centers} in \texttt{sets.csv}.
        \item \texttt{cost} (float): The one-time cost if the DC is opened in any period.
    \end{itemize}
    \item \textit{Default:} Omitted DCs are treated as having zero opening cost.
\end{itemize}

\item \texttt{operating\_cost\_sites.csv}
\begin{itemize}
    \item \textit{Description:} Fixed operating cost per period for each production site while it is open (time-invariant per period).
    \item \textit{Schema:}
    \begin{itemize}
        \item \texttt{site\_id} (str): Must match an element with \texttt{set\_name = production\_sites} in \texttt{sets.csv}.
        \item \texttt{cost} (float): The cost incurred in each period the site is open.
    \end{itemize}
    \item \textit{Default:} Omitted sites are treated as having zero operating cost.
\end{itemize}

\item  \texttt{operating\_cost\_dcs.csv}
\begin{itemize}
    \item \textit{Description:} Fixed operating cost per period for each distribution center while it is open (time-invariant per period).
    \item \textit{Schema:}
    \begin{itemize}
        \item \texttt{dc\_id} (str): Must match an element with \texttt{set\_name = distribution\_centers} in \texttt{sets.csv}.
        \item \texttt{cost} (float): The cost incurred in each period the DC is open.
    \end{itemize}
    \item \textit{Default:} Omitted DCs are treated as having zero operating cost.
\end{itemize}

\item  \texttt{revenue.csv}
\begin{itemize}
    \item \textit{Description:} Unit revenue per product (customer and time-invariant).
    \item \textit{Schema:}
    \begin{itemize}
        \item \texttt{product\_id} (str): Must match an element with \texttt{set\_name = products} in \texttt{sets.csv}.
        \item \texttt{unit\_revenue} (float): Revenue earned per unit sold to any customer in any period.
    \end{itemize}
    \item \textit{Default:} Omitted products are treated as having zero revenue.
\end{itemize}

\item \texttt{initial\_inventory\_sites.csv}(Optional)
\begin{itemize}
    \item \textit{Description:} Initial inventory level for each product at each production site at the start of period 1 (before any production or shipments).
    \item \textit{Schema:}
    \begin{itemize}
        \item \texttt{site\_id} (str): Must match an element with \texttt{set\_name = production\_sites} in \texttt{sets.csv}.
        \item \texttt{product\_id} (str): Must match an element with \texttt{set\_name = products} in \texttt{sets.csv}.
        \item \texttt{quantity} (float): The initial inventory quantity.
    \end{itemize}
    \item \textit{Default:} If omitted, the initial inventory is assumed to be 0 for all site–product pairs.
\end{itemize}

\item \texttt{initial\_inventory\_dcs.csv} (Optional)
\begin{itemize}
    \item \textit{Description:} Initial inventory level for each product at each distribution center at the start of period 1 (before any inbound flows or shipments).
    \item \textit{Schema:}
    \begin{itemize}
        \item \texttt{dc\_id} (str): Must match an element with \texttt{set\_name = distribution\_centers} in \texttt{sets.csv}.
        \item \texttt{product\_id} (str): Must match an element with \texttt{set\_name = products} in \texttt{sets.csv}.
        \item \texttt{quantity} (float): The initial inventory quantity.
    \end{itemize}
    \item \textit{Default:} If omitted, the initial inventory is assumed to be 0 for all DC–product pairs.
\end{itemize}
\end{enumerate}

Please place all files in the specified folder and let me know when they are ready.

Please place the CSV files in: \texttt{ORPilot/data}

Type \textbf{ready} when the files are in place.\\

\textbf{User:} Ready.

\section{Generated Solver Code for Problem in Section \ref{sec:MIP}}
\label{sec:code_for_mip}

\begin{lstlisting}[language=Python]
import gurobipy as gp
from gurobipy import GRB

def solve(data: dict[str, list[dict]], time_limit: int | None = None, show_solver_log: bool = False) -> dict:
    SEP = "\x1f"

    # --- Load sets ---
    production_sites = [r["element"] for r in data["sets"] if r["set_name"] == "production_sites"]
    distribution_centers = [r["element"] for r in data["sets"] if r["set_name"] == "distribution_centers"]
    customers = [r["element"] for r in data["sets"] if r["set_name"] == "customers"]
    products = [r["element"] for r in data["sets"] if r["set_name"] == "products"]
    periods = [r["element"] for r in data["sets"] if r["set_name"] == "periods"]

    # --- Load bigM ---
    bigM = float(data["bigM"][0]["bigM"])

    # --- Load parameters ---
    production_cap = {(r["site_id"], str(r["period_id"])): 
            float(r["capacity"]) for r in data["production_capacity"]}
    throughput_cap = {(r["dc_id"], str(r["period_id"])): 
            float(r["capacity"]) for r in data["throughput_capacity"]}
    storage_cap_site = {r["site_id"]: float(r["capacity"]) for r in data["storage_capacity_sites"]}
    storage_cap_dc = {r["dc_id"]: float(r["capacity"]) for r in data["storage_capacity_dcs"]}

    production_cost = {(r["site_id"], r["product_id"]): float(r["unit_cost"]) for r in data["production_cost"]}
    transport_cost_pd = {(r["site_id"], r["dc_id"]): 
            float(r["unit_cost"]) for r in data["transport_cost_prod_to_dc"]}
    transport_cost_dc = {(r["dc_id"], r["customer_id"]): 
            float(r["unit_cost"]) for r in data["transport_cost_dc_to_cust"]}
    holding_cost_site = {(r["site_id"], r["product_id"]): float(r["unit_cost"]) for r in data["holding_cost_sites"]}
    holding_cost_dc = {(r["dc_id"], r["product_id"]): float(r["unit_cost"]) for r in data["holding_cost_dcs"]}

    fixed_cost_site = {r["site_id"]: float(r["cost"]) for r in data["fixed_cost_open_sites"]}
    fixed_cost_dc = {r["dc_id"]: float(r["cost"]) for r in data["fixed_cost_open_dcs"]}
    operating_cost_site = {r["site_id"]: float(r["cost"]) for r in data["operating_cost_sites"]}
    operating_cost_dc = {r["dc_id"]: float(r["cost"]) for r in data["operating_cost_dcs"]}

    revenue = {r["product_id"]: float(r["unit_revenue"]) for r in data["revenue"]}

    demand = {(r["customer_id"], r["product_id"], str(r["period_id"])): float(r["demand"]) for r in data["demand"]}

    init_inv_site = {(r["site_id"], r["product_id"]): 
            float(r["quantity"]) for r in data.get("initial_inventory_sites", [])}
    init_inv_dc = {(r["dc_id"], r["product_id"]): 
            float(r["quantity"]) for r in data.get("initial_inventory_dcs", [])}

    # --- Sparse links ---
    prod_to_dc_links = set((r["site_id"], r["dc_id"]) for r in data["transport_cost_prod_to_dc"])
    dc_to_cust_links = set((r["dc_id"], r["customer_id"]) for r in data["transport_cost_dc_to_cust"])

    # --- Build model ---
    env = gp.Env(empty=True)
    env.setParam("LogToConsole", 1 if show_solver_log else 0)
    env.start()
    m = gp.Model(env=env)
    if time_limit is not None:
        m.Params.TimeLimit = time_limit

    # --- Decision variables ---
    # Production
    prod = {}
    for i in production_sites:
        for p in products:
            if (i, p) in production_cost:
                for t in periods:
                    prod[i, p, t] = m.addVar(lb=0, name=f"prod{SEP}{i}{SEP}{p}{SEP}{t}")

    # Flow prod -> dc
    flow_pd = {}
    for (i, j) in prod_to_dc_links:
        for p in products:
            for t in periods:
                flow_pd[i, j, p, t] = m.addVar(lb=0, name=f"flow_pd{SEP}{i}{SEP}{j}{SEP}{p}{SEP}{t}")

    # Flow dc -> cust
    flow_dc = {}
    for (j, k) in dc_to_cust_links:
        for p in products:
            for t in periods:
                flow_dc[j, k, p, t] = m.addVar(lb=0, name=f"flow_dc{SEP}{j}{SEP}{k}{SEP}{p}{SEP}{t}")

    # Inventory at sites
    inv_site = {}
    for i in production_sites:
        for p in products:
            for t in periods:
                inv_site[i, p, t] = m.addVar(lb=0, name=f"inv_site{SEP}{i}{SEP}{p}{SEP}{t}")

    # Inventory at DCs
    inv_dc = {}
    for j in distribution_centers:
        for p in products:
            for t in periods:
                inv_dc[j, p, t] = m.addVar(lb=0, name=f"inv_dc{SEP}{j}{SEP}{p}{SEP}{t}")

    # Binary open variables
    open_site = {}
    for i in production_sites:
        for t in periods:
            open_site[i, t] = m.addVar(vtype=GRB.BINARY, name=f"open_site{SEP}{i}{SEP}{t}")

    open_dc = {}
    for j in distribution_centers:
        for t in periods:
            open_dc[j, t] = m.addVar(vtype=GRB.BINARY, name=f"open_dc{SEP}{j}{SEP}{t}")

    m.update()

    # --- Objective ---
    obj = gp.LinExpr()

    # Revenue
    for (j, k) in dc_to_cust_links:
        for p in products:
            rev_p = revenue.get(p, 0.0)
            for t in periods:
                obj += rev_p * flow_dc[j, k, p, t]

    # Production cost
    for (i, p, t), var in prod.items():
        obj -= production_cost.get((i, p), 0.0) * var

    # Transport cost prod -> dc
    for (i, j, p, t), var in flow_pd.items():
        obj -= transport_cost_pd.get((i, j), 0.0) * var

    # Transport cost dc -> cust
    for (j, k, p, t), var in flow_dc.items():
        obj -= transport_cost_dc.get((j, k), 0.0) * var

    # Holding cost sites
    for (i, p, t), var in inv_site.items():
        obj -= holding_cost_site.get((i, p), 0.0) * var

    # Holding cost dcs
    for (j, p, t), var in inv_dc.items():
        obj -= holding_cost_dc.get((j, p), 0.0) * var

    # Fixed opening cost (incurred when first opened = first period open_site[i,t]=1)
    # Fixed cost when opened: open_site[i,t] - open_site[i,t-1] = 1
    # Use auxiliary: open_site is monotone, so fixed_cost * open_site[i, first_t] captures opening cost
    # Actually: fixed cost incurred once = fixed_cost * (open_site[i,t] - open_site[i,t-1]) summed over t
    # = fixed_cost * open_site[i, last_period] (because telescoping = open_site at last period since initial=0)
    # But simpler: add fixed_cost * open_site[i,t] contribution for first period minus previous
    # We'll model as: sum_t (open_site[i,t] - open_site[i,t-1]) * fixed_cost
    # which equals fixed_cost * open_site[i, last_t] since open_site[i,0]=0
    # Better to keep it explicit in the loop for correctness with monotone constraint
    sorted_periods = periods  # already in order from sets

    for i in production_sites:
        fc = fixed_cost_site.get(i, 0.0)
        oc = operating_cost_site.get(i, 0.0)
        for idx, t in enumerate(sorted_periods):
            # Operating cost per period open
            obj -= oc * open_site[i, t]
            # Fixed opening cost: incurred when opened (first period with open=1)
            if idx == 0:
                obj -= fc * open_site[i, t]
            else:
                t_prev = sorted_periods[idx - 1]
                obj -= fc * (open_site[i, t] - open_site[i, t_prev])

    for j in distribution_centers:
        fc = fixed_cost_dc.get(j, 0.0)
        oc = operating_cost_dc.get(j, 0.0)
        for idx, t in enumerate(sorted_periods):
            obj -= oc * open_dc[j, t]
            if idx == 0:
                obj -= fc * open_dc[j, t]
            else:
                t_prev = sorted_periods[idx - 1]
                obj -= fc * (open_dc[j, t] - open_dc[j, t_prev])

    m.setObjective(obj, GRB.MAXIMIZE)

    # --- Constraints ---

    # Production capacity
    for i in production_sites:
        for t in periods:
            cap = production_cap.get((i, t), float('inf'))
            prods_it = [prod[i, p, t] for p in products if (i, p, t) in prod]
            if prods_it and cap < float('inf'):
                m.addConstr(gp.quicksum(prods_it) <= cap * open_site[i, t],
                            name=f"prod_cap_{i}_{t}")
            elif prods_it:
                # unlimited capacity but still tied to open
                m.addConstr(gp.quicksum(prods_it) <= bigM * open_site[i, t],
                            name=f"prod_cap_{i}_{t}")

    # Throughput capacity (outbound from DC)
    for j in distribution_centers:
        for t in periods:
            cap = throughput_cap.get((j, t), float('inf'))
            flows_jt = [flow_dc[j, k, p, t] for (jj, k) in dc_to_cust_links if jj == j
                        for p in products if (j, k, p, t) in flow_dc]
            if flows_jt:
                if cap < float('inf'):
                    m.addConstr(gp.quicksum(flows_jt) <= cap * open_dc[j, t],
                                name=f"throughput_{j}_{t}")
                else:
                    m.addConstr(gp.quicksum(flows_jt) <= bigM * open_dc[j, t],
                                name=f"throughput_{j}_{t}")

    # Storage capacity sites
    for i in production_sites:
        cap = storage_cap_site.get(i, float('inf'))
        if cap < float('inf'):
            for t in periods:
                inv_it = [inv_site[i, p, t] for p in products if (i, p, t) in inv_site]
                if inv_it:
                    m.addConstr(gp.quicksum(inv_it) <= cap, name=f"storage_site_{i}_{t}")

    # Storage capacity DCs
    for j in distribution_centers:
        cap = storage_cap_dc.get(j, float('inf'))
        if cap < float('inf'):
            for t in periods:
                inv_jt = [inv_dc[j, p, t] for p in products if (j, p, t) in inv_dc]
                if inv_jt:
                    m.addConstr(gp.quicksum(inv_jt) <= cap, name=f"storage_dc_{j}_{t}")

    # Inbound flow to DC tied to open_dc
    for j in distribution_centers:
        for t in periods:
            flows_in = [flow_pd[i, j, p, t] for (ii, jj) in prod_to_dc_links if jj == j
                        for i in [ii] for p in products if (i, j, p, t) in flow_pd]
            if flows_in:
                m.addConstr(gp.quicksum(flows_in) <= bigM * open_dc[j, t],
                            name=f"dc_open_inflow_{j}_{t}")

    # Demand fulfillment
    for k in customers:
        for p in products:
            for t in periods:
                d = demand.get((k, p, t), 0.0)
                # rebuild properly
                flows = [flow_dc[j, k, p, t] for j in distribution_centers
                         if (j, k) in dc_to_cust_links and (j, k, p, t) in flow_dc]
                m.addConstr(gp.quicksum(flows) == d, name=f"demand_{k}_{p}_{t}")

    # Monotone opening: once open, stays open
    for i in production_sites:
        for idx in range(1, len(sorted_periods)):
            t = sorted_periods[idx]
            t_prev = sorted_periods[idx - 1]
            m.addConstr(open_site[i, t] >= open_site[i, t_prev],
                        name=f"mono_site_{i}_{t}")

    for j in distribution_centers:
        for idx in range(1, len(sorted_periods)):
            t = sorted_periods[idx]
            t_prev = sorted_periods[idx - 1]
            m.addConstr(open_dc[j, t] >= open_dc[j, t_prev],
                        name=f"mono_dc_{j}_{t}")

    # Inventory balance at production sites
    for i in production_sites:
        for p in products:
            for idx, t in enumerate(sorted_periods):
                # inbound: production
                prod_var = prod.get((i, p, t))
                # outbound: flows to DCs
                flows_out = [flow_pd[i, j, p, t] for (ii, j) in prod_to_dc_links
                             if ii == i and (i, j, p, t) in flow_pd]
                # current inventory
                inv_cur = inv_site.get((i, p, t))
                # previous inventory
                if idx == 0:
                    inv_prev_val = init_inv_site.get((i, p), 0.0)
                    lhs = gp.LinExpr(inv_prev_val)
                else:
                    t_prev = sorted_periods[idx - 1]
                    inv_prev = inv_site.get((i, p, t_prev))
                    lhs = gp.LinExpr()
                    if inv_prev is not None:
                        lhs += inv_prev

                if prod_var is not None:
                    lhs += prod_var

                rhs = gp.LinExpr()
                if inv_cur is not None:
                    rhs += inv_cur
                for fv in flows_out:
                    rhs += fv

                m.addConstr(lhs == rhs, name=f"inv_bal_site_{i}_{p}_{t}")

    # Inventory balance at distribution centers
    for j in distribution_centers:
        for p in products:
            for idx, t in enumerate(sorted_periods):
                # inbound flows from production sites
                flows_in = [flow_pd[i, j, p, t] for (i, jj) in prod_to_dc_links
                            if jj == j and (i, j, p, t) in flow_pd]
                # outbound flows to customers
                flows_out = [flow_dc[j, k, p, t] for (jj, k) in dc_to_cust_links
                             if jj == j and (j, k, p, t) in flow_dc]
                # current inventory
                inv_cur = inv_dc.get((j, p, t))
                # previous inventory
                if idx == 0:
                    inv_prev_val = init_inv_dc.get((j, p), 0.0)
                    lhs = gp.LinExpr(inv_prev_val)
                else:
                    t_prev = sorted_periods[idx - 1]
                    inv_prev = inv_dc.get((j, p, t_prev))
                    lhs = gp.LinExpr()
                    if inv_prev is not None:
                        lhs += inv_prev

                for fv in flows_in:
                    lhs += fv

                rhs = gp.LinExpr()
                if inv_cur is not None:
                    rhs += inv_cur
                for fv in flows_out:
                    rhs += fv

                m.addConstr(lhs == rhs, name=f"inv_bal_dc_{j}_{p}_{t}")

    # --- Write and optimize ---
    try:
        m.write("model.lp")
    except Exception:
        pass

    m.optimize()

    _status_map = {
        GRB.OPTIMAL: "optimal",
        GRB.SUBOPTIMAL: "feasible",
        GRB.INFEASIBLE: "infeasible",
        GRB.UNBOUNDED: "unbounded",
    }
    status = _status_map.get(m.Status, "error")
    obj_val = m.ObjVal if m.Status in (GRB.OPTIMAL, GRB.SUBOPTIMAL) else None
    variables = {v.VarName: v.X for v in m.getVars()} if status in ("optimal", "feasible") else {}

    # --- Build variable groups ---
    _groups = {}
    for key, val in variables.items():
        prefix = key.split(SEP, 1)[0] if SEP in key else key
        _groups.setdefault(prefix, {})[key] = val

    _dim_labels = {
        "prod": ["site_id", "product_id", "period_id"],
        "flow_pd": ["site_id", "dc_id", "product_id", "period_id"],
        "flow_dc": ["dc_id", "customer_id", "product_id", "period_id"],
        "inv_site": ["site_id", "product_id", "period_id"],
        "inv_dc": ["dc_id", "product_id", "period_id"],
        "open_site": ["site_id", "period_id"],
        "open_dc": ["dc_id", "period_id"],
    }

    variable_groups = [
        {"group_name": g, "dimension_labels": _dim_labels.get(g, []), "variables": gvars}
        for g, gvars in _groups.items()
    ]

    return {
        "status": status,
        "objective_value": obj_val,
        "variables": variables,
        "variable_groups": variable_groups,
    }
\end{lstlisting}

\section{Solution Report for Problem in Section \ref{sec:MIP}}
\label{sec:report_for_MIP}
\textbf{Multi-period Facility Opening and Flow Optimization}

\begin{enumerate}

\item \textbf{Executive Summary}

The optimization model determined the maximum-profit network design and operating plan for a supply chain spanning \textbf{25 candidate production sites}, \textbf{25 distribution centers}, \textbf{250 customers}, and \textbf{500 products} over \textbf{12 monthly periods}. The solver found a provably optimal solution with a \textbf{total network profit of \$50,051,375.31}.

The model jointly decided which facilities to open and when, how much to produce at each site, and how to route goods to customers — all while accounting for fixed facility opening costs, per-period operating costs, production and transportation costs, and inventory holding costs. All 250 customers are fully served across every product and period.

\item \textbf{Facility Opening Decisions}

\textbf{Production Sites}

The optimizer selected \textbf{13 of the 25 candidate production sites} to open. Twelve sites open from the very first period; one site (PS\_030) is held back until Period 7, representing a deliberate mid-year capacity expansion triggered by rising demand.

\begin{itemize}
    \item \textbf{Sites opened in Period 1 (12 sites):} PS\_013, PS\_014, PS\_017, PS\_020, PS\_021, PS\_022, PS\_024, PS\_029, PS\_035, PS\_036, PS\_042, PS\_050
    \item \textbf{Sites opened in Period 7 (1 site):} PS\_030
    \item \textbf{Sites never opened (12 sites):} PS\_001, PS\_003, PS\_004, PS\_005, PS\_008, PS\_015, PS\_025, PS\_039, PS\_041, PS\_044, PS\_046, PS\_047
\end{itemize}

The 12 idle sites were not economical to open — their fixed opening costs, operating costs, or production costs were too high relative to the additional profit they would generate. Total fixed opening costs for the 13 opened sites amount to approximately \textbf{\$6.3 million}.

\textbf{Distribution Centers}

\textbf{All 25 distribution centers are opened in Period 1.} The optimizer determined it was profitable to activate the entire DC network from the start, paying a combined fixed opening cost of approximately \textbf{\$7.4 million} to ensure broad customer coverage throughout the planning horizon.

\item \textbf{Production Plan}

Total production across all 13 active sites and 12 periods is \textbf{2,717,067 units} — exactly matching total customer demand, confirming zero wasted production and full demand satisfaction.

Production ramps up steadily through the year, peaking in Period 12:

\begin{center}
\begin{tabular}{|c|c|}
\hline
\textbf{Period} & \textbf{Total Production (units)} \\
\hline
1 & 200,947 \\
2 & 193,708 \\
3 & 207,132 \\
4 & 220,270 \\
5 & 223,755 \\
6 & 231,467 \\
7 & 246,178 \\
8 & 241,197 \\
9 & 232,810 \\
10 & 229,023 \\
11 & 228,720 \\
12 & 261,860 \\
\hline
\end{tabular}
\end{center}

The jump in Period 7 (from 231,467 to 246,178) coincides precisely with PS\_030 coming online, confirming that the mid-year opening is driven by a capacity need rather than cost savings alone.

The top five production sites by total 12-month output are:

\begin{center}
\begin{tabular}{|c|c|}
\hline
\textbf{Site} & \textbf{Total Output (units)} \\
\hline
PS\_050 & 400,717 \\
PS\_042 & 379,648 \\
PS\_024 & 263,380 \\
PS\_014 & 260,802 \\
PS\_036 & 233,349 \\
\hline
\end{tabular}
\end{center}

PS\_050 and PS\_042 together account for roughly \textbf{29\% of total network production}.

\item \textbf{Distribution}

Goods flow through \textbf{56 active production-site-to-DC lanes} out of the full set of possible combinations. The top five distribution centers by total outbound volume are:

\begin{center}
\begin{tabular}{|c|c|}
\hline
\textbf{DC} & \textbf{Total Volume (units)} \\
\hline
DC\_024 & 386,112 \\
DC\_004 & 155,628 \\
DC\_028 & 149,629 \\
DC\_025 & 146,516 \\
DC\_017 & 145,764 \\
\hline
\end{tabular}
\end{center}

DC\_024 handles approximately \textbf{14\% of all customer deliveries}, making it the single most critical node in the network. The remaining 24 DCs share the other 86\% of volume.

\item \textbf{Inventory}

Both production sites and distribution centers carry stock across periods, with inventory peaking in the early-to-mid year before being drawn down toward zero by year-end.

\begin{center}
\begin{tabular}{|c|c|c|}
\hline
\textbf{Period} & \textbf{Site Inventory (units)} & \textbf{DC Inventory (units)} \\
\hline
1 & 3,929 & 8,698 \\
2 & 8,841 & 20,100 \\
3 & 12,201 & 24,151 \\
4 & 13,037 & 21,440 \\
5 & 9,593 & 16,034 \\
6 & 6,977 & 6,114 \\
7 & 1,551 & 3,120 \\
8 & 57 & 1,700 \\
9 & 0 & 1,433 \\
10 & 1,909 & 6,280 \\
11 & 10,737 & 15,516 \\
12 & 0 & 0 \\
\hline
\end{tabular}
\end{center}

Site inventory peaks in \textbf{Period 4 (13,037 units)} and DC inventory peaks in \textbf{Period 3 (24,151 units)}. Both decline sharply through mid-year, hitting near-zero by Periods 8–9. A secondary inventory build in Periods 10–11 pre-positions stock to fulfil the elevated Period 12 demand. Zero ending inventory across the entire network in Period 12 confirms the horizon-end drawdown is complete with no unsold stock.

\item \textbf{Notable Findings and Concerns}

\begin{itemize}
    \item \textbf{12 idle production sites}: Half of the candidate production network remains unopened. Before dismissing these sites, consider whether they would be activated under a demand increase scenario or if any of the opened sites face a disruption. They represent latent capacity that could be critical in contingency planning.
    \item \textbf{PS\_030 mid-year activation}: The optimizer chose to defer opening PS\_030 until Period 7 rather than open it in Period 1. This saves six months of fixed and operating costs. However, it means the network is running at tighter production capacity in the first half of the year. A sensitivity run could test whether moving the opening to Period 5 or 6 improves profit by reducing mid-year inventory build-up.
    \item \textbf{DC\_024 concentration risk}: With 14\% of all deliveries flowing through a single DC, any disruption to DC\_024 would affect a large portion of the customer base. A scenario run capping DC\_024's throughput to 80\% of current utilization would reveal the profit impact and identify which alternative DCs would absorb the overflow.
    \item \textbf{Full DC network activation}: Opening all 25 DCs from Period 1 incurs \textasciitilde\$7.4M in fixed costs immediately. If some DCs serve only a small number of customers, a variant model allowing DCs to open later (as with PS\_030) might reduce early-period fixed cost outlay. The current formulation constrains DCs to open no later than the first period they handle flow — examining DC utilization rates would identify candidates for deferred opening.
    \item \textbf{End-of-horizon inventory drawdown}: Zero inventory at the end of Period 12 is optimal within the model but leaves no buffer if demand continues into Period 13. A rolling-horizon plan or a terminal inventory credit would produce more conservative end-state stock levels.
    \item \textbf{Facility cost significance}: Combined fixed opening costs for sites and DCs total approximately \textbf{\$13.7 million} — roughly 27\% of the total net profit. Accurately estimating these one-time costs in the data is therefore critical; a 10\% error in fixed costs shifts the reported profit by \textasciitilde\$1.4M.
\end{itemize}

\item \textbf{Suggested Next Steps}

\begin{itemize}
    \item \textbf{Deferred DC opening analysis}: Relax the assumption that all DCs must open in Period 1. Allow DCs to open in any period and re-solve to test whether later openings for low-utilization DCs reduce total cost.
    \item \textbf{PS\_030 opening timing sensitivity}: Test opening PS\_030 in Periods 5 and 6 to see if earlier activation reduces the inventory build-up in Periods 1–4 and improves overall profit.
    \item \textbf{DC\_024 stress test}: Cap DC\_024 throughput at 80\% and re-solve to quantify concentration risk and identify which alternative DCs absorb the displaced volume.
    \item \textbf{Idle site contingency value}: Estimate the option value of the 12 idle sites by solving a disruption scenario in which one or two opened sites become unavailable mid-horizon.
    \item \textbf{LP relaxation comparison}: Solving the LP relaxation (removing binary constraints) gives an upper bound on profit and reveals how much value is lost to the integrality requirement. A large gap suggests room to improve the formulation.
    \item \textbf{Extend to a rolling horizon}: Incorporate Period 13+ to eliminate end-of-horizon effects and generate a steady-state operating policy.
\end{itemize}

\item \textbf{Output Files}
\begin{itemize}
\item \texttt{solution\_prod.csv} \\
Records how many units of each product were produced at each production site in each period. Only sites that were opened have non-zero production. Use this to schedule factory workloads month by month. \\
Columns: \texttt{site\_id} (which production site), \texttt{product\_id} (which product), \texttt{period\_id} (month 1–12), \texttt{value} (units produced).

\item \texttt{solution\_flow\_pd.csv}\\
Records the volume of each product shipped from each production site to each distribution center in each period. Use this to plan outbound factory logistics — freight bookings and DC receiving schedules. \\
Columns: \texttt{site\_id} (origin site), \texttt{dc\_id} (destination DC), \texttt{product\_id}, \texttt{period\_id}, \texttt{value} (units shipped).

\item \texttt{solution\_flow\_dc.csv} \\
Records the volume of each product delivered from each DC to each customer in each period. This is the last-mile fulfillment plan — use it to generate customer delivery orders and communicate committed quantities. \\
Columns: \texttt{dc\_id} (origin DC), \texttt{customer\_id} (destination customer), \texttt{product\_id}, \texttt{period\_id}, \texttt{value} (units delivered).

\item \texttt{solution\_inv\_site.csv} \\
Records the end-of-period inventory of each product held at each production site. Use this to manage warehouse space at factories and monitor storage capacity utilization, particularly in Periods 3–4 when site stock is highest. \\
Columns: \texttt{site\_id}, \texttt{product\_id}, \texttt{period\_id}, \texttt{value} (units in stock at end of period).

\item \texttt{solution\_inv\_dc.csv} \\
Records the end-of-period inventory of each product held at each distribution center. Use this to manage DC warehouse space and plan receiving and putaway workloads — DC inventory peaks in Period 3. \\
Columns: \texttt{dc\_id}, \texttt{product\_id}, \texttt{period\_id}, \texttt{value} (units in stock at end of period).

\item \texttt{solution\_open\_site.csv} \\
Records the binary open/closed status of each production site in each period (1 = open, 0 = closed). Once a site opens it remains open for all subsequent periods (monotone constraint). Use this to plan staffing, utilities, and facility activation timelines. \\
Columns: \texttt{site\_id}, \texttt{period\_id}, \texttt{value} (1 if open, 0 if closed).

\item \texttt{solution\_open\_dc.csv} \\
Records the binary open/closed status of each distribution center in each period. All 25 DCs are open from Period 1 onward. Use this to confirm DC activation schedules and cross-reference with lease and staffing plans. \\
Columns: \texttt{dc\_id}, \texttt{period\_id}, \texttt{value} (1 if open, 0 if closed).
\end{itemize}
\end{enumerate}

\section{Generated IR for Problem in Section \ref{sec:MIP}}
\label{sec:ir_MIP}
\begin{lstlisting}[language=Python]
{
  "problem_class": "SupplyChainNetwork",
  "model_type": "Mixed Integer Program",
  "sense": "maximize",
  "sets": {
    "ProductionSites": {
      "size": null,
      "index_symbol": "i",
      "source": "sets.csv",
      "column": "element",
      "filter_column": "set_name",
      "filter_value": "production_sites",
      "ordered": false
    },
    "DistributionCenters": {
      "size": null,
      "index_symbol": "j",
      "source": "sets.csv",
      "column": "element",
      "filter_column": "set_name",
      "filter_value": "distribution_centers",
      "ordered": false
    },
    "Customers": {
      "size": null,
      "index_symbol": "k",
      "source": "sets.csv",
      "column": "element",
      "filter_column": "set_name",
      "filter_value": "customers",
      "ordered": false
    },
    "Products": {
      "size": null,
      "index_symbol": "p",
      "source": "sets.csv",
      "column": "element",
      "filter_column": "set_name",
      "filter_value": "products",
      "ordered": false
    },
    "Periods": {
      "size": null,
      "index_symbol": "t",
      "source": "sets.csv",
      "column": "element",
      "filter_column": "set_name",
      "filter_value": "periods",
      "ordered": true
    }
  },
  "parameters": {
    "production_capacity": {
      "domain": [
        "ProductionSites",
        "Periods"
      ],
      "type": "float",
      "source": "production_capacity.csv",
      "column": "capacity",
      "index_columns": [
        "site_id",
        "period_id"
      ],
      "missing_default": "inf"
    },
    "throughput_capacity": {
      "domain": [
        "DistributionCenters",
        "Periods"
      ],
      "type": "float",
      "source": "throughput_capacity.csv",
      "column": "capacity",
      "index_columns": [
        "dc_id",
        "period_id"
      ],
      "missing_default": "inf"
    },
    "storage_capacity_site": {
      "domain": [
        "ProductionSites"
      ],
      "type": "float",
      "source": "storage_capacity_sites.csv",
      "column": "capacity",
      "index_columns": [
        "site_id"
      ],
      "missing_default": "inf"
    },
    "storage_capacity_dc": {
      "domain": [
        "DistributionCenters"
      ],
      "type": "float",
      "source": "storage_capacity_dcs.csv",
      "column": "capacity",
      "index_columns": [
        "dc_id"
      ],
      "missing_default": "inf"
    },
    "production_cost": {
      "domain": [
        "ProductionSites",
        "Products"
      ],
      "type": "float",
      "source": "production_cost.csv",
      "column": "unit_cost",
      "index_columns": [
        "site_id",
        "product_id"
      ],
      "missing_default": "inf"
    },
    "transport_cost_prod_to_dc": {
      "domain": [
        "ProductionSites",
        "DistributionCenters"
      ],
      "type": "float",
      "source": "transport_cost_prod_to_dc.csv",
      "column": "unit_cost",
      "index_columns": [
        "site_id",
        "dc_id"
      ],
      "missing_default": "inf"
    },
    "transport_cost_dc_to_cust": {
      "domain": [
        "DistributionCenters",
        "Customers"
      ],
      "type": "float",
      "source": "transport_cost_dc_to_cust.csv",
      "column": "unit_cost",
      "index_columns": [
        "dc_id",
        "customer_id"
      ],
      "missing_default": "inf"
    },
    "holding_cost_site": {
      "domain": [
        "ProductionSites",
        "Products"
      ],
      "type": "float",
      "source": "holding_cost_sites.csv",
      "column": "unit_cost",
      "index_columns": [
        "site_id",
        "product_id"
      ],
      "missing_default": "inf"
    },
    "holding_cost_dc": {
      "domain": [
        "DistributionCenters",
        "Products"
      ],
      "type": "float",
      "source": "holding_cost_dcs.csv",
      "column": "unit_cost",
      "index_columns": [
        "dc_id",
        "product_id"
      ],
      "missing_default": "inf"
    },
    "fixed_cost_open_site": {
      "domain": [
        "ProductionSites"
      ],
      "type": "float",
      "source": "fixed_cost_open_sites.csv",
      "column": "cost",
      "index_columns": [
        "site_id"
      ],
      "missing_default": "inf"
    },
    "fixed_cost_open_dc": {
      "domain": [
        "DistributionCenters"
      ],
      "type": "float",
      "source": "fixed_cost_open_dcs.csv",
      "column": "cost",
      "index_columns": [
        "dc_id"
      ],
      "missing_default": "inf"
    },
    "operating_cost_site": {
      "domain": [
        "ProductionSites"
      ],
      "type": "float",
      "source": "operating_cost_sites.csv",
      "column": "cost",
      "index_columns": [
        "site_id"
      ],
      "missing_default": "inf"
    },
    "operating_cost_dc": {
      "domain": [
        "DistributionCenters"
      ],
      "type": "float",
      "source": "operating_cost_dcs.csv",
      "column": "cost",
      "index_columns": [
        "dc_id"
      ],
      "missing_default": "inf"
    },
    "revenue": {
      "domain": [
        "Products"
      ],
      "type": "float",
      "source": "revenue.csv",
      "column": "unit_revenue",
      "index_columns": [
        "product_id"
      ],
      "missing_default": "zero"
    },
    "demand": {
      "domain": [
        "Customers",
        "Products",
        "Periods"
      ],
      "type": "float",
      "source": "demand.csv",
      "column": "demand",
      "index_columns": [
        "customer_id",
        "product_id",
        "period_id"
      ],
      "missing_default": "zero"
    },
    "initial_inventory_site": {
      "domain": [
        "ProductionSites",
        "Products"
      ],
      "type": "float",
      "source": "initial_inventory_sites.csv",
      "column": "quantity",
      "index_columns": [
        "site_id",
        "product_id"
      ],
      "missing_default": "zero",
      "optional": true
    },
    "initial_inventory_dc": {
      "domain": [
        "DistributionCenters",
        "Products"
      ],
      "type": "float",
      "source": "initial_inventory_dcs.csv",
      "column": "quantity",
      "index_columns": [
        "dc_id",
        "product_id"
      ],
      "missing_default": "zero",
      "optional": true
    },
    "bigM": {
      "domain": [],
      "type": "float",
      "source": "bigM.csv",
      "column": "bigM",
      "index_columns": null,
      "missing_default": "inf"
    }
  },
  "variables": {
    "prod": {
      "description": "Production quantity of product p at production site i in period t",
      "label": "production_quantities",
      "domain": [
        "ProductionSites",
        "Products",
        "Periods"
      ],
      "type": "continuous",
      "lower_bound": 0,
      "upper_bound": null,
      "domain_filter": "production_cost"
    },
    "flow_prod_to_dc": {
      "description": "Quantity shipped from production site i to distribution center j of product p in period t",
      "label": "prod_to_dc_flows",
      "domain": [
        "ProductionSites",
        "DistributionCenters",
        "Products",
        "Periods"
      ],
      "type": "continuous",
      "lower_bound": 0,
      "upper_bound": null,
      "domain_filter": "transport_cost_prod_to_dc"
    },
    "flow_dc_to_cust": {
      "description": "Quantity shipped from distribution center j to customer k of product p in period t",
      "label": "dc_to_customer_flows",
      "domain": [
        "DistributionCenters",
        "Customers",
        "Products",
        "Periods"
      ],
      "type": "continuous",
      "lower_bound": 0,
      "upper_bound": null,
      "domain_filter": "transport_cost_dc_to_cust"
    },
    "inv_site": {
      "description": "Inventory of product p at production site i at end of period t",
      "label": "site_inventory",
      "domain": [
        "ProductionSites",
        "Products",
        "Periods"
      ],
      "type": "continuous",
      "lower_bound": 0,
      "upper_bound": null
    },
    "inv_dc": {
      "description": "Inventory of product p at distribution center j at end of period t",
      "label": "dc_inventory",
      "domain": [
        "DistributionCenters",
        "Products",
        "Periods"
      ],
      "type": "continuous",
      "lower_bound": 0,
      "upper_bound": null
    },
    "open_site": {
      "description": "Binary indicating if production site i is open in period t",
      "label": "site_open_status",
      "domain": [
        "ProductionSites",
        "Periods"
      ],
      "type": "binary",
      "lower_bound": 0,
      "upper_bound": 1
    },
    "open_dc": {
      "description": "Binary indicating if distribution center j is open in period t",
      "label": "dc_open_status",
      "domain": [
        "DistributionCenters",
        "Periods"
      ],
      "type": "binary",
      "lower_bound": 0,
      "upper_bound": 1
    }
  },
  "constraints": {
    "prod_capacity": {
      "domain": [
        "ProductionSites",
        "Periods"
      ],
      "expression": {
        "operation": "indexed_sum",
        "over": [
          "Products"
        ],
        "body": {
          "type": "variable",
          "name": "prod",
          "indices": [
            "i",
            "p",
            "t"
          ]
        }
      },
      "sense": "<=",
      "rhs": {
        "type": "parameter",
        "name": "production_capacity",
        "indices": [
          "i",
          "t"
        ]
      }
    },
    "prod_bigM": {
      "domain": [
        "ProductionSites",
        "Periods"
      ],
      "expression": {
        "operation": "indexed_sum",
        "over": [
          "Products"
        ],
        "body": {
          "type": "variable",
          "name": "prod",
          "indices": [
            "i",
            "p",
            "t"
          ]
        }
      },
      "sense": "<=",
      "rhs": {
        "operation": "multiply",
        "left": {
          "type": "parameter",
          "name": "bigM",
          "indices": []
        },
        "right": {
          "type": "variable",
          "name": "open_site",
          "indices": [
            "i",
            "t"
          ]
        }
      }
    },
    "throughput_capacity": {
      "domain": [
        "DistributionCenters",
        "Periods"
      ],
      "expression": {
        "operation": "indexed_sum",
        "over": [
          "Customers",
          "Products"
        ],
        "body": {
          "type": "variable",
          "name": "flow_dc_to_cust",
          "indices": [
            "j",
            "k",
            "p",
            "t"
          ]
        }
      },
      "sense": "<=",
      "rhs": {
        "type": "parameter",
        "name": "throughput_capacity",
        "indices": [
          "j",
          "t"
        ]
      }
    },
    "dc_open_inflow": {
      "domain": [
        "DistributionCenters",
        "Periods"
      ],
      "expression": {
        "operation": "indexed_sum",
        "over": [
          "ProductionSites",
          "Products"
        ],
        "body": {
          "type": "variable",
          "name": "flow_prod_to_dc",
          "indices": [
            "i",
            "j",
            "p",
            "t"
          ]
        }
      },
      "sense": "<=",
      "rhs": {
        "operation": "multiply",
        "left": {
          "type": "parameter",
          "name": "bigM",
          "indices": []
        },
        "right": {
          "type": "variable",
          "name": "open_dc",
          "indices": [
            "j",
            "t"
          ]
        }
      }
    },
    "dc_open_outflow": {
      "domain": [
        "DistributionCenters",
        "Periods"
      ],
      "expression": {
        "operation": "indexed_sum",
        "over": [
          "Customers",
          "Products"
        ],
        "body": {
          "type": "variable",
          "name": "flow_dc_to_cust",
          "indices": [
            "j",
            "k",
            "p",
            "t"
          ]
        }
      },
      "sense": "<=",
      "rhs": {
        "operation": "multiply",
        "left": {
          "type": "parameter",
          "name": "bigM",
          "indices": []
        },
        "right": {
          "type": "variable",
          "name": "open_dc",
          "indices": [
            "j",
            "t"
          ]
        }
      }
    },
    "storage_capacity_site": {
      "domain": [
        "ProductionSites",
        "Periods"
      ],
      "expression": {
        "operation": "indexed_sum",
        "over": [
          "Products"
        ],
        "body": {
          "type": "variable",
          "name": "inv_site",
          "indices": [
            "i",
            "p",
            "t"
          ]
        }
      },
      "sense": "<=",
      "rhs": {
        "type": "parameter",
        "name": "storage_capacity_site",
        "indices": [
          "i"
        ]
      }
    },
    "storage_capacity_dc": {
      "domain": [
        "DistributionCenters",
        "Periods"
      ],
      "expression": {
        "operation": "indexed_sum",
        "over": [
          "Products"
        ],
        "body": {
          "type": "variable",
          "name": "inv_dc",
          "indices": [
            "j",
            "p",
            "t"
          ]
        }
      },
      "sense": "<=",
      "rhs": {
        "type": "parameter",
        "name": "storage_capacity_dc",
        "indices": [
          "j"
        ]
      }
    },
    "demand_satisfaction": {
      "domain": [
        "Customers",
        "Products",
        "Periods"
      ],
      "expression": {
        "operation": "indexed_sum",
        "over": [
          "DistributionCenters"
        ],
        "body": {
          "type": "variable",
          "name": "flow_dc_to_cust",
          "indices": [
            "j",
            "k",
            "p",
            "t"
          ]
        }
      },
      "sense": "=",
      "rhs": {
        "type": "parameter",
        "name": "demand",
        "indices": [
          "k",
          "p",
          "t"
        ]
      }
    },
    "monotone_site": {
      "domain": [
        "ProductionSites",
        "Periods"
      ],
      "expression": {
        "operation": "subtract",
        "left": {
          "type": "variable",
          "name": "open_site",
          "indices": [
            "i",
            "t"
          ]
        },
        "right": {
          "type": "variable",
          "name": "open_site",
          "indices": [
            "i",
            "t"
          ],
          "lag": -1
        }
      },
      "sense": ">=",
      "rhs": {
        "type": "constant",
        "value": 0
      }
    },
    "monotone_dc": {
      "domain": [
        "DistributionCenters",
        "Periods"
      ],
      "expression": {
        "operation": "subtract",
        "left": {
          "type": "variable",
          "name": "open_dc",
          "indices": [
            "j",
            "t"
          ]
        },
        "right": {
          "type": "variable",
          "name": "open_dc",
          "indices": [
            "j",
            "t"
          ],
          "lag": -1
        }
      },
      "sense": ">=",
      "rhs": {
        "type": "constant",
        "value": 0
      }
    },
    "inv_balance_site_follow_init": {
      "domain": [
        "ProductionSites",
        "Products"
      ],
      "expression": {
        "operation": "subtract",
        "left": {
          "operation": "sum",
          "left": {
            "type": "variable",
            "name": "inv_site",
            "indices": [
              "i",
              "p",
              "Periods[0]"
            ]
          },
          "right": {
            "operation": "indexed_sum",
            "over": [
              "DistributionCenters"
            ],
            "body": {
              "type": "variable",
              "name": "flow_prod_to_dc",
              "indices": [
                "i",
                "j",
                "p",
                "Periods[0]"
              ]
            }
          }
        },
        "right": {
          "operation": "sum",
          "left": {
            "type": "parameter",
            "name": "initial_inventory_site",
            "indices": [
              "i",
              "p"
            ]
          },
          "right": {
            "type": "variable",
            "name": "prod",
            "indices": [
              "i",
              "p",
              "Periods[0]"
            ]
          }
        }
      },
      "sense": "=",
      "rhs": {
        "type": "constant",
        "value": 0
      }
    },
    "inv_balance_site_follow": {
      "domain": [
        "ProductionSites",
        "Products",
        "Periods"
      ],
      "expression": {
        "operation": "subtract",
        "left": {
          "operation": "sum",
          "left": {
            "type": "variable",
            "name": "inv_site",
            "indices": [
              "i",
              "p",
              "t"
            ]
          },
          "right": {
            "operation": "indexed_sum",
            "over": [
              "DistributionCenters"
            ],
            "body": {
              "type": "variable",
              "name": "flow_prod_to_dc",
              "indices": [
                "i",
                "j",
                "p",
                "t"
              ]
            }
          }
        },
        "right": {
          "operation": "sum",
          "left": {
            "type": "variable",
            "name": "inv_site",
            "indices": [
              "i",
              "p",
              "t"
            ],
            "lag": -1
          },
          "right": {
            "type": "variable",
            "name": "prod",
            "indices": [
              "i",
              "p",
              "t"
            ]
          }
        }
      },
      "sense": "=",
      "rhs": {
        "type": "constant",
        "value": 0
      }
    },
    "inv_balance_dc_follow_init": {
      "domain": [
        "DistributionCenters",
        "Products"
      ],
      "expression": {
        "operation": "subtract",
        "left": {
          "operation": "sum",
          "left": {
            "type": "variable",
            "name": "inv_dc",
            "indices": [
              "j",
              "p",
              "Periods[0]"
            ]
          },
          "right": {
            "operation": "indexed_sum",
            "over": [
              "Customers"
            ],
            "body": {
              "type": "variable",
              "name": "flow_dc_to_cust",
              "indices": [
                "j",
                "k",
                "p",
                "Periods[0]"
              ]
            }
          }
        },
        "right": {
          "operation": "sum",
          "left": {
            "type": "parameter",
            "name": "initial_inventory_dc",
            "indices": [
              "j",
              "p"
            ]
          },
          "right": {
            "operation": "indexed_sum",
            "over": [
              "ProductionSites"
            ],
            "body": {
              "type": "variable",
              "name": "flow_prod_to_dc",
              "indices": [
                "i",
                "j",
                "p",
                "Periods[0]"
              ]
            }
          }
        }
      },
      "sense": "=",
      "rhs": {
        "type": "constant",
        "value": 0
      }
    },
    "inv_balance_dc_follow": {
      "domain": [
        "DistributionCenters",
        "Products",
        "Periods"
      ],
      "expression": {
        "operation": "subtract",
        "left": {
          "operation": "sum",
          "left": {
            "type": "variable",
            "name": "inv_dc",
            "indices": [
              "j",
              "p",
              "t"
            ]
          },
          "right": {
            "operation": "indexed_sum",
            "over": [
              "Customers"
            ],
            "body": {
              "type": "variable",
              "name": "flow_dc_to_cust",
              "indices": [
                "j",
                "k",
                "p",
                "t"
              ]
            }
          }
        },
        "right": {
          "operation": "sum",
          "left": {
            "type": "variable",
            "name": "inv_dc",
            "indices": [
              "j",
              "p",
              "t"
            ],
            "lag": -1
          },
          "right": {
            "operation": "indexed_sum",
            "over": [
              "ProductionSites"
            ],
            "body": {
              "type": "variable",
              "name": "flow_prod_to_dc",
              "indices": [
                "i",
                "j",
                "p",
                "t"
              ]
            }
          }
        }
      },
      "sense": "=",
      "rhs": {
        "type": "constant",
        "value": 0
      }
    }
  },
  "objective": {
    "sense": "maximize",
    "expression": {
      "operation": "subtract",
      "left": {
        "operation": "subtract",
        "left": {
          "operation": "subtract",
          "left": {
            "operation": "subtract",
            "left": {
              "operation": "subtract",
              "left": {
                "operation": "subtract",
                "left": {
                  "operation": "indexed_sum",
                  "over": [
                    "DistributionCenters",
                    "Customers",
                    "Products",
                    "Periods"
                  ],
                  "body": {
                    "operation": "multiply",
                    "left": {
                      "type": "parameter",
                      "name": "revenue",
                      "indices": [
                        "p"
                      ]
                    },
                    "right": {
                      "type": "variable",
                      "name": "flow_dc_to_cust",
                      "indices": [
                        "j",
                        "k",
                        "p",
                        "t"
                      ]
                    }
                  }
                },
                "right": {
                  "operation": "indexed_sum",
                  "over": [
                    "ProductionSites",
                    "Products",
                    "Periods"
                  ],
                  "body": {
                    "operation": "multiply",
                    "left": {
                      "type": "parameter",
                      "name": "production_cost",
                      "indices": [
                        "i",
                        "p"
                      ]
                    },
                    "right": {
                      "type": "variable",
                      "name": "prod",
                      "indices": [
                        "i",
                        "p",
                        "t"
                      ]
                    }
                  }
                }
              },
              "right": {
                "operation": "indexed_sum",
                "over": [
                  "ProductionSites",
                  "DistributionCenters",
                  "Products",
                  "Periods"
                ],
                "body": {
                  "operation": "multiply",
                  "left": {
                    "type": "parameter",
                    "name": "transport_cost_prod_to_dc",
                    "indices": [
                      "i",
                      "j"
                    ]
                  },
                  "right": {
                    "type": "variable",
                    "name": "flow_prod_to_dc",
                    "indices": [
                      "i",
                      "j",
                      "p",
                      "t"
                    ]
                  }
                }
              }
            },
            "right": {
              "operation": "indexed_sum",
              "over": [
                "DistributionCenters",
                "Customers",
                "Products",
                "Periods"
              ],
              "body": {
                "operation": "multiply",
                "left": {
                  "type": "parameter",
                  "name": "transport_cost_dc_to_cust",
                  "indices": [
                    "j",
                    "k"
                  ]
                },
                "right": {
                  "type": "variable",
                  "name": "flow_dc_to_cust",
                  "indices": [
                    "j",
                    "k",
                    "p",
                    "t"
                  ]
                }
              }
            }
          },
          "right": {
            "operation": "indexed_sum",
            "over": [
              "ProductionSites",
              "Products",
              "Periods"
            ],
            "body": {
              "operation": "multiply",
              "left": {
                "type": "parameter",
                "name": "holding_cost_site",
                "indices": [
                  "i",
                  "p"
                ]
              },
              "right": {
                "type": "variable",
                "name": "inv_site",
                "indices": [
                  "i",
                  "p",
                  "t"
                ]
              }
            }
          }
        },
        "right": {
          "operation": "indexed_sum",
          "over": [
            "DistributionCenters",
            "Products",
            "Periods"
          ],
          "body": {
            "operation": "multiply",
            "left": {
              "type": "parameter",
              "name": "holding_cost_dc",
              "indices": [
                "j",
                "p"
              ]
            },
            "right": {
              "type": "variable",
              "name": "inv_dc",
              "indices": [
                "j",
                "p",
                "t"
              ]
            }
          }
        }
      },
      "right": {
        "operation": "sum",
        "left": {
          "operation": "sum",
          "left": {
            "operation": "indexed_sum",
            "over": [
              "ProductionSites"
            ],
            "body": {
              "operation": "sum",
              "left": {
                "operation": "multiply",
                "left": {
                  "type": "parameter",
                  "name": "fixed_cost_open_site",
                  "indices": [
                    "i"
                  ]
                },
                "right": {
                  "type": "variable",
                  "name": "open_site",
                  "indices": [
                    "i",
                    "Periods[-1]"
                  ]
                }
              },
              "right": {
                "operation": "indexed_sum",
                "over": [
                  "Periods"
                ],
                "body": {
                  "operation": "multiply",
                  "left": {
                    "type": "parameter",
                    "name": "operating_cost_site",
                    "indices": [
                      "i"
                    ]
                  },
                  "right": {
                    "type": "variable",
                    "name": "open_site",
                    "indices": [
                      "i",
                      "t"
                    ]
                  }
                }
              }
            }
          },
          "right": {
            "operation": "indexed_sum",
            "over": [
              "DistributionCenters"
            ],
            "body": {
              "operation": "sum",
              "left": {
                "operation": "multiply",
                "left": {
                  "type": "parameter",
                  "name": "fixed_cost_open_dc",
                  "indices": [
                    "j"
                  ]
                },
                "right": {
                  "type": "variable",
                  "name": "open_dc",
                  "indices": [
                    "j",
                    "Periods[-1]"
                  ]
                }
              },
              "right": {
                "operation": "indexed_sum",
                "over": [
                  "Periods"
                ],
                "body": {
                  "operation": "multiply",
                  "left": {
                    "type": "parameter",
                    "name": "operating_cost_dc",
                    "indices": [
                      "j"
                    ]
                  },
                  "right": {
                    "type": "variable",
                    "name": "open_dc",
                    "indices": [
                      "j",
                      "t"
                    ]
                  }
                }
              }
            }
          }
        },
        "right": {
          "type": "constant",
          "value": 0
        }
      }
    }
  }
}
\end{lstlisting}

\section{Conversation History between Interview Agent and User for Problem in Section \ref{sec:freight_assignment}}
\label{sec:conv_freight_assignment_interview}

\textbf{Agent:} Thank you for reaching out! I'd love to help you build a mathematical optimization model for your business problem. To get started, could you briefly describe the business problem you're trying to solve or the decision you're trying to optimize?"\\

\textbf{User:} I have a number of shipments and a number of carriers. I want to assign the shipments to carriers so that I can maximize the profit. \\

\textbf{Agent:} Great. That's a classic assignment problem. To understand the structure, I need to ask about the nature of the profit. Is the profit for assigning a specific shipment to a specific carrier a fixed amount that you can determine in advance, or does it depend on other factors like the carrier's capacity or the total number of shipments they receive?"\\
   
\textbf{User:} The revenue for a shipment is a fixed amount that depends on only the shipment itself, and does not depend on which carrier this shipment is assigned to. The cost for a shipment depends on which carrier it is assigned to. The profit is revenue minus cost.\\

\textbf{Agent:} Perfect. That clarifies the objective. So, the profit for assigning a shipment to a carrier is simply the (fixed revenue for that shipment) minus the (cost of using that carrier for that shipment). Now, I need to understand the constraints. Are there any limits on how many shipments a single carrier can handle? For example, does each carrier have a maximum capacity or a maximum number of shipments they can take?\\

\textbf{User:} Each carrier has a weight capacity. Each shipment has its weight, but the weigh capacity each shipment will occupy on each carrier is not exactly the weight of that shipment. It will be a little deviated from its own weight. So the weight capacity each shipment will occupy on each carrier depends both on the shipment and carrier (varying due to truck configuration and loading efficiency). Each shipment can only be assigned to one carrier\\

\textbf{Agent:} Excellent. That's a crucial detail. So, the capacity consumption of a shipment is specific to the carrier-shipment pair, not just the shipment's own weight. Now, are there any other constraints or requirements? For example,  does every shipment have to be assigned to a carrier, or can some be left unassigned (perhaps at a penalty), are there any other carrier limitations besides weight capacity, like a maximum number of shipments, volume, or specific route restrictions?\\

\textbf{User:} If not all shipments can be assigned to a carrier, you can leave some shipments unassigned and there is no penalty. No other carrier limitations.

\textbf{Agent:} Thank you. That helps simplify the model. Let me summarize my understanding so far to ensure I have the complete structure.\\

\textbf{Summary of the Problem Structure}\\
\begin{enumerate}
\item \textbf{Objective:}

Maximize total profit. Profit for assigning a shipment to a carrier is defined as (revenue of the shipment) minus (cost of assigning that specific shipment to that specific carrier).

\item \textbf{Decision Variables:}

For each possible pair of a shipment and a carrier, the model will decide whether to assign that shipment to that carrier (a binary yes/no decision).

\item \textbf{Constraints:}
\begin{itemize}
\item  Assignment Constraint: Each shipment can be assigned to at most one carrier. (It can be left unassigned).

\item Capacity Constraint: For each carrier, the total capacity consumption of all shipments assigned to that carrier must not exceed that carrier's weight capacity. The capacity consumption for a shipment-carrier pair is a specific value that depends on both the shipment and the carrier.
\end{itemize}

\item \textbf{Parameters Needed (with Indices):}

Let $i$ index the shipments and $j$ index the carriers.
\begin{itemize}
\item \texttt{revenue[i]}: The fixed revenue earned if shipment $i$ is delivered.

\item \texttt{cost[i, j]}: The cost incurred if shipment $i$ is assigned to carrier $j$.

\item \texttt{capacity\_consumption[i, j]}: The amount of carrier $j$'s weight capacity that is consumed if shipment $i$ is assigned to it.

\item \texttt{carrier\_capacity[j]}: The total weight capacity available for carrier $j$.
\end{itemize}
\end{enumerate}

Is there anything else you'd like to add, or anything I may have missed?\\

\textbf{User:} No.

\section{Conversation History between Data Collection Agent and User for Problem in Section \ref{sec:freight_assignment}}
\label{sec:conv_freight_assignment_data_collection}
\textbf{Agent}: Based on our discussion, here are the exact CSV files you need to provide:

\textbf{Required CSV Files}
\begin{enumerate}
\item \texttt{sets.csv}
One row per set member. All sets go here.

\begin{center}
\begin{tabular}{|l|l|l|}
\hline
\textbf{Column} & \textbf{Type} & \textbf{Description} \\
\hline
\texttt{set\_name} & str & The name of the set. Values: \texttt{"shipments"} or \texttt{"carriers"} \\
\texttt{element} & str & The ID of the element (e.g., \texttt{"S\_001"}, \texttt{"S\_002"}, \texttt{"C\_001"} etc.) \\
\hline
\end{tabular}
\end{center}

\textbf{Example:}
\begin{verbatim}
set_name,element
shipments,S_001
shipments,S_002
carriers,C_001
carriers,C_002
\end{verbatim}

\item \texttt{revenue.csv}
Revenue for each shipment. One row per shipment (dense table — every shipment in \texttt{sets.csv} must have a row).

\begin{center}
\begin{tabular}{|l|l|l|}
\hline
\textbf{Column} & \textbf{Type} & \textbf{Description} \\
\hline
\texttt{shipment\_id} & str & The shipment ID (must match an element in sets.csv with set\_name ``shipments") \\
\texttt{revenue} & float & The fixed revenue earned if this shipment is delivered \\
\hline
\end{tabular}
\end{center}

\textbf{Example:}
\begin{verbatim}
shipment_id,revenue
S_001,1500.0
S_002,2200.0
\end{verbatim}

\item \texttt{cost.csv}
Cost for each shipment-carrier pair. One row per possible assignment. If you omit a row for a particular (shipment, carrier) pair, that assignment will be treated as unavailable (infinite cost — not allowed by the optimizer).

\begin{center}
\begin{tabular}{|l|l|l|}
\hline
\textbf{Column} & \textbf{Type} & \textbf{Description} \\
\hline
\texttt{shipment\_id} & str & The shipment ID \\
\texttt{carrier\_id} & str & The carrier ID \\
\texttt{cost} & float & The cost incurred if this shipment is assigned to this carrier \\
\hline
\end{tabular}
\end{center}

\textbf{Example:}
\begin{verbatim}
shipment_id,carrier_id,cost
S_001,C_001,800.0
S_001,C_002,950.0
S_002,C_001,1100.0
S_002,C_002,1000.0
\end{verbatim}

\item \texttt{capacity\_consumption.csv}
Capacity consumption for each shipment-carrier pair. One row per possible assignment. If you omit a row for a particular (shipment, carrier) pair, that pair will be treated as having \textbf{unlimited} capacity consumption (i.e., no restriction — the shipment can be assigned there regardless of capacity).

\begin{center}
\begin{tabular}{|l|l|l|}
\hline
\textbf{Column} & \textbf{Type} & \textbf{Description} \\
\hline
\texttt{shipment\_id} & str & The shipment ID \\
\texttt{carrier\_id} & str & The carrier ID \\
\texttt{capacity\_consumption} & float & The amount of the carrier's weight capacity consumed \\
\hline
\end{tabular}
\end{center}

\textbf{Example:}
\begin{verbatim}
shipment_id,carrier_id,capacity_consumption
S_001,C_001,120.0
S_001,C_002,130.0
S_002,C_001,200.0
S_002,C_002,180.0
\end{verbatim}

\item \texttt{carrier\_capacity.csv}
Capacity limit for each carrier. One row per carrier (dense -- every carrier in \texttt{sets.csv} must have a row).

\begin{center}
\begin{tabular}{|l|l|l|}
\hline
\textbf{Column} & \textbf{Type} & \textbf{Description} \\
\hline
\texttt{carrier\_id} & str & The carrier ID (must match an element in sets.csv with set\_name ``carriers") \\
\texttt{carrier\_capacity} & float & The total weight capacity available for this carrier \\
\hline
\end{tabular}
\end{center}

\textbf{Example:}
\begin{verbatim}
carrier_id,carrier_capacity
C_001,500.0
C_002,400.0
\end{verbatim} 
\end{enumerate}

\textbf{File Relationship Summary}

\begin{center}
\begin{tabular}{|l|l|l|}
\hline
\textbf{File} & \textbf{Index set(s)} & \textbf{Sparsity note} \\
\hline
\texttt{sets.csv} & None & Required: all members listed \\
\texttt{revenue.csv} & Indexed by shipment & Dense — all shipments must have a row \\
\texttt{cost.csv} & Indexed by (shipment, carrier) & Sparse — omitted rows = assignment unavailable \\
\texttt{capacity\_consumption.csv} & Indexed by (shipment, carrier) & Sparse — omitted rows = unlimited consumption \\
\texttt{carrier\_capacity.csv} & Indexed by carrier & Dense — all carriers must have a row \\
\hline
\end{tabular}
\end{center}

Please place all five CSV files in the designated input folder. Once you have placed them, let me know and I will proceed.

Please place the CSV files in: \texttt{ORPilot/data}

Type \textbf{ready} when the files are in place.\\

\textbf{User:} Ready. 

\section{Generated Solver Code for Problem in Section \ref{sec:freight_assignment}}
\label{sec:code_for_freight_assignment}
\begin{lstlisting}[language=Python]
import gurobipy as gp
from gurobipy import GRB

def solve(data, time_limit=None, show_solver_log=False):
    SEP = "\x1f"

    # Load sets from sets.csv
    shipments = [r["element"] for r in data["sets"] if r["set_name"] == "shipments"]
    carriers = [r["element"] for r in data["sets"] if r["set_name"] == "carriers"]

    # Revenue per shipment
    revenue = {r["shipment_id"]: float(r["revenue"]) for r in data["revenue"]}

    # Sparse: only valid (shipment, carrier) pairs from cost table
    cost_pairs = {(r["shipment_id"], r["carrier_id"]): float(r["cost"]) for r in data["cost"]}
    valid_pairs = set(cost_pairs.keys())

    # Capacity consumption per pair (sparse, only valid pairs)
    cap_cons = {(r["shipment_id"], r["carrier_id"]): float(r["capacity_consumption"]) for r in data["capacity_consumption"]}

    # Carrier capacity
    carrier_cap = {r["carrier_id"]: float(r["carrier_capacity"]) for r in data["carrier_capacity"]}

    env = gp.Env(empty=True)
    env.setParam("LogToConsole", 1 if show_solver_log else 0)
    env.start()
    m = gp.Model(env=env)
    if time_limit is not None:
        m.Params.TimeLimit = time_limit

    # Decision variables: x[i,j] binary for valid pairs only
    x = {}
    for i, j in valid_pairs:
        var_name = f"x{SEP}{i}{SEP}{j}"
        x[i, j] = m.addVar(vtype=GRB.BINARY, name=var_name, lb=0, ub=1)
    m.update()

    # Objective: maximize total profit = revenue - cost for assigned shipments
    profit = gp.quicksum(
        (revenue.get(i, 0.0) - cost_pairs[i, j]) * x[i, j]
        for i, j in valid_pairs
    )
    m.setObjective(profit, GRB.MAXIMIZE)

    # Constraint 1: each shipment assigned to at most one carrier
    for i in shipments:
        m.addConstr(
            gp.quicksum(x[i, j] for j in carriers if (i, j) in valid_pairs) <= 1,
            name=f"assignment_{i}"
        )

    # Constraint 2: carrier capacity
    for j in carriers:
        cap = carrier_cap.get(j, float("inf"))
        m.addConstr(
            gp.quicksum(cap_cons.get((i, j), 0.0) * x[i, j] for i in shipments if (i, j) in  
            valid_pairs) <= cap,
            name=f"capacity_{j}"
        )

    try:
        m.write("model.lp")
    except Exception:
        pass
    m.optimize()

    _status_map = {
        GRB.OPTIMAL: "optimal",
        GRB.SUBOPTIMAL: "feasible",
        GRB.INFEASIBLE: "infeasible",
        GRB.UNBOUNDED: "unbounded",
    }
    status = _status_map.get(m.Status, "error")
    obj = m.ObjVal if m.Status in (GRB.OPTIMAL, GRB.SUBOPTIMAL) else None
    variables = {v.VarName: v.X for v in m.getVars()} if status in ("optimal", "feasible") else {}

    _groups = {}
    for key, val in variables.items():
        prefix = key.split(SEP, 1)[0] if SEP in key else key
        _groups.setdefault(prefix, {})[key] = val
    _dim_labels = {"x": ["shipment_id", "carrier_id"]}
    variable_groups = [
        {"group_name": g, "dimension_labels": _dim_labels.get(g, []), "variables": gvars}
        for g, gvars in _groups.items()
    ]

    return {
        "status": status,
        "objective_value": obj,
        "variables": variables,
        "variable_groups": variable_groups,
    }
\end{lstlisting}

\section{Solution Report for Problem in Section \ref{sec:freight_assignment}}
\label{sec:report_for_freight_assignment}

\textbf{Shipment-to-Carrier Assignment for Maximum Profit}
\begin{enumerate}

\item \textbf{Executive Summary}

The optimization model determined the maximum-profit assignment of shipments to carriers across a network of \textbf{400 carriers} and \textbf{3,200 shipments}, with \textbf{1,280,000 valid shipment-carrier pairs} to choose from. The solver found a provably optimal solution with a \textbf{total profit of \$1,158,684.22}.

Every one of the 3,200 shipments was assigned to a carrier, achieving a \textbf{100\% assignment rate}. Total revenue from fulfilled shipments is \textbf{\$1,248,103.30} against a total assignment cost of \textbf{\$89,419.08}, yielding a net profit margin of approximately \textbf{92.8\%}. The optimally low cost-to-revenue ratio reflects the solver's ability to select low-cost carrier-shipment pairs without sacrificing coverage.

\item \textbf{Assignment Decisions}
\begin{itemize}
\item \textbf{Coverage}

\begin{center}
\begin{tabular}{|l|l|}
\hline
\textbf{Metric} & \textbf{Value} \\
\hline
Shipments assigned & 3,200 of 3,200 (100\%) \\
Carriers utilised & 398 of 400 \\
Carriers unused & 2 (C\_0085, C\_0255) \\
Valid pairs considered & 1,280,000 \\
Pairs selected & 3,200 (0.25\% of all options) \\
\hline
\end{tabular}
\end{center}

All shipments are covered, meaning no revenue was left on the table due to capacity constraints. The two unused carriers (C\_0085 and C\_0255) were not cost-competitive — routing shipments through them would have reduced overall profit compared to the alternatives chosen.

\item \textbf{Shipments per Carrier}

Carrier workloads are broadly distributed, with most carriers handling between 6 and 11 shipments:

\begin{center}
\begin{tabular}{|c|c|}
\hline
\textbf{Shipments assigned} & \textbf{Number of carriers} \\
\hline
1 & 3 \\
2 & 10 \\
3 & 16 \\
4 & 26 \\
5 & 33 \\
6 & 46 \\
7 & 42 \\
8 & 53 \\
9 & 39 \\
10 & 41 \\
11 & 37 \\
12 & 17 \\
13 & 13 \\
14 & 11 \\
15 & 7 \\
16 & 3 \\
18 & 1 \\
\hline
\end{tabular}
\end{center}

The busiest carrier (C\_0146) handles 18 shipments. Three carriers handle only a single shipment, suggesting they were selected to absorb specific shipments that could not be routed more cheaply through other carriers within capacity limits.
\end{itemize}
\item \textbf{Financial Summary}

\begin{center}
\begin{tabular}{|l|l|}
\hline
\textbf{Item} & \textbf{Amount} \\
\hline
Total revenue from assigned shipments & \$1,248,103.30 \\
Total assignment cost & \$89,419.08 \\
\textbf{Net profit (objective value)} & \textbf{\$1,158,684.22} \\
Average revenue per shipment & \$390.03 \\
Average cost per assignment & \$27.94 \\
Average profit per assignment & \$362.09 \\
\hline
\end{tabular}
\end{center}

Assignment costs range from \textbf{\$20.56 to \$50.64} per shipment-carrier pair, while shipment revenues range from \textbf{\$57.43 to \$2,181.03}. Because even the lowest-revenue shipment generates a strong positive margin over the highest possible assignment cost, the model assigns all shipments rather than leaving any unserved.

\item \textbf{Carrier Capacity Utilisation}

The average utilisation across the 398 active carriers is \textbf{76.7\%}, indicating healthy but not excessive network loading overall.

\textbf{73 carriers are running at or near full capacity ($\ge$99\% utilised)}, including:

\begin{center}
\begin{tabular}{|l|c|c|}
\hline
\textbf{Carrier} & \textbf{Utilisation} & \textbf{Used / Total} \\
\hline
C\_0039 & 100.0\% & 694 / 694 \\
C\_0159 & 100.0\% & 706 / 706 \\
C\_0341 & 100.0\% & 919 / 919 \\
C\_0190 & 100.0\% & 1,026 / 1,026 \\
C\_0181 & 100.0\% & 1,151 / 1,151 \\
\hline
\end{tabular}
\end{center}

These fully loaded carriers are binding constraints — any additional demand requiring their involvement could not be satisfied without exceeding capacity limits. At the other end, a small number of active carriers have very low utilisation (ranging from 10.3\% to 14.9\% for the bottom five), selected for their optimal cost-revenue profile for specific shipments despite significant remaining headroom.

\item \textbf{Notable Findings and Concerns}

\begin{itemize}
    \item \textbf{100\% shipment coverage}: The carrier network has sufficient total capacity to accommodate all 3,200 shipments at optimality. There is no unmet demand at current capacity levels.
    \item \textbf{73 carriers at full capacity}: Nearly 18\% of active carriers are fully loaded. Any growth in shipment volume could trigger capacity shortfalls and force rerouting through higher-cost alternatives.
    \item \textbf{2 unused carriers}: C\_0085 and C\_0255 were not competitive. Verify if they activate under a different demand mix or if renegotiated rates would make them viable.
    \item \textbf{Low cost-to-revenue ratio (7.2\%)}: Profit is primarily driven by revenue rather than cost minimisation. Revenue risk deserves more attention than cost risk in scenario planning.
    \item \textbf{Highly concentrated top carrier (C\_0146)}: While manageable, reliance on a single carrier for 18 shipments creates a dependency. Disruption would require reassigning these shipments potentially into already-full carriers.
\end{itemize}

\item \textbf{Suggested Next Steps}

\begin{itemize}
    \item \textbf{Capacity sensitivity analysis}: Reduce carrier capacities by 10--20\% to identify which shipments would first go unassigned and quantify profit loss.
    \item \textbf{Cost renegotiation}: Determine the rate reduction needed for C\_0085 and C\_0255 to bring them into the optimal solution.
    \item \textbf{Demand growth planning}: Add 100--200 new shipments to find network saturation points.
    \item \textbf{Carrier disruption scenario}: Remove C\_0146 from the set to measure the impact of losing the highest-utilisation carrier.
    \item \textbf{Revenue threshold analysis}: Identify the shadow price of capacity to rank shipments by strategic importance.
\end{itemize}

\item \textbf{Output Files}
\begin{itemize}
\item \texttt{solution\_x.csv}
Records the binary assignment decision for every valid shipment-carrier pair (1 = assigned, 0 = not assigned). Only the 3,200 rows with value = 1 represent active assignments. Use this file to generate carrier dispatch instructions and routing confirmations.

Columns: \texttt{shipment\_id}, \texttt{carrier\_id}, \texttt{value} (1 if assigned, 0 otherwise).
\end{itemize}
\end{enumerate}

\section{Generated IR for Problem in Section \ref{sec:freight_assignment}}
\label{sec:ir_assignment}
\begin{lstlisting}[language=Python]
{
  "problem_class": "Assignment",
  "model_type": "Integer Program",
  "sense": "maximize",
  "sets": {
    "Shipments": {
      "size": null,
      "index_symbol": "i",
      "source": "sets.csv",
      "column": "element",
      "filter_column": "set_name",
      "filter_value": "shipments",
      "ordered": false
    },
    "Carriers": {
      "size": null,
      "index_symbol": "j",
      "source": "sets.csv",
      "column": "element",
      "filter_column": "set_name",
      "filter_value": "carriers",
      "ordered": false
    }
  },
  "parameters": {
    "revenue": {
      "domain": [
        "Shipments"
      ],
      "type": "float",
      "source": "revenue.csv",
      "column": "revenue",
      "index_columns": [
        "shipment_id"
      ],
      "missing_default": "zero"
    },
    "cost": {
      "domain": [
        "Shipments",
        "Carriers"
      ],
      "type": "float",
      "source": "cost.csv",
      "column": "cost",
      "index_columns": [
        "shipment_id",
        "carrier_id"
      ],
      "missing_default": "inf"
    },
    "capacity_consumption": {
      "domain": [
        "Shipments",
        "Carriers"
      ],
      "type": "float",
      "source": "capacity_consumption.csv",
      "column": "capacity_consumption",
      "index_columns": [
        "shipment_id",
        "carrier_id"
      ],
      "missing_default": "inf"
    },
    "carrier_capacity": {
      "domain": [
        "Carriers"
      ],
      "type": "float",
      "source": "carrier_capacity.csv",
      "column": "carrier_capacity",
      "index_columns": [
        "carrier_id"
      ],
      "missing_default": "inf"
    }
  },
  "variables": {
    "x": {
      "description": "Binary assignment of shipment i to carrier j",
      "label": "assignment",
      "domain": [
        "Shipments",
        "Carriers"
      ],
      "type": "binary",
      "lower_bound": 0,
      "upper_bound": 1,
      "upper_bound_set": null,
      "exclude_diagonal": false,
      "domain_filter": "cost"
    }
  },
  "constraints": {
    "assignment_limit": {
      "domain": [
        "Shipments"
      ],
      "expression": {
        "operation": "indexed_sum",
        "over": [
          "Carriers"
        ],
        "body": {
          "type": "variable",
          "name": "x",
          "indices": [
            "i",
            "j"
          ]
        }
      },
      "sense": "<=",
      "rhs": {
        "type": "constant",
        "value": 1
      }
    },
    "carrier_capacity_constraint": {
      "domain": [
        "Carriers"
      ],
      "expression": {
        "operation": "indexed_sum",
        "over": [
          "Shipments"
        ],
        "body": {
          "operation": "multiply",
          "left": {
            "type": "parameter",
            "name": "capacity_consumption",
            "indices": [
              "i",
              "j"
            ]
          },
          "right": {
            "type": "variable",
            "name": "x",
            "indices": [
              "i",
              "j"
            ]
          }
        }
      },
      "sense": "<=",
      "rhs": {
        "type": "parameter",
        "name": "carrier_capacity",
        "indices": [
          "j"
        ]
      },
      "sparse_filter": "carrier_capacity"
    }
  },
  "objective": {
    "sense": "maximize",
    "expression": {
      "operation": "indexed_sum",
      "over": [
        "Shipments",
        "Carriers"
      ],
      "body": {
        "operation": "multiply",
        "left": {
          "operation": "subtract",
          "left": {
            "type": "parameter",
            "name": "revenue",
            "indices": [
              "i"
            ]
          },
          "right": {
            "type": "parameter",
            "name": "cost",
            "indices": [
              "i",
              "j"
            ]
          }
        },
        "right": {
          "type": "variable",
          "name": "x",
          "indices": [
            "i",
            "j"
          ]
        }
      }
    }
  }
}
\end{lstlisting}
\end{appendices}

\end{document}